\documentclass[10pt,journal,compsoc]{IEEEtran}
\ifCLASSOPTIONcompsoc
  \usepackage[nocompress]{cite}
\else
  \usepackage{cite}
\fi
\ifCLASSINFOpdf
\else
\fi
\usepackage{times}
\usepackage{epsfig}
\usepackage{graphicx}
\usepackage{amsmath}
\usepackage{amssymb}
\usepackage{array}
\usepackage{multirow}
\usepackage{caption}
\usepackage{xcolor}
\usepackage{color, colortbl}
\usepackage{cite}
\usepackage{subcaption}
\usepackage{adjustbox}
\usepackage[pagebackref,breaklinks,colorlinks]{hyperref}
\definecolor{LightGray}{rgb}{0.94,0.94,0.94}
\definecolor{XL_color}{rgb}{0.858, 0.188, 0.478}

\newcommand{\VA}[1]{\textcolor{black}{#1}}

\newcommand{\vect}[1]{\boldsymbol{#1}}

\newcommand{\minisection}[1]{\vspace{0.5mm}\noindent{\textbf{#1}.}}

\newcommand{\Paragraph}[1]{\vspace{0.5mm} \noindent \textbf{#1} \hspace{0mm}}
\newcommand{\etal}{\textit{et~al}.}
\newcommand{\ie}{\textit{i}.\textit{e}.}
\newcommand{\eg}{\textit{e}.\textit{g}.}
\newcommand{\vs}{\textit{vs}. }

\DeclareMathOperator*{\argmin}{arg\,min}

\usepackage{xcolor,pifont}
\newcommand*\colourcheck[1]{%
  \expandafter\newcommand\csname #1check\endcsname{\textcolor{#1}{\ding{52}}}%
}
\newcommand*\colourcross[1]{%
  \expandafter\newcommand\csname #1check\endcsname{\textcolor{#1}{\ding{55}}}%
}
\colourcheck{green}
\colourcross{red}

\newcolumntype{P}[1]{>{\centering\arraybackslash}p{#1}}

\begin{document}
\title{Reverse Engineering of Generative Models:\\Inferring Model Hyperparameters from Generated Images}

\author{Vishal Asnani \thanks{Vishal Asnani and Xiaoming Liu are with the Department of Computer Science and Engineering at Michigan State University. Xi Yin and Tal Hassner are with Meta AI. All data, experiments, and code were collected, performed, and developed at Michigan State University.}, Xi Yin, Tal Hassner, Xiaoming Liu }

\markboth{Asnani \textit{et al.} Reverse Engineering of Generative Models}%
{Shell \MakeLowercase{\textit{et al.}}: Bare Demo of IEEEtran.cls for Computer Society Journals}



\IEEEtitleabstractindextext{%
\begin{abstract}
State-of-the-art (SOTA) Generative Models (GMs) can synthesize photo-realistic images that are hard for humans to distinguish from genuine photos. Identifying and understanding manipulated media are crucial to mitigate the social concerns on the potential misuse of GMs. We propose to perform reverse engineering of GMs to infer model hyperparameters from the images generated by these models. We define a novel problem, ``model parsing", as estimating GM network architectures and training loss functions by examining their generated images -- a task seemingly impossible for human beings. To tackle this problem, we propose a framework with two components: a Fingerprint Estimation Network (FEN), which estimates a GM fingerprint from a generated image by training with four constraints to encourage the fingerprint to have desired properties, and a Parsing Network (PN), which predicts network architecture and loss functions from the estimated fingerprints. To evaluate our approach, we collect a fake image dataset with $100$K images generated by $116$ different GMs. Extensive experiments show encouraging results in parsing the hyperparameters of the unseen models. Finally, our fingerprint estimation can be leveraged for deepfake detection and image attribution, as we show by reporting SOTA results on both the deepfake detection (Celeb-DF) and image attribution benchmarks.
\end{abstract}

}

\maketitle

\IEEEdisplaynontitleabstractindextext

% \fi
\IEEEpeerreviewmaketitle

\IEEEraisesectionheading{\section{Introduction}\label{sec:introduction}}

%\secvspace
Image generation techniques have improved significantly in recent years, especially after the breakthrough of Generative Adversarial Networks (GANs)~\cite{NIPS2014_5423}. Many Generative Models (GMs), including both GAN and Variational Autoencoder (VAE)~\cite{64,61,57,42,43,45,dhariwal2021diffusion}, can generate photo-realistic images that are hard for \VA{humans} to distinguish from genuine photos. This photo-realism, however, raises increasing concerns for the potential misuse of these models, \eg, by launching coordinated misinformation attack~\cite{Disinformation2020,From2019}. As a result, deepfake detection~\cite{1,2,3,5,7,nirkin2020deepfake} has recently attracted \VA{growing} attention. Going beyond the binary genuine \vs fake classification as in deepfake detection, Yu~\etal~\cite{11} proposed source model classification given a generated image. This {\em image attribution} problem assumes a {\it closed set} of GMs, used in both training and testing.

\begin{figure}[t!]
\begin{center}
\includegraphics[width=\columnwidth]{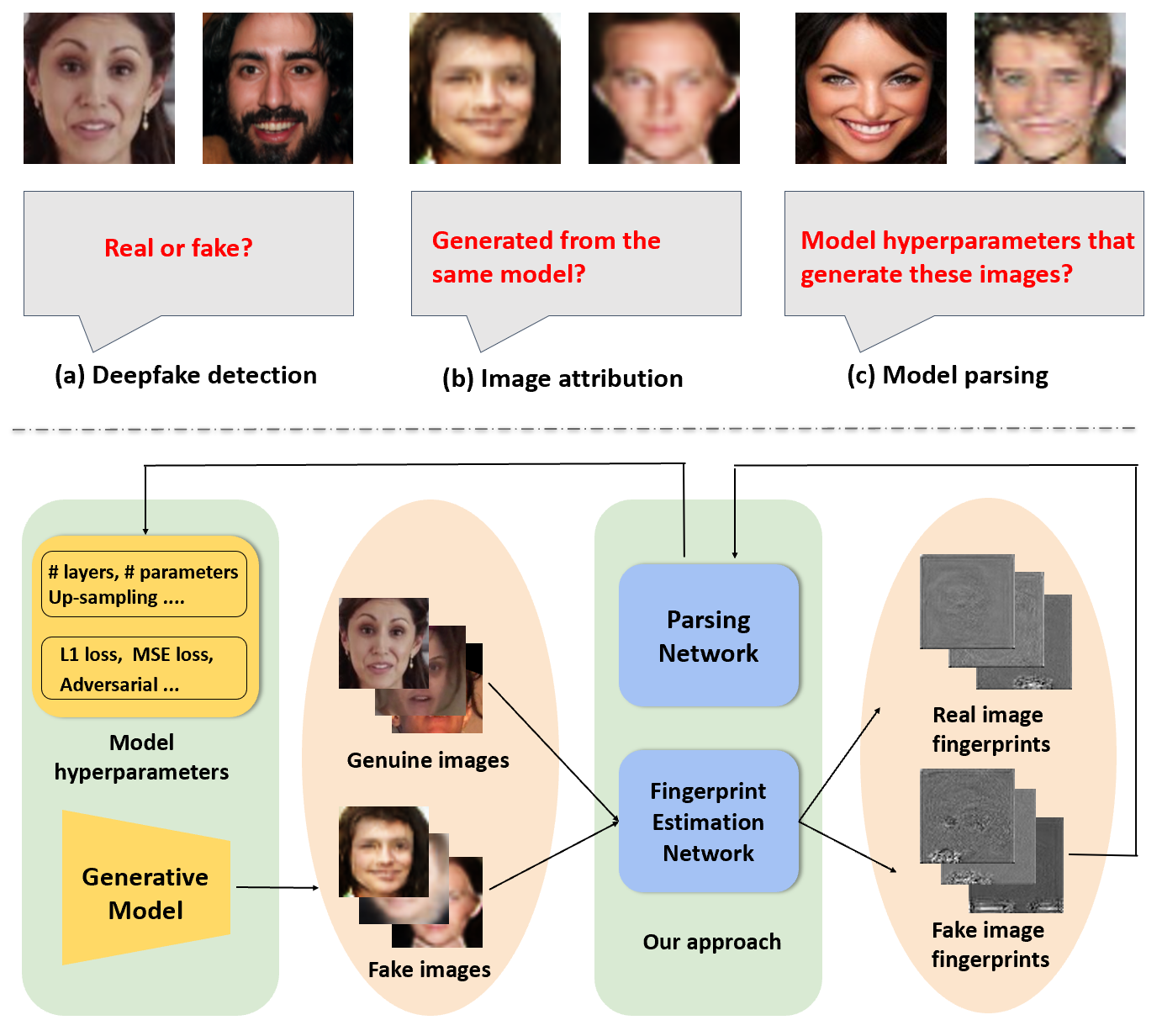}
\caption{\small Top: Three increasingly difficult tasks: (a) {\em deepfake detection} classifies an image as genuine or fake; (b) {\em image attribution} predicts which of a closed set of GMs generated a fake image; and (c) {\em model parsing}, proposed here, infers hyperparameters of the GM used to generate an image, for those models unseen during training. Bottom: We present a framework for model parsing, which can also be applied to simpler tasks of deepfake detection and image attribution.}
\label{fig:teaser}
\end{center}
\end{figure}

It is desirable to generalize image attribution to open-set recognition, {\it i.e.}, classify an image generated by GMs which were {\it not} seen during training. However, one may wonder what else \VA{we can do} beyond recognizing a GM as an {\em unseen} or {\em new} model. Can we know more about how this new GM was designed? How its architecture differs from known GMs in the training set? Answering these questions is valuable when we, as defenders, strive to understand the source of images generated by malicious attackers or identify coordinated misinformation attacks which use the same GM. We view this as the grand challenge of reverse engineering of GMs.

%\vspace{-1mm}
% To promote reproduction of our results, we will publicly release our dataset, source code, and trained models. 
%%
%% =======================================================
%%

\begin{table*}[t!]
%\begin{center}
\centering
\caption{\small Comparison of our approach with prior works on reverse engineering of models, fingerprint estimation and deepfake detection. We compare on the basis of \VA{input} and output of methods, whether the testing is done on multiple unseen GMs and whether the testing is done on multiple datasets. [KEYS: R.E.: reverse engineering, I.A.: image attribution, D.D.: deepfake detection, Fing. est.: fingerprint estimation, mul.: multiple, un.: unknown, N.A.: network architecture, L.F.: Loss function, para.: parameters, sup.: supervised, unsup.: unsupervised]}
\scalebox{1}{
\begin{tabular}{P{2cm}|P{2cm}|P{3.5cm}|P{2cm}|P{0.7cm}|P{1.2cm}|P{1.2cm}|P{1.2cm}}
\hline
Method (Year) & Purpose & Input & Output & Fing. est.  & Test on mul. GMs & Test on un. GMs & Test on mul. data\\\hline
\cite{14} ($2016$) & R.E. & Attack on models & Training data & \redcheck & \redcheck & \redcheck & \redcheck\\
\cite{15} ($2018$) & R.E. & Input-output images & N.A. para. & \redcheck & \redcheck & \redcheck & \redcheck\\
\cite{18} ($2018$) & R.E. & Memory access patterns & Model weights & \redcheck & \redcheck & \redcheck & \redcheck\\
\cite{19} ($2018$) & R.E. & Electromagnetic emanations &  N.A. para. & \redcheck & \redcheck & \redcheck & \redcheck\\
\cite{10} ($2019$) & I.A. & Image & \redcheck & Sup. & \greencheck & \redcheck & \greencheck\\
\cite{11} ($2019$) & I.A. & Image& \redcheck  & Sup. &  \greencheck & \greencheck & \greencheck\\
\cite{12} ($2020$) & I.A. & Image& \redcheck  & Sup. & \greencheck & \redcheck & \greencheck\\
\cite{13} ($2019$) & I.A. & Image& \redcheck  & Sup. &  \greencheck & \redcheck & \greencheck\\
\cite{1} ($2019$) & D.D. & Image& \redcheck & \redcheck &  \redcheck & \redcheck & \greencheck\\
\cite{3} ($2020$) & D.D. & Image& \redcheck & \redcheck &  \redcheck & \redcheck & \greencheck\\
\cite{2} ($2019$) & D.D. & Image& \redcheck & \redcheck &  \redcheck & \redcheck & \greencheck\\
\cite{5} ($2019$) & D.D. & Image& \redcheck & \redcheck &  \redcheck & \redcheck & \greencheck\\
\cite{7} ($2020$) & D.D. & Image& \redcheck & \redcheck &  \redcheck & \redcheck & \greencheck\\
\cite{nirkin2020deepfake} ($2020$) & D.D. & Image& \redcheck & \redcheck &  \redcheck & \redcheck & \greencheck\\
\cite{masi2020two} ($2020$) & D.D. & Image& \redcheck & \redcheck &  \redcheck & \redcheck & \greencheck\\
\cite{liu2021spatial} ($2021$) & D.D. & Image& \redcheck & \redcheck &  \redcheck & \redcheck & \greencheck\\
Ours ($2022$) & R.E., I.A.,D.D. & Image & N.A. \& L.F. para. &  Unsup. & \greencheck &  \greencheck & \greencheck\\
 \hline \hline
\end{tabular}}
\label{tab:rel_works}
%\end{center}
\end{table*}

While image attribution of GMs is both exciting and challenging, our work aims to take one step further with the following observation. When different GMs are designed, they mainly differ in their model hyperparameters, including the network architectures ({\it e.g.}, the number of layers/blocks, the type of normalization) and training loss functions.
If we could map the generated images to the embedding space of the model hyperparameters used to generate them, there is a potential to tackle a new problem we termed as {\em model parsing}, {\it i.e.}, estimating hyperparameters of an {\it unseen} GM from only its generated image (Figure~\ref{fig:teaser}).
Reverse engineering machine learning models has been done before by relying on a model's input and output~\cite{14,15}, or accessing the hardware usage during inference~\cite{18,19}. 
To the best of our knowledge, however, reverse engineering has not been explored for GMs, especially with only generated images as input.

There are many publicly available GMs that generate images of diverse contents, including faces, digits, and generic scenes. 
%Our framework is not specific to a particular content. 
To improve the generalization of model parsing, we collect a large-scale fake image dataset with various contents so that our framework is not specific to a particular content.
It consists of images generated from $116$ CNN-based GMs, including $81$ GANs, $13$ VAEs, $6$ Adversarial Attack models (AAs), $11$ Auto-Regressive models (ARs) and $5$ Normalizing Flow models (NFs). 
%Although AA model is different from a GM, 
While GANs or VAEs  generate an image by feeding a genuine image or latent code to the network, AAs modify a genuine image based on its objectives via back-propagation. 
%ARs and NFs generate images by adopting different methods.
ARs  generate each pixel of a fake image sequentially, and NFs generate images via a flow-based function. 
Despite such differences, we call all these models as GMs for simplicity.  
%Each GM generates $1,000$ images, which are included in our dataset. 
For each GM, our dataset includes $1,000$ generated images.
We use each model's hyperparameters, including network architecture parameters and training loss types, as the ground-truth for model parsing training. 
We propose a framework to peek inside the black boxes of these GMs by estimating their hyperparameters from the generated images.
Unlike the closed-set setting in~\cite{11}, we venture into quantifying the generalization ability of our method in parsing {\it unseen} GMs.

Our framework consists of two components (Figure~\ref{fig:teaser}, bottom). A {\em Fingerprint Estimation Network} (FEN) infers the subtle yet unique patterns left by GMs on their generated images. Image fingerprint was first applied to images captured by camera sensors~\cite{8,30,31,32,33,34,35} and then extended to GMs~\cite{10,11}. 
%Prior works estimate fingerprints of GMs either by hand-crafted features~\cite{10} or learnt features~\cite{11,12,13}. 
We estimate fingerprints using different constraints which are based on the general properties of fingerprint, including the fingerprint magnitude, repetitive nature, frequency range and symmetrical frequency response. 
Different loss functions are defined to apply these constraints so that the estimated fingerprints manifest these desired properties. 
These constraints enable us to estimate fingerprints of GMs without ground truth.
%These constraints are useful in improving model generalization. 

The estimated fingerprints are discriminative and can serve as the cornerstone for subsequent tasks.
The second part of our framework is a {\em Parsing Network} (PN), which takes the fingerprint as input and predicts the model hyperparameters. 
We consider parameters representing network architectures and loss function types. 
For the former, we form $15$ parameters and categorize them into discrete and continuous types. 
For the latter, we form a $10$-dimensional vector where each parameter represents the usage of a particular loss function type. 
Classification is used for estimating discrete parameters such as the normalization type, and regression is used for continuous parameters such as the number of layers. 
To leverage the similarity between different GMs, we group the GMs into several clusters based on their ground-truth hyperparameters. The mean and deviation are calculated for each GM. We use two different parsers: cluster parser and instance parser to predict the mean and deviation of these parameters, which are then combined as the final predictions.
%We propose to estimate these parameters by adopting the idea of estimating a mean and a deviation prediction. We group the GMs into various clusters using the concatenated ground truth vectors for network architecture and loss function. The formed clusters are then used to estimate a mean using a cluster parser. Further, we use another parser to estimate a deviation which is then combined with mean to output model hyperparameters. 

%Specifically, $18$ types of parameters representing the network architectures are normalized and form a $18$-D vector. $L_2$ loss is used for network architecture prediction. 
%For loss function type prediction, we group nine different loss functions into three categories in a novel hierarchical learning, designed to make training easier.  

Among the $116$ GMs in our collected dataset, there are $47$ models for face generation and $69$ for non-face image generation. We partition all GMs into two categories: face \vs non-face. We carefully curate four evaluation sets for face and non-face categories respectively, where every set well represents the GM population. 
Cross-validation is used in our experiments. 
%To mimic real-world applications where testing images may come from unseen GMs, we perform leave-one-out experiments on our collected dataset. Our framework works remarkably well on such a challenging task, presumably impossible for humans: We achieve an $L_1$ error of $0.179$ for network architecture prediction and $72.5\%$ accuracy in loss function classification. 
%In addition to model parsing, our FEN can be used for deepfake detection and image attribution.
In addition to model parsing, our FEN can be used for deepfake detection and image attribution.
%, which are easier tasks than model parsing. We evaluate our framework under these settings to compare with the state of the art (SOTA). 
For both tasks, we add a shallow network that inputs the estimated fingerprint and performs binary (deepfake detection) or multi-class classification (image attribution). Although our FEN is not tailored for these tasks, we still achieve state-of-the-art (SOTA) performance, indicating the superior generalization ability of our fingerprint estimation. 
Finally, in coordinated misinformation attack,  attackers may use the same GM to generate multiple fake images. 
To detect such attacks, we also define a new task to evaluate how well our model parsing results can be used to determine if two fake images are generated from the same GM.

In summary, this paper makes the following contributions. 
%\vspace{-1mm}
\begin{itemize}
\item We are the first to go beyond model classification by formulating a novel problem of model parsing for GMs. 

\item We propose a novel framework with fingerprint estimation and clustering of GMs to predict the network architecture and loss functions, given a single generated image. 

\item We assemble a dataset of generated images from $116$ GMs, including ground-truth labels on the network architectures and loss function types. % and demonstrate the effectiveness of our proposed approach on this dataset.

\item We show promising results for model parsing and our fingerprint estimation generalizes well to deepfake detection on the Celeb-DF benchmark~\cite{21} and image attribution~\cite{11}, in both cases reporting results comparable or better than existing SOTA~\cite{7,11}.
The parsed model parameters can also be used in detecting coordinated misinformation attacks.

\end{itemize}

\section{Related work}
%\secvspace

\minisection{Reverse engineering of models} There is a growing area of \VA{interest in} reverse engineering the hyperparameters of machine learning models, with two types of approaches. 
First, some methods treat a model as a black box API by examining its input and output pairs.
For example, Tramer~\etal~\cite{14} developed an avatar method to estimate training data and model architectures, while Oh~\etal~\cite{15} trained a set of while-box models to estimate model hyperparameters. 
The second type of \VA{approach} assumes that the intermediate hardware information is available during model inference.
Hua~\etal~\cite{18} estimated both the structure and the weights of a CNN model running on a hardware accelerator, by using information leaks of memory access patterns. Batina~\etal~\cite{19} estimated the network architecture by using side-channel information such as timing and electromagnetic emanations.

Unlike prior methods which require access to the models or their inputs, our approach can reverse engineer GMs by examining {\it only} the images generated by these models, making it more suitable for real-world applications. We summarize our approach with previous works in Tab.~\ref{tab:rel_works}.

\minisection{Fingerprint estimation} Every acquisition device leaves a subtle but unique pattern on its captured image, due to manufacturing imperfections. Such patterns are referred to as {\em device fingerprints}. Device fingerprint estimation~\cite{8,9} was extended to fingerprint estimation of GMs by Marra~\etal~\cite{10}, who showed that hand-crafted fingerprints are unique to each GM and can be used to identify an image's source. Ning~\etal~\cite{11} extended this idea to learning-based fingerprint estimation. Both methods rely on the noise signals in the image. 
Others explored frequency domain information. 
For example, Wang~\etal~\cite{12} showed that CNN generated images have unique patterns in their frequency domain, regarded as model fingerprints. 
Zhang~\etal~\cite{13} showed that features extracted from the middle and high frequencies of the spectrum domain were useful in detecting upsampling artifacts produced by GANs. 

Unlike prior methods which derive fingerprints directly from noise signals or the frequency domain, we propose several novel loss functions to learn GM fingerprints in an unsupervised manner (Tab.~\ref{tab:rel_works}). We further show that our fingerprint estimation can generalize well to other related tasks.

\minisection{Deepfake detection} Deepfake detection is a new and active field with many recent developments. Rossler~\etal~\cite{1} evaluated different methods for detecting face and mouth replacement manipulation. 
Others proposed SVM classifiers on colour difference features~\cite{2}. Guarnera~\etal~\cite{3} used Expectation Maximization~\cite{4} algorithm to extract features and convolution traces for classification. 
Marra~\etal~\cite{5} proposed a multi-task incremental learning to classify new GAN generated images. Chai~\etal ~\cite{chai2020makes} introduced a patch-based classifier to exaggerate regions that are more easily detectable. An attention mechanism~\cite{6} was proposed by Hao~\etal~\cite{7} to improve the performance of deepfake detection.
Masi~\etal~\cite{masi2020two} amplifies the artifacts produced by deepfake methods to perform the detection. 
Nirkin~\etal~\cite{nirkin2020deepfake} seek discrepancies between face regions and their context~\cite{nirkin2018face} as telltale signs of manipulation.
Finally, Liu~\cite{liu2021spatial} uses the spatial information as an additional channel for the classifier.
In our work, the estimated fingerprint is fed into a classifier for genuine \vs~fake classification.

\begin{figure}[t]
\centering
\resizebox{\columnwidth}{!}{
\includegraphics[]{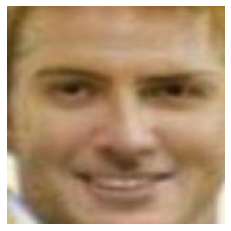}\!
\includegraphics[]{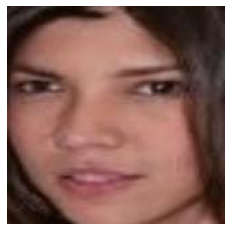}\!
\includegraphics[]{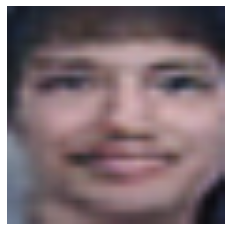}\!
\includegraphics[]{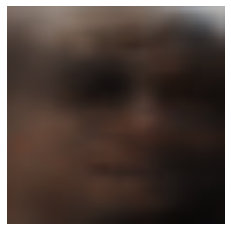}\!
\includegraphics[]{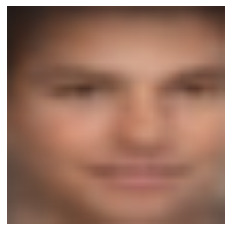}\!
\includegraphics[]{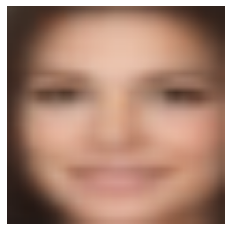}\!
\includegraphics[]{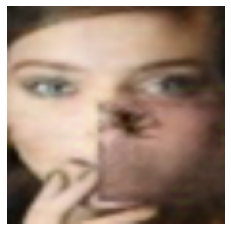}\!
\includegraphics[]{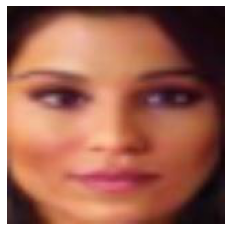}\!
\includegraphics[]{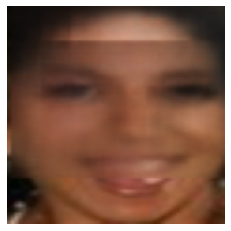}\!
\includegraphics[]{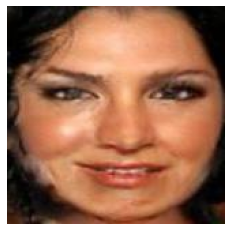}
}

\resizebox{\columnwidth}{!}{
\includegraphics[]{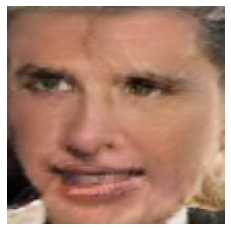}\!
\includegraphics[]{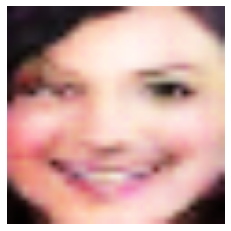}\!
\includegraphics[]{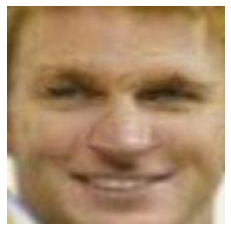}\!
\includegraphics[]{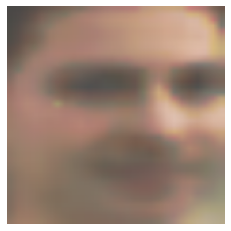}\!
\includegraphics[]{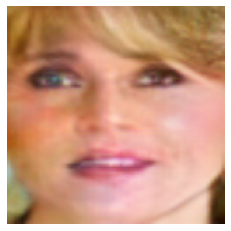}\!
\includegraphics[]{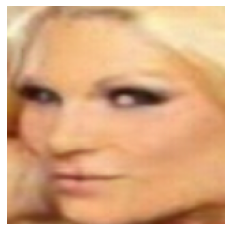}\!
\includegraphics[]{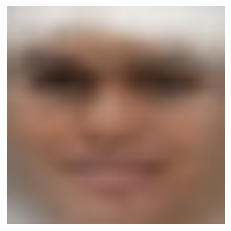}\!
\includegraphics[]{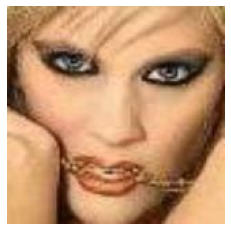}\!
\includegraphics[]{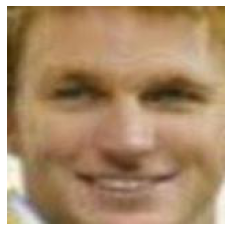}\!
\includegraphics[]{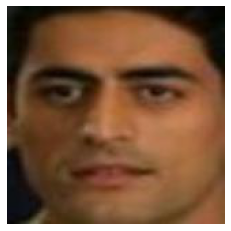}
}

\resizebox{\columnwidth}{!}{
\includegraphics[]{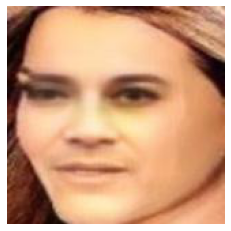}\!
\includegraphics[]{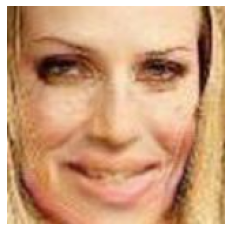}\!
\includegraphics[]{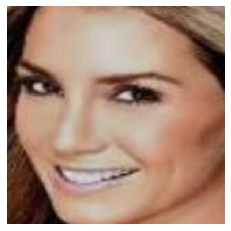}\!
\includegraphics[]{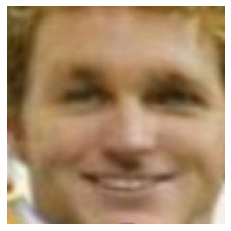}\!
\includegraphics[]{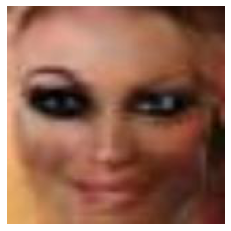}\!
\includegraphics[]{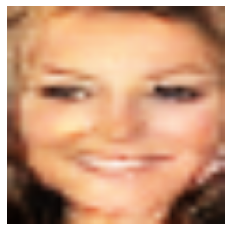}\!
\includegraphics[]{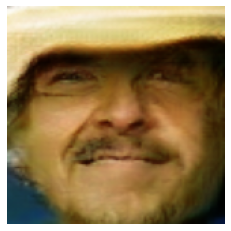}\!
\includegraphics[]{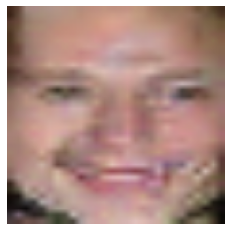}\!
\includegraphics[]{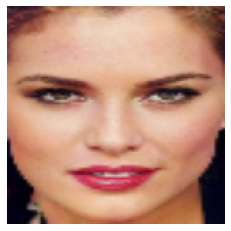}\!
\includegraphics[]{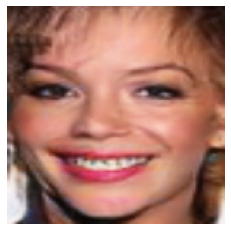}
}

\resizebox{\columnwidth}{!}{
\includegraphics[]{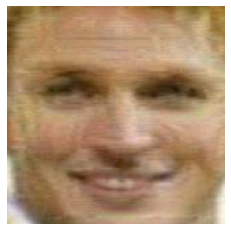}\!
\includegraphics[]{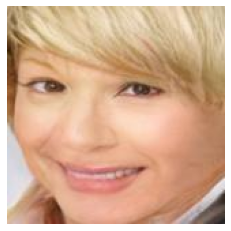}\!
\includegraphics[]{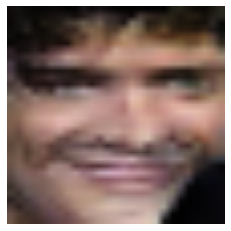}\!
\includegraphics[]{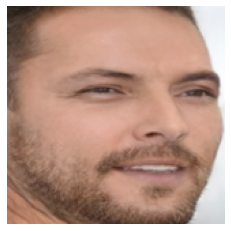}\!
\includegraphics[]{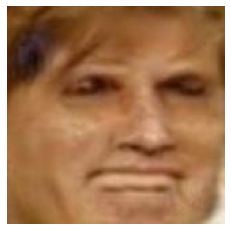}\!
\includegraphics[]{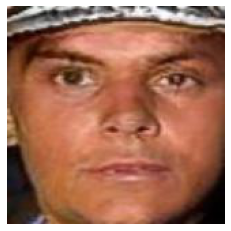}\!
\includegraphics[]{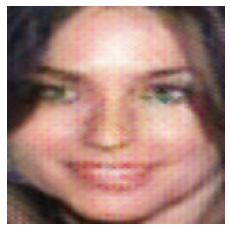}\!
\includegraphics[]{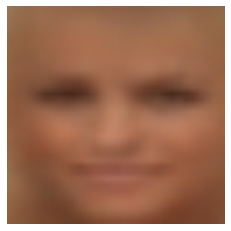}\!
\includegraphics[]{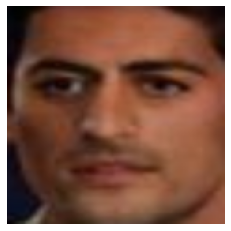}\!
\includegraphics[]{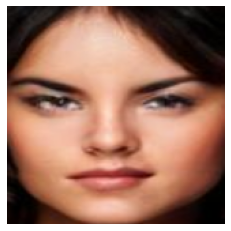}
}

\resizebox{\columnwidth}{!}{
\includegraphics[]{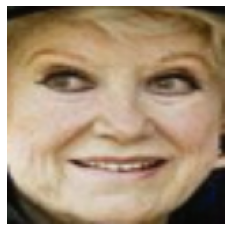}\!
\includegraphics[]{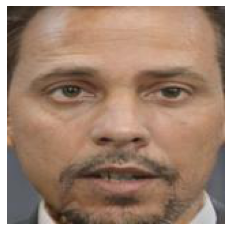}\!
\includegraphics[]{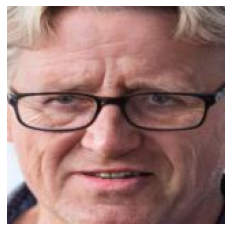}\!
\includegraphics[]{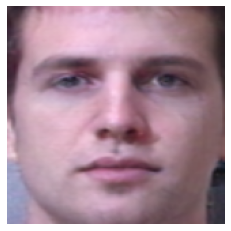}\!
\includegraphics[]{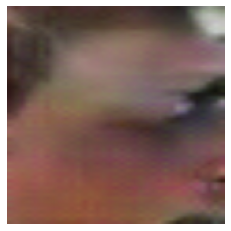}\!
\includegraphics[]{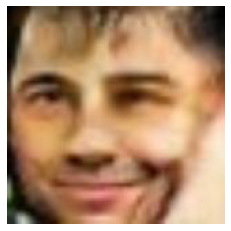}\!
\includegraphics[]{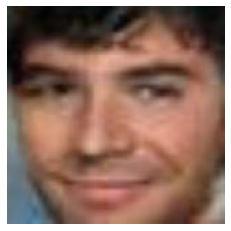}\!
\includegraphics[]{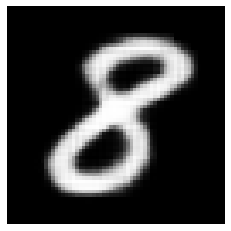}\!
\includegraphics[]{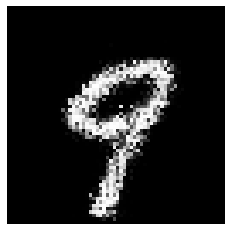}\!
\includegraphics[]{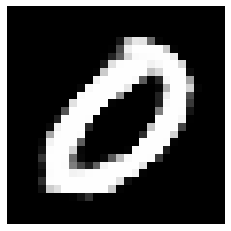}
}

\resizebox{\columnwidth}{!}{
\includegraphics[]{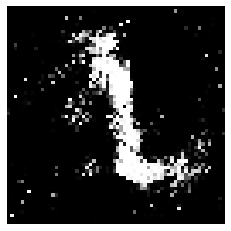}\!
\includegraphics[]{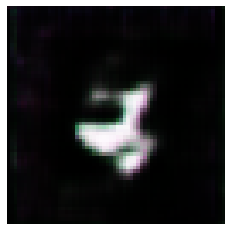}\!
\includegraphics[]{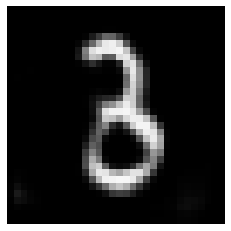}\!
\includegraphics[]{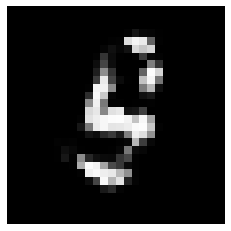}\!
\includegraphics[]{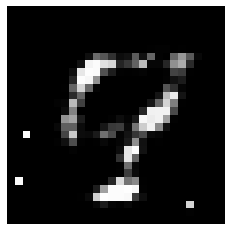}\!
\includegraphics[]{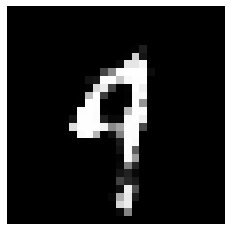}\!
\includegraphics[]{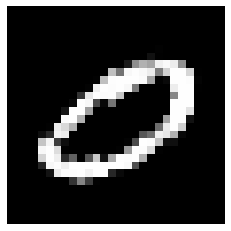}\!
\includegraphics[]{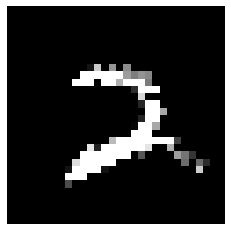}\!
\includegraphics[]{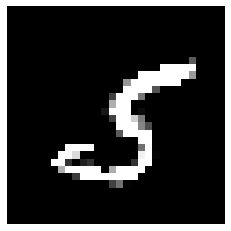}\!
\includegraphics[]{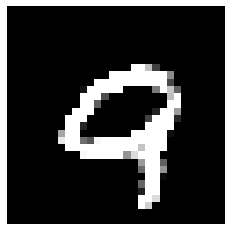}
}

\resizebox{\columnwidth}{!}{
\includegraphics[]{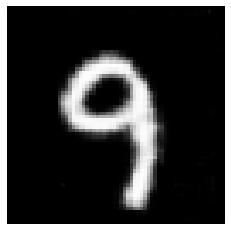}\!
\includegraphics[]{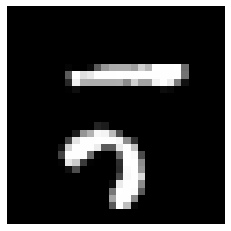}\!
\includegraphics[]{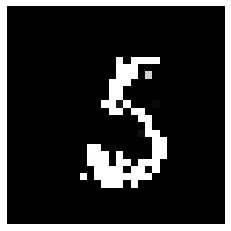}\!
\includegraphics[]{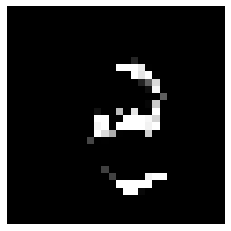}\!
\includegraphics[]{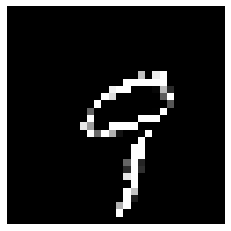}\!
\includegraphics[]{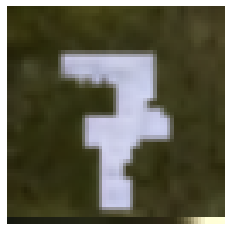}\!
\includegraphics[]{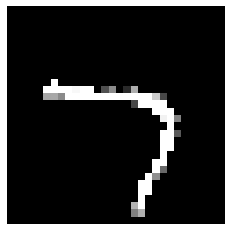}\!
\includegraphics[]{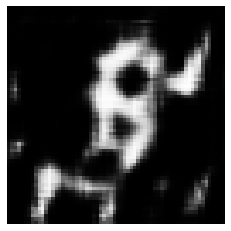}\!
\includegraphics[]{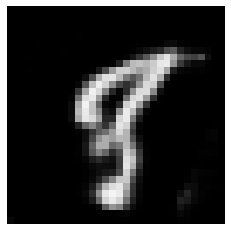}\!
\includegraphics[]{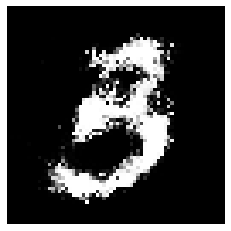}
}

\resizebox{\columnwidth}{!}{
\includegraphics[]{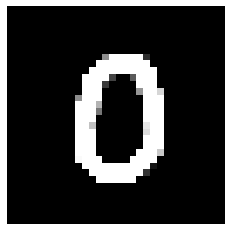}\!
\includegraphics[]{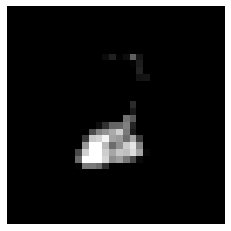}\!
\includegraphics[]{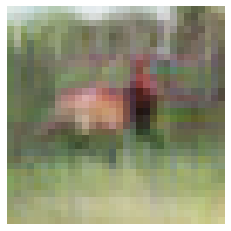}\!
\includegraphics[]{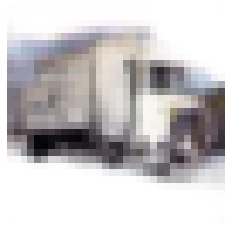}\!
\includegraphics[]{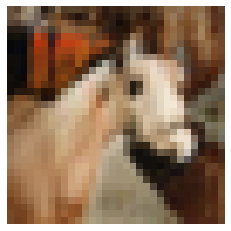}\!
\includegraphics[]{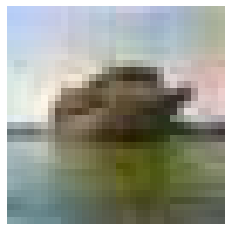}\!
\includegraphics[]{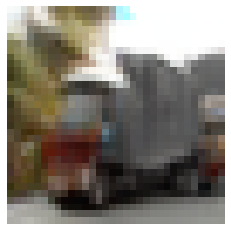}\!
\includegraphics[]{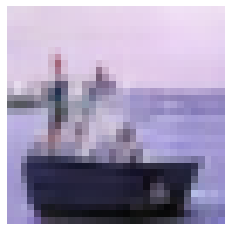}\!
\includegraphics[]{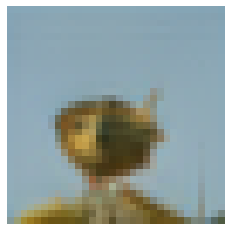}\!
\includegraphics[]{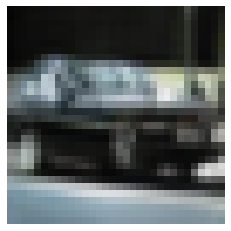}
}

\resizebox{\columnwidth}{!}{
\includegraphics[]{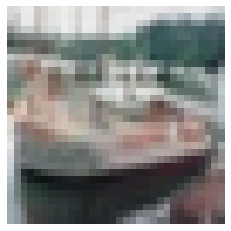}\!
\includegraphics[]{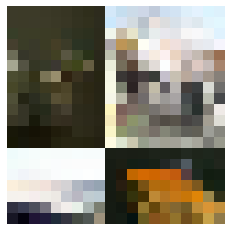}\!
\includegraphics[]{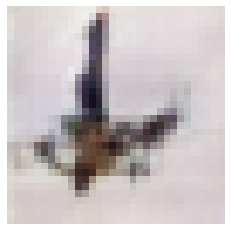}\!
\includegraphics[]{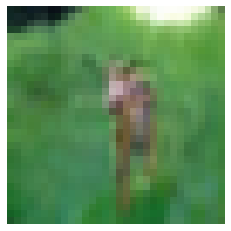}\!
\includegraphics[]{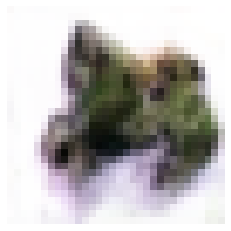}\!
\includegraphics[]{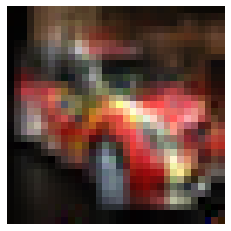}\!
\includegraphics[]{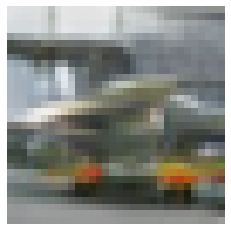}\!
\includegraphics[]{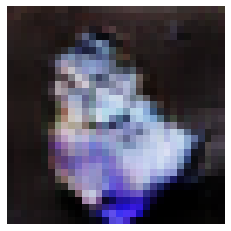}\!
\includegraphics[]{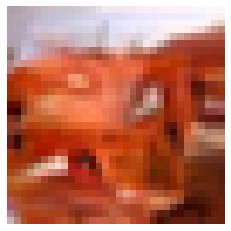}\!
\includegraphics[]{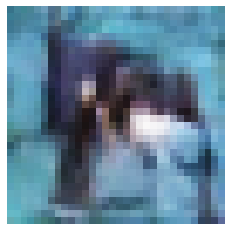}
}

\resizebox{\columnwidth}{!}{
\includegraphics[]{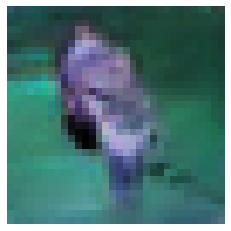}\!
\includegraphics[]{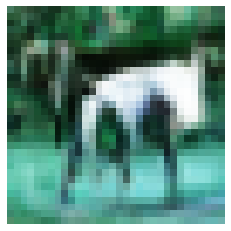}\!
\includegraphics[]{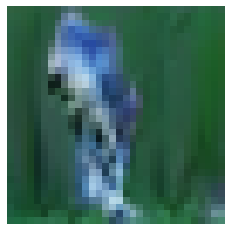}\!
\includegraphics[]{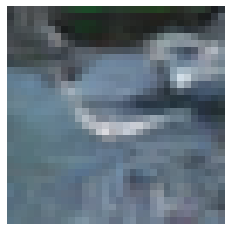}\!
\includegraphics[]{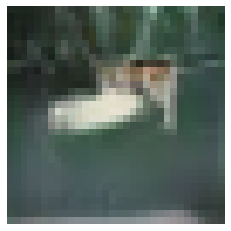}\!
\includegraphics[]{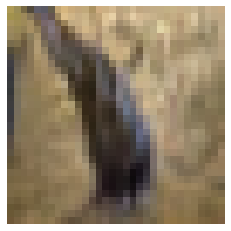}\!
\includegraphics[]{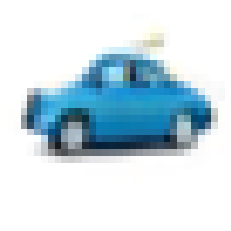}\!
\includegraphics[]{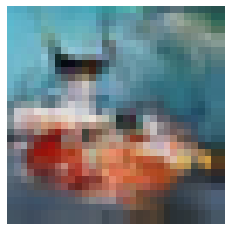}\!
\includegraphics[]{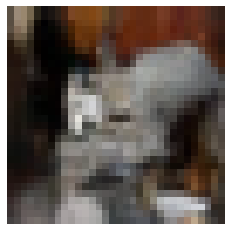}\!
\includegraphics[]{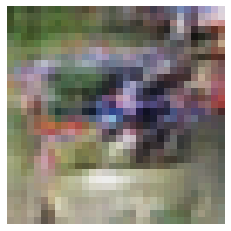}

}

\resizebox{\columnwidth}{!}{
\includegraphics[]{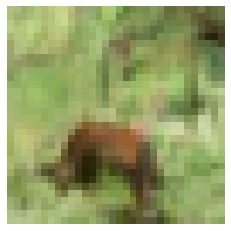}\!
\includegraphics[]{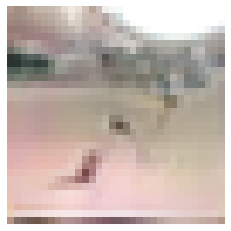}\!
\includegraphics[]{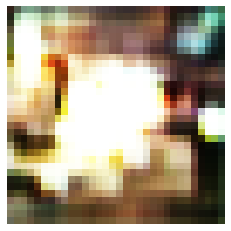}\!
\includegraphics[]{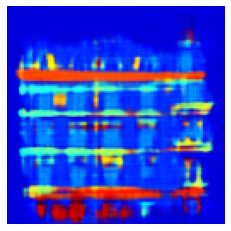}\!
\includegraphics[]{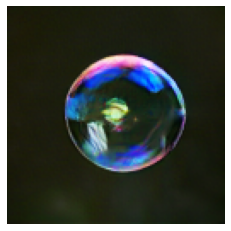}\!
\includegraphics[]{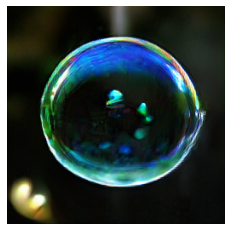}\!
\includegraphics[]{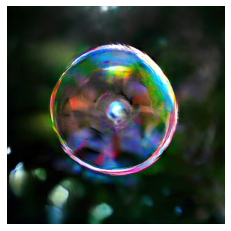}\!
\includegraphics[]{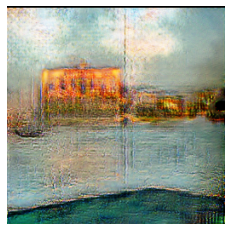}\!
\includegraphics[]{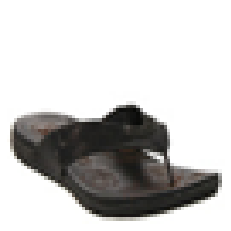}\!
\includegraphics[]{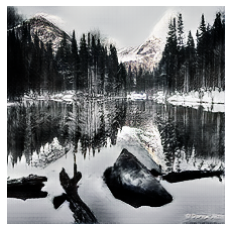}

}
\begin{center}
\resizebox{0.6\columnwidth}{!}{
\includegraphics[]{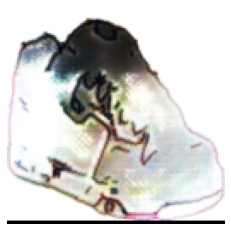}\!
\includegraphics[]{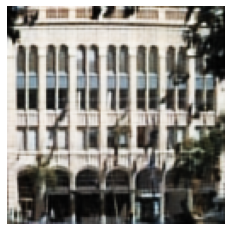}\!
\includegraphics[]{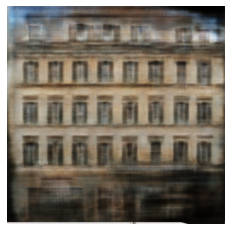}\!
\includegraphics[]{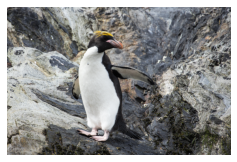}\!
\includegraphics[]{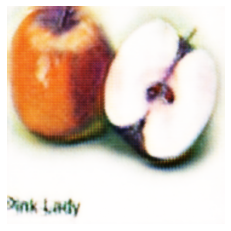}\!
\includegraphics[]{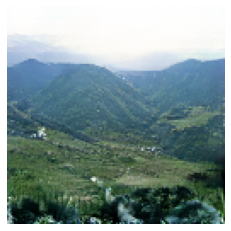}
}
\end{center}
\vspace{2mm}
\caption{\small Example images generated by all $116$ GMs in our collected dataset (one image per model).}
\label{gan_imagess}
\end{figure}

\begin{figure*}[t]
\begin{center}
\includegraphics[width=\linewidth]{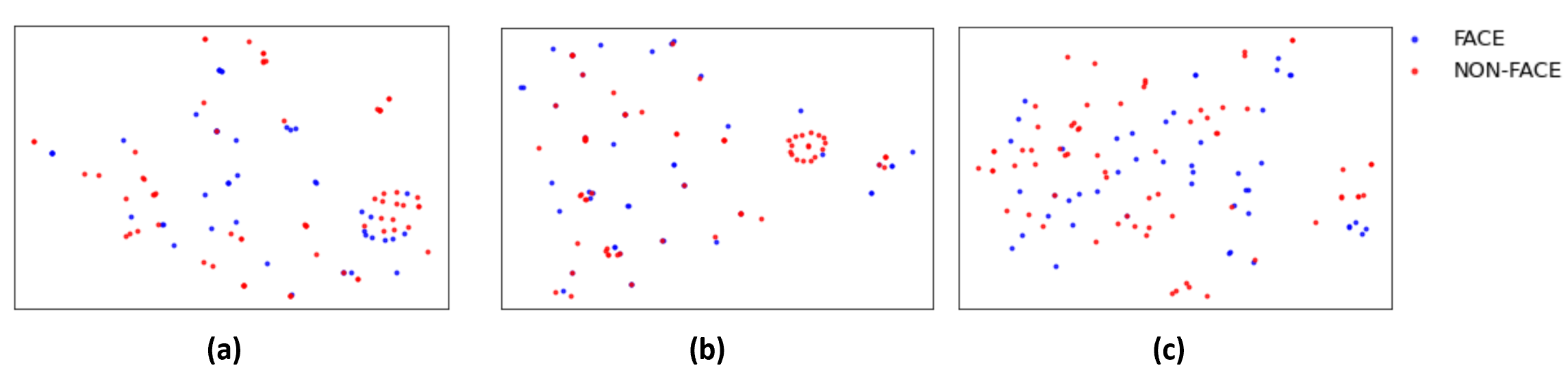}
\caption{\small t-SNE visualization for ground-truth vectors for (a) network architecture, (b) loss function and (c) network architecture and loss function combined. The ground-truth vectors are fairly distributed across the embedding space regardless of the face/non-face data.}
\label{fig:tsne}
\end{center}
\end{figure*}

\begin{figure*}[t!]
\begin{center}
\includegraphics[width=1\textwidth]{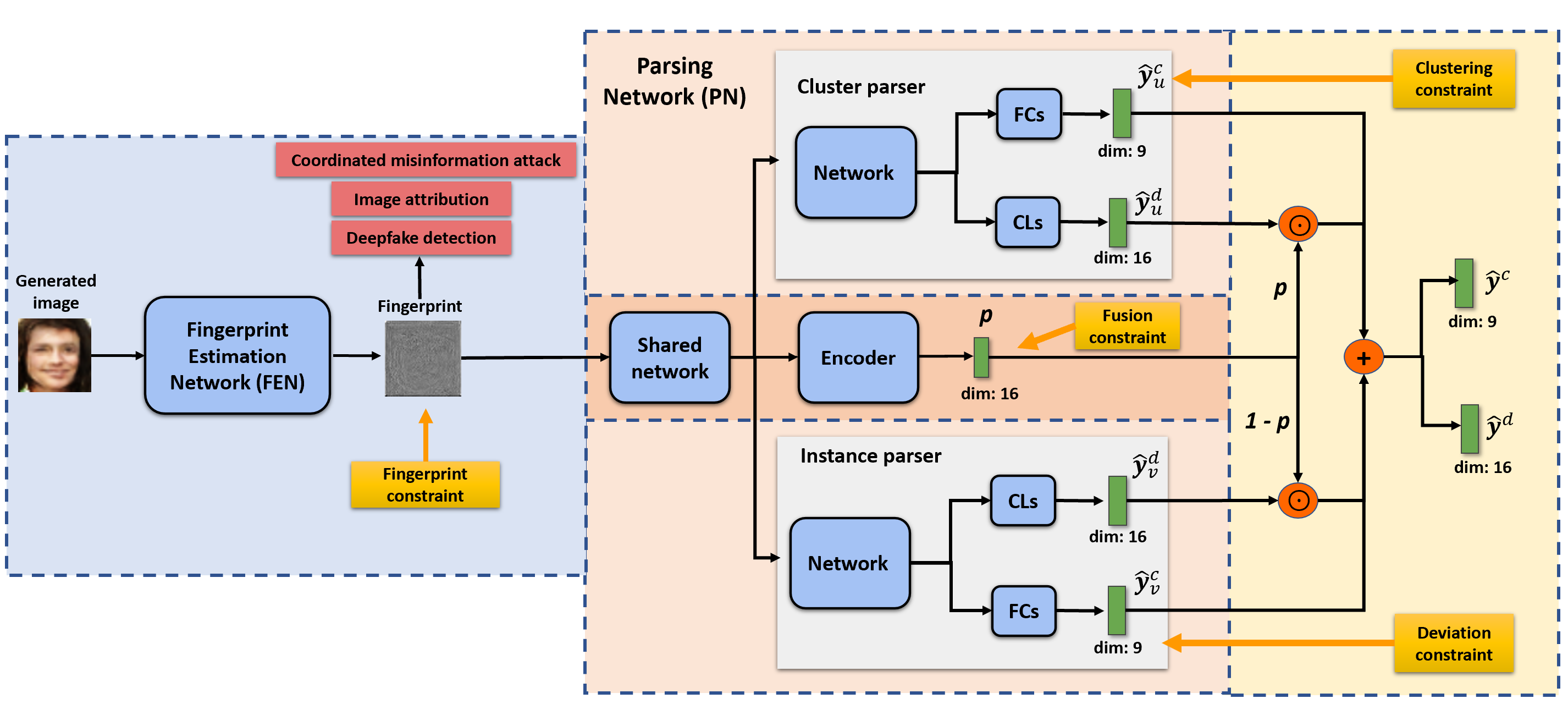}
\caption{\small Our framework includes two components: 1) the FEN is trained with four objectives for fingerprint estimation; and 2) the PN consists of a shared network, two parsers to estimate mean and deviation for each parameter, an encoder to estimate fusion parameter, fully connected layers (FCs) for continuous type parameters and separate classifiers (CLs) for discrete type parameters in network architecture and loss function prediction. 
Blue boxes denote trainable components; green boxes denote feature vectors; orange boxes denote loss functions; red boxes denote other tasks our framework can handle; black arrows denote data flow; orange arrows denote loss supervisions. Best viewed in color.}
\label{fig:overview}
\end{center}
\end{figure*}

%%
%% =======================================================
%%

\section{Proposed approach}
%\secvspace
In this section, we first introduce our collected dataset in Sec.~\ref{sec:data}. 
We then present the fingerprint estimation method in Sec.~\ref{sec:fen} and model parsing in Sec.~\ref{sec:pn}. 
Finally, we apply our estimated fingerprints to deepfake detection, image attribution, and detecting coordinated misinformation attacks, as described in Sec.~\ref{sec:other_app}.

\subsection{Data collection}
\label{sec:data}

We make the first attempt to study the model parsing problem.
Since data drives research, it is essential to collect a dataset for our new research problem.
Given the large number of GMs published in recent years~\cite{wang2019generative, jabbar2020survey}, we consider a few factors while deciding which GMs to be included in our dataset.
First of all, since it is desirable to study if model parsing is content-dependent, we hope to collect GMs with as diverse content as possible, such as the face, digits, and generic scenes.
Secondly, we give preference to GMs where either the authors have publicly released pre-trained models, generated images, or the training script. % to train models from scratch. 
Third, the network architecture of the GM should be clearly described in the respective paper.

%Importantly, our framework can be applied to general fake images in different domains. 
%We focus on faces, because face generation techniques have advanced significantly, resulting in the availability of many public GMs. 
To this end, we assemble a list of $116$ publicly available GMs, including ProGan~\cite{57}, StyleGAN~\cite{64}, and others.
A complete list is provided in the supplementary material. 
For each GM, we collect $1,000$ generated images. Therefore, our dataset $\mathcal{D}$ comprises of $116,000$ images.
We show example images in Figure~\ref{gan_imagess}. 
These GMs were trained on datasets with various contents, such as CelebA~\cite{liu2018large}, MNIST~\cite{deng2012mnist}, CIFAR10~\cite{krizhevsky2009learning}, ImageNet~\cite{deng2009imagenet}, facades~\cite{CycleGAN2017}, edges2shoes~\cite{CycleGAN2017}, and apple2oranges~\cite{CycleGAN2017}. The dataset is available \href{https://drive.google.com/file/d/1bAmC_9aMkWJB_scGvOOWvNeLa9FBoMUr/view?usp=sharing}{here}.
%These models are selected because either the authors have released their pre-trained models or there is a training script available publicly to train models from scratch. 

\begin{table*}[t]
\begin{center}
\small
\caption{\small Hyper-parameters representing the network architectures of GMs. (KEYS: cont. int.: continuous integer.)}
\scalebox{0.9}{
\label{tab:net}
\begin{tabular}{@{}l@{\hspace{2mm}}c@{\hspace{2mm}}c|@{\hspace{2mm}}l@{\hspace{2mm}}c@{\hspace{2mm}}c|@{\hspace{2mm}}l@{\hspace{2mm}}c@{\hspace{2mm}}c@{}}
\hline
Parameter & Type & Range & Parameter & Type & Range & Parameter & Type & Range \\ \hline\hline

$\#$ layers & cont. int. & [$5$, $95$] & $\#$ filter & cont. int. & [$0$, $8365$] & non-linearity type in blocks & multi-class & $0$, $1$, $2$, $3$ \\

$\#$ convolutional layers & cont. int. & [$0$, $92$] & $\#$ parameters & cont. int. & [$0.36M$, $267M$] &  non-linearity type in last layer & multi-class & $0$, $1$, $2$, $3$ \\ 

$\#$ fully connected layers & cont. int. & [$0$, $40$] & $\#$ blocks & cont. int. & [$0$, $16$] & up-sampling type & binary & $0$, $1$ \\

$\#$ pooling layers & cont. int. & [$0$, $4$] & $\#$ layers per block & cont. int. & [$0$, $9$]&skip connection & binary & $0$, $1$ \\

$\#$ normalization layers & cont. int. & [$0$, $57$] & normalization type & multi-class & $0,1,2,3$ & down-sampling & binary & $0$, $1$ \\
\hline \hline
\end{tabular}
}
\end{center}
\end{table*}
\begin{table}[t]
\begin{center}
\caption{\small Loss function types used by all GMs. We group the $10$ loss functions into three categories. We use the binary representation to indicate presence of each loss type in training the respective GM. }
\scalebox{1}{
\begin{tabular}{@{\hspace{1mm}}c|@{\hspace{5mm}}c@{\hspace{1mm}}}
\hline
Category & Loss  function\\ \hline \hline
\multirow{5}{*}{Pixel-level} & $L_1$ \\
 & $L_2$ \\
 & Mean squared error (MSE) \\
 & Maximum mean discrepancy (MMD) \\
 & Least squares (LS) \\\hline
\multirow{4}{*}{Discriminator} & Wasserstein loss for GAN (WGAN) \\
 & Kullback–Leibler (KL) divergence \\
 & Adversarial \\
 & Hinge\\ \hline
Classification & Cross-entropy (CE)\\ \hline \hline
\end{tabular}
}
\label{tab:loss_type}
\end{center}
\end{table}

We further document the model hyperparameters for each GM as reported in their papers. 
Specifically, we investigate two aspects: network architecture and training loss functions. 
We form a super-set of $15$ network architecture parameters (\eg, number of layers, normalization type) and $10$ different loss function types.
%For each GM, we further document hyperparameters representing model network architectures and the training loss functions, as reported in their papers. 
We obtain a large-scale fake image dataset $\mathbb{D} = \{{\bf{X}}_i, {\bf{y}}^n_i, {\bf{y}}^l_i\}_{i=1}^{N}$ where ${\bf{X}}_i$ is a fake image, ${\bf{y}}^n_i\in\mathbb{R}^{15}$ and ${\bf{y}}^l_i\in\mathbb{R}^{10}$ represent the ground-truth network architecture and loss functions, respectively. 
We also show the t{-}SNE distribution for both network architecture and loss functions in Figure \ref{fig:tsne} for different \VA{types} of models and datasets. 
We observe that the ground-truth vectors for both network architecture and loss function are evenly distributed across the space for both types of data: face and non-face.

\subsection{Fingerprint estimation}
\label{sec:fen}
%Fingerprint estimation relies the assumption that each GM leaves a unique pattern on the images it generated~\cite{12}; this pattern arises from network operators such as convolution, average pooling, \etc. The estimated fingerprints are discriminative in that can be used to identify the model~\cite{11}, and, in our case, parse model hyperparameters. 
We adopt a network structure similar to the DnCNN model used in~\cite{72}. As shown in Figure~\ref{fig:overview}, the input to FEN is a generated image ${\bf{X}}$, and the output is a fingerprint image ${\bf{F}}$ of the same size. Motivated \VA{by} prior works on physical fingerprint estimation~\cite{40,13,12,11,10}, we define the following four constraints to guide our estimated fingerprints to have the desirable properties. 

\minisection{Magnitude loss} Fingerprints can be considered as image noise patterns with small magnitudes. 
Similar assumptions were made by others when estimating spoof noise for spoofed face images~\cite{40} and sensor noise for genuine images~\cite{8}. 
The first constraint is thus proposed to regularize the fingerprint image to have a low magnitude with an $L_2$ loss:
\begin{equation}
J_m = ||{\bf{F}}||_2^2.\label{eq:Jm}
\end{equation}
    
\minisection{Spectrum loss} Previous work observed that fingerprints primarily lie in the middle and high-frequency bands of an image~\cite{13}. We thus propose to minimize the low-frequency content in a fingerprint image by applying a low pass filter to its frequency domain:
\begin{equation}
J_s = ||\mathcal{L}(\mathcal{F}({\bf{F}}),f)\label{eq:Js}||_2^2,
\end{equation}
where $\mathcal{F}$ is the Fourier transform, $\mathcal{L}$ is the low pass filter selecting the $f \times f$ region in the center of the $2$D Fourier spectrum and making everything else zero.
    
\minisection{Repetitive loss} Amin~\etal~\cite{40} noted that the noise characteristics of an image are repetitive and exist everywhere in its spatial domain. Such repetitive patterns will result in a large magnitude in the high-frequency band of the fingerprint. Therefore, we propose to maximize the high-frequency information to encourage this \VA{repetitive} pattern: 
\begin{equation}
J_r = -\text{max}\{\mathcal{H}(\mathcal{F}({\bf{F}}),f )\},\label{eq:Jr}
\end{equation}
where $\mathcal{H}$ is a high pass filter assigning the $f \times f$ region in the center of the $2$D Fourier spectrum to zero. 

\Paragraph{Energy loss.} Wang~\etal~\cite{12} showed that unique patterns exist in the Fourier spectrum of the image generated by CNN networks. These patterns have similar energy in the vertical and horizontal directions of the Fourier spectrum. Our final constraint is proposed to incorporate this observation: 
\begin{equation}
J_e=||\mathcal{F}({\bf{F}})-\mathcal{F}({\bf{F}})^T||_2^2,
\label{eq:Je}
\end{equation}
where $\mathcal{F}({\bf{F}})^T$ is the transpose of $\mathcal{F}({\bf{F}})$.

These constraints guide the training of our fingerprint estimation.  As shown in Figure~\ref{fig:overview}, the fingerprint constraint is given by:
\begin{equation}
J_f = \lambda_1{J_m}+\lambda_2{J_s}+\lambda_3{J_r}+\lambda_4{J_e},
\label{eqn:fen}
\end{equation}
where $\lambda_1$, $\lambda_2$, $\lambda_3$, $\lambda_4$ are the loss weights for each term. 

\subsection{Model parsing}
\label{sec:pn}
The estimated fingerprint is expected to capture unique patterns generated from a GM. Prior works adopted fingerprints for deepfake detection~\cite{2,3} and image attribution~\cite{11}. However, we go beyond those efforts by parsing the hyperparameters of GMs. 
As shown in Figure~\ref{fig:overview}, we perform prediction using two parsers, namely, cluster parser and instance  parser. We combine both outputs for network architecture and loss function prediction. We will now discuss the ground truth calculation and our framework in detail.  

\subsubsection{Ground truth hyperparamters}

\minisection{Network architecture} In this work, we do not aim to recover the network parameters.
The reason is that a typical deep network has millions of network parameters, which reside in a very high dimensional space and is thus hard to predict.
%We do not aim to recover the exact model structure. 
Instead, we propose to infer the hyperparameters that define the network architecture, which \VA{are} much fewer than the network parameters. 
Motivated by prior works in neural architecture search~\cite{68,69,70}, we form a set of  $15$ network architecture parameters covering various aspects of architectures. 
%In total, we study $15$ hyperparameters for network architecture representations. 
As shown in Tab.~\ref{tab:net}, these parameters fall into different data types and have different ranges. We further split the network architecture parameters ${\bf{y}}^n$ into two parts: ${\bf{y}}^{n_c}\in\mathbb{R}^{9}$ for continuous data type and ${\bf{y}}^{n_d}\in\mathbb{R}^{6}$ for discrete data type. 

\minisection{Loss function} In addition to the network architectures, the learned network parameters of trained GM can also impact the fingerprints left on the generated images. These network parameters are determined mainly by the training data and the loss functions used to train these models. We, therefore, explore the possibility of also predicting the training loss functions from the estimated fingerprints. 
The $116$ GMs were trained with $10$ types of loss functions as shown in Tab. \ref{tab:loss_type}. 
For each model, we compose a ground-truth vector ${\bf{y}}^l\in\mathbb{R}^{10}$, where each element is a binary value indicating whether the corresponding loss is used or not in training this model.

Our framework parses two types of hyperparameters: continuous and discrete. The former includes the continuous network architecture parameters. The latter includes discrete network architecture parameters and loss function parameters. For clarity, we group these parameters into continuous and discrete types in the remaining of this section to describe the model parsing objectives. We use ${\bf{y}}^{c}$ and ${\bf{y}}^{d}$ to denote continuous and discrete parameters respectively.
%The training for both data types is different, but it is similar for parameters among each data type. Therefore, for simplicity purposes, we denote the network architecture parameters for continuous data type ${\bf{y}}^{n_c}$ as ${\bf{y}}^{c}$. For discrete parameters, we denote the vectors, ${\bf{y}}^{n_d}$ and ${\bf{y}}^l$ as ${\bf{y}}^{d}$.

\begin{figure}[t!]
\begin{center}
\includegraphics[width=1\columnwidth]{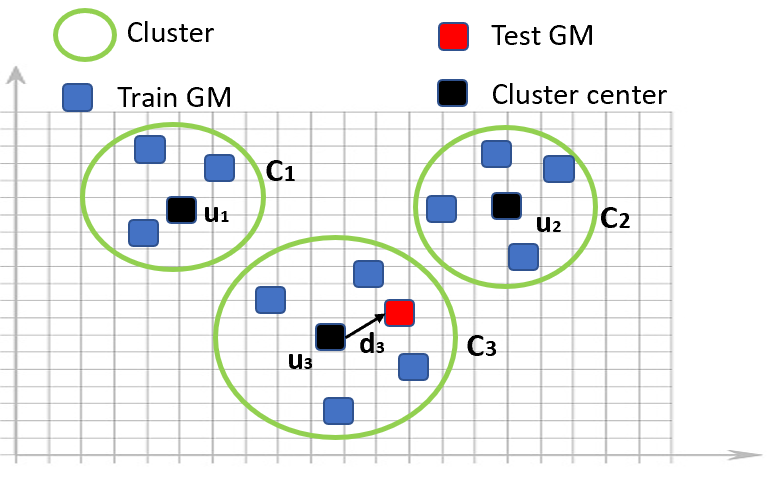}
\caption{\small The idea of grouping various GMs into different clusters. For the test GM, we estimate its cluster mean and the deviation from that mean to predict network architecture and loss function type. }
\label{fig:cluster}
\end{center}
\end{figure}

\subsubsection{Cluster parser prediction}
We have observed that directly \VA{estimating} the hyperparameters independently for each GM \VA{yields} inferior results. 
In fact, some of the GMs in our dataset have similar network architectures and/or loss functions. 
It is intuitive to leverage the similarities among different GMs for better hyperparameter estimation. 
To do this, we perform k-means clustering to group all GMs into different clusters, as shown in Figure~\ref{fig:cluster}. 
Then we propose to perform cluster-level coarse prediction and GM-level fine prediction, which are subsequently combined to obtain the final prediction results.

%Some of the GMs in our dataset have either similar network architecture or similar types of loss functions. As we show in our experiments, estimating the network architecture and loss functions for individual GMs by directly performing regression or using classifiers might be too naive as there is no usage of the similarity between the ground truth of various GMs. To address this issue, we perform the grouping of GMs to assign them into different clusters as shown in Fig. \ref{fig:cluster}. Every cluster would represent GMs that are similar to each other. Cluster-level prediction is intuitively the coarse-level mean  prediction for every GM. The main question is how this grouping should be performed for all the GMs in the training set. 

As we aim to estimate the parameters for network architecture and loss function, it is intuitive to combine them to perform grouping. Thus, we concatenate the ground truth network architecture parameters ${\bf{y}}^n$ and loss function parameters ${\bf{y}}^l$, denoted as ${\bf{y}}^{nl}$. 
We use these ground truth vectors to perform k-means clustering to find the optimal \textit{k}-clusters in the dataset ${\bf\mathcal{D}}=\{\vect{C}_1, \vect{C}_2,...\vect{C}_k\}$. Our clustering objective can be written as:
\begin{equation}
    \argmin_{\bf\mathcal{D}}\sum_{i=1}^{k}\sum_{{\bf y}^{nl}_j\in \vect{C}_i} ||{\bf y}^{nl}_j-{\bf \mu}_i||^2,
\end{equation}
where ${\bf \mu}_i$ is the mean of the ground truth of the GMs in $\vect{C}_i$. 

Our dataset comprises different kinds of GMs, namely GANs, VAEs, AAs, ARs, and NFs. We perform clustering after separating the training data into different kinds of GMs. This is done to ensure that each cluster would belong to one particular kind of GM. Next, we select the value of \textit{k} \ie, the number of clusters, using the elbow method adopted by previous works~\cite{bholowalia2014ebk, kodinariya2013review}.
After determining the clusters comprising of similar GMs, we estimate the ground truth ${\bf{y}}_u$ to represent the respective cluster. We estimate this cluster ground truth using different ways for continuous and discrete parameters. For the former, we take the average of each parameter using the ground truth for all GMs in the respective cluster. For the latter, we perform majority voting for every parameter to find the most common class across all GMs in the cluster. 

We use different loss functions to perform cluster-level prediction. For continuous parameters, we perform regression for parameter estimation. As these parameters are in different ranges, we further perform a min-max normalization to bring all parameters to the range of [$0$, $1$].
An $L_2$ loss is used to estimate the prediction error:
\begin{equation}
J^{c}_u = ||{\bf{\hat{y}}}_u^{c} - {\bf{y}}_u^{c}||_2^2,
\label{eq:J_u_nc}
\end{equation}
where ${\bf{\hat{y}}}_u^{c}$ is the cluster mean prediction and ${\bf{y}}_u^{c}$ is the normalized ground-truth cluster mean.
%cluster ground-truth parameters of continuous data type.  

For discrete parameters, the prediction is made via individual classifiers. Specifically, we train $M=16$ classifiers ($6$ for network architecture and $10$ for loss function parameters), one for each discrete parameter. The loss term for discrete parameters cluster-prediction is defined as:
\begin{equation}
J^{d}_u = - \sum_{m=1}^{M}{\mathrm{sum}({\bf{y}}_{u_m}^{d}\odot \mathrm{log}(\mathcal{S}({\bf{\hat{y}}}_{u_m}^{d})))},
\label{eq:J_u_d}
\end{equation}
where ${\bf{y}}_{u_m}^{d}$ is the ground-truth one-hot vector for the respective class in the \VA{$m$-th} discrete type parameter, ${\bf{\hat{y}}}_{u_m}^{d}$ are the class logits, $\mathcal{S}$ is the Softmax function that maps the class logits into the range of $[0,1]$,  $\odot$ is the element-wise multiplication, and $\mathrm{sum}()$ computes the summation of a vector's elements.

As shown in Figure~\ref{fig:overview}, the clustering constraint is given by:
\begin{equation}
J_{u} = \gamma_1{J^{c}_u}+\gamma_2{J^{d}_u},
\label{eq:Ju}
\end{equation}
where $\gamma_1$ and $\gamma_2$ are the loss weights for each term.

\subsubsection{Instance  parser prediction}
%We use a cluster parser to perform prediction at the coarse-level, as explained before. 
The cluster parser performs coarse-level prediction.
To obtain a more fine-level prediction, we use an instance  parser to estimate a GM-level prediction, which ignores any similarity among GMs. This parser aims to predict the deviation of every parameter from the coarse-level prediction. The ground truth deviation vector ${\bf{y}}_v$ can be estimated in different ways for two types of parameters. For continuous type parameters, the deviation can be the difference between the ground truth of the GM and the ground truth of the cluster the GM was assigned. 
However, in the case of discrete parameters, the actual ground truth class for the parameters can act as the deviation from the most common class estimated in cluster ground truth. We use different loss functions to perform deviation-level prediction. Specifically, we use an $L_2$ loss to estimate the prediction error for continuous parameters:
\begin{equation}
J^{c}_v = ||{\bf{\hat{y}}}_v^{c} - {\bf{y}}_v^{c}||_2^2,
\label{eq:J_v_c}
\end{equation}
where ${\bf{\hat{y}}}_v^{c}$ is the deviation prediction and ${\bf{y}}_v^{c}$ is the deviation ground-truth of continuous data type.

We have noticed the class distribution for some discrete parameters is imbalanced. Therefore, we apply the weighted cross-entropy loss for every parameter to handle this challenge. We train $M=16$ classifiers, one for each of the discrete parameters. For the $m$-th classifier with $N_m$ classes ($N_m=2$ or $4$ in our case), we calculate a loss weight for each class as $w_{m}^i = \frac{N}{N_{m}^i}$ where $N_m^i$ is the number of training examples for the $i$th class of $m$-th classifier, and $N$ is the number of total training examples. As a result, the class with more examples is down-weighted, and the class with fewer examples is up-weighted to overcome the imbalance issue, which will be empirically demonstrated in Figure.~\ref{fig:con_mat}. The loss term for discrete parameters deviation-prediction is defined as:
\begin{equation}
J^{d}_v = - \sum_{m=1}^{M}{\mathrm{sum}({\bf{w}}_m\odot {\bf{y}}_{v_m}^{d}\odot \mathrm{log}(\mathcal{S}({\bf{\hat{y}}}_{v_m}^{d})))},
\label{eq:J_v_d}
\end{equation}
where ${\bf{y}}_{v_m}^{d}$ is the ground-truth one-hot deviation vector for the $m$-th classifier, ${{\bf{{w}}}_m}$ is a weight vector for all classes in the $m$-th classifier and ${\bf{\hat{y}}}_{v_m}^{d}$ are the class logits.

As shown in Figure \ref{fig:overview}, the deviation constraint is given by:
\begin{equation}
J_{v} = \gamma_3{J^{c}_v}+\gamma_4{J^{d}_v}.
\label{eq:Jv}
\end{equation}
where $\gamma_3$ and $\gamma_4$ are the loss weights for each term.

\subsubsection{Combining predictions}
We use a cluster parser to perform a coarse-level mean prediction and an instance  parser to predict a deviation prediction for each GM. The final prediction of our framework, \ie, the prediction at the fine-level is the combination of the outputs of these two parsers. 
%However, as we have two types of parameters, combining the output of both parsers would be different. 
For continuous parameters, we perform the element-wise addition of the coarse-level mean and deviation prediction: 
\begin{equation}
{\bf{\hat{y}}}^{c}={\bf{\hat{y}}}_u^{c}+{\bf{\hat{y}}}_v^{c},
\label{eq:y_net_c_final}
\end{equation}

For discrete parameters, we have observed that element-wise addition of the logits for every classifier in both parsers didn't perform well. Therefore, to integrate the outputs, we train an encoder network to predict a fusion parameter $\hat{p^{d}} \in [0,1]$ for each classifier. For any parameter, the value of the fusion parameter is $1$ if the cluster class is the same as the GM class, encouraging the parsing network to give importance to the cluster parser output. The value of the fusion parameter is $0$ if the GM class is different from the cluster class. Therefore, for $m$-th classifier, the training of the model is supervised by the ground truth $p^{d}_m$ as defined below:
\begin{equation}
p^{d}_m =
    \begin{cases} 
      1, & {\bf{y}}_{u_m}^{d}={\bf{y}}_{v_m}^{d}\\
      0, & {\bf{y}}_{u_m}^{d}\neq {\bf{y}}_{v_m}^{d}.\\
   \end{cases}
\label{eq:p_d}
\end{equation}

To train our encoder, we use the ground truth fusion parameter ${\bf p}^{d}$ which is the concatenation for all parameters. The training is done via cross-entropy loss as shown below:
\begin{equation}
%J_{p} = - \sum_{m=1}^{M}{\mathrm{sum}(p^{d}_m\odot \mathrm{log}(\mathcal{S}(\hat{p}^{d}_m)))}.
J_{p} = - \sum_{m=1}^{M}{(p^{d}_m\mathrm{log}(\mathcal{G}(\hat{p}^{d}_m)) + (1-p^{d}_m)\mathrm{log}(1-\mathcal{G}(\hat{p}^{d}_m)))}.
\label{eq:Jp}
\end{equation}
where $\mathcal{G}$ is the Sigmoid function that maps the class logits into the range of $[0,1]$. 

As shown in Figure~\ref{fig:overview} for discrete parameters, the final prediction is given by:
\begin{equation}
{\bf{\hat{y}}}^{d}={\bf{\hat{p}}}^{d}\odot{\bf{\hat{y}}}_u^{d}+({\bf{1}} -{\bf{\hat{p}}}^{d})\odot {\bf{\hat{y}}}_v^{d}.
\label{eq:y_d_final}
\end{equation}

The overall loss function for model parsing is given by:
\begin{equation}
J = J_f+J_u+J_v+\gamma_5J_p.
\label{eq:J_final}
\end{equation}
where $\gamma_5$ is the loss weight for fusion constraint.
Our framework is trained end-to-end with fingerprint estimation (Eqn.~\ref{eqn:fen}) and model parsing (Eqn.~\ref{eq:J_final}).

\subsection{Other applications}
\label{sec:other_app}

%We show the application of our framework in various classification problems. 
%Our model parsing prediction can be used to identify a coordinated misinformation attack. 
In addition to model parsing, our fingerprint estimation can be easily leveraged for other applications such as detecting coordinated misinformation attacks, deepfake detection and image attribution.

\minisection{Coordinated misinformation attack}
In coordinated misinformation attacks, the attackers often use the same model to generate multiple fake images.
One way to detect such attacks is to classify whether two fake images are generated from the same GM, despite that this GM might be unseen to the classifier. 
This task is not straightforward to perform by prior works.
However, given the ability of our model parsing, this is the ideal task that we can contribute.
%The primary use of parsing GM's hyperparameters can be on estimating whether two images came from the same GM or not. Estimating this information is desirable because then we would be able to identify whether there has been a coordinated misinformation attack or not. Therefore,
To perform this binary classification task, we use the parsed network architecture and loss function parameters to calculate the similarity score between two test images. We calculate the cosine similarity for continuous type parameters and fraction of the number of parameters having same class for discrete type. Both cosine similarity and fraction of parameters are averaged to get the similarity score. 
%These images can come either from the same GM or two different GMs.
%Comparing the cosine similarity with a threshold will lead to the binary classification decision.
Comparing the cosine similarity with a threshold will lead to the binary classification decision of whether two images come from the same GM or not. 
%Further, we use these cosine similarities to perform binary classification.

\minisection{Deepfake detection} 
We consider the binary classification of an image as either genuine or fake. 
We add a shallow network on the generated fingerprint to predict the probabilities of being genuine or fake. 
The shallow network consists of five convolution layers and two fully connected layers. 
Both genuine and fake face images are used for training. Both FEN and the shallow network are trained end-to-end with the proposed fingerprint constraints (Eqn.~\ref{eqn:fen}) and a cross-entropy loss for genuine \vs fake classification. 
Note that the fingerprint constraints (Eqn.~\ref{eqn:fen}) are not applied to the genuine input face images.

\minisection{Image attribution} We aim to learn a mapping from a given image to the model that generated it if it is fake or classified as genuine otherwise. 
%We aim to learn a mapping from a given fake image to the model that generated it. 
All models are known during training. We solve image attribution as a closed-set classification problem. Similar to deepfake detection, we add a shallow network on the generated fingerprint for model classification with the cross-entropy loss. 
The shallow network consists of two convolutional layers and two fully connected layers. 

%%
%% =======================================================
%%prediction

\section{Experiments}

\subsection{Settings}
\label{sec:exp:setting}
\minisection{Dataset} As described in Sec.~\ref{sec:data}, we have collected a fake image dataset \VA{consisting} of $116K$ images from $116$ GMs ($1K$ images per model) for model parsing experiments. 
These models can be split into two parts: $47$ face models and $69$ non-face models. 
%We focus our evaluation mainly on face models. 
Instead of performing one split of training and testing sets, we carefully construct four different splits with a focus on curating well-represented test sets.
%testing sets. For each set, we select representative models in order to evaluate the generalization ability of our algorithm. 
Specifically, each testing set includes six GANs, two VAEs, two ARs, one AA and one NF model.
We perform cross-validation to train on $104$ models and evaluate on the remaining $12$ models in testing sets. The performance is averaged across four testing sets.

%\VA{\noindent{\bf Dataset.} As described in Sec.~\ref{sec:data}, we collected a fake face dataset consists of $100K$ images from $100$ GMs ($1K$ images per model). To evaluate our model's generalization ability, we create six different testing sets, each containing images from the GMs unseen in training. Each set consists of six different GMs. We conduct experiments based on face models and non-face models in testing set. More details of the all the testing sets can be found in supplementary material.}
%We conduct leave-one-model-out experiments to evaluate our generalization ability to parse {\it unseen} GMs.

For deepfake detection experiments, we conduct experiments on the recently released Celeb-DF dataset~\cite{21}, consisting of $590$ real and $5,639$ fake videos. 
For image attribution experiments, a source database with genuine images needs to be selected, from which the fake images can be generated by various GAN models.
We select two source datasets: CelebA~\cite{21} and LSUN~\cite{73}, for two experiments. 
From each source dataset, we construct a training set of $100K$ genuine and $100K$ fake face images produced by each of the same four GAN models used in Yu~\etal~\cite{11}, and a testing set with $10K$ genuine and $10K$ fake images per model. 
%We do this experiment for two different datasets: CelebA~\cite{21} and LSUN~\cite{73}.

%For deepfake detection, we conduct experiments on the recently released Celeb-df dataset~\cite{21}, consisting of $180K$ genuine and $140K$ fake face images. For image attribution, we use a training set of $100K$ genuine and $100K$ fake face images produced by each of the same four GAN models used by Yu~\etal~\cite{11}, and a testing set with $10K$ genuine and $10K$ fake images per model. We do this experiment for two different datasets: CelebA \cite{21} and LSUN \cite{73} %For image attribution, we use the dataset released by Yu~\etal~\cite{11}, which offers a training set of $100K$ genuine and $100K$ fake face images produced by four GAN models, and a testing set with $10K$ genuine and $10K$ fake images per model. 

\minisection{Implementation details} Our framework is trained end-to-end with the loss functions of Eqn.~\ref{eqn:fen} and Eqn.~\ref{eq:J_final}. The loss weights are set to make the magnitudes of all loss terms comparable: $\lambda_1=0.05$, $\lambda_2 = 0.001$, $\lambda_3 = 0.1$, $\lambda_4 = 1$, $\gamma_1 = 5$, $\gamma_2 = 5$, $\gamma_3 = 5$, $\gamma_4 = 5$, $\gamma_5 = 5$, $\gamma_6 = 5$, $\gamma_7 = 1$, $\gamma_8 = 1$. The value of $f$ for spectrum loss and repetitive loss in the fingerprint estimation is set to $50$. For each of the four test sets, we calculate the number of clusters \textit{k} using the elbow method. We divide the data into different GM types and perform k-means clustering separately for each type. 
According to the sets defined in the supplementary, we obtain the value of \textit{k} as $11$, $11$, $15$, and $13$. We use Adam optimizer with a learning rate of $0.0001$. Our framework is trained with a batch size of $32$ for $10$ epochs. 
All the experiments are conducted using NVIDIA Tesla K$80$ GPUs. 

\minisection{Evaluation metrics} For continuous type parameters, we report the $L_1$ error for the regression estimation of continuous type parameters. We also report the p-value of t-test, correlation coefficient, coefficient of determination~\cite{srivastava1995coefficient} and slope of the RANSAC regression line~\cite{fischler1981random} to show the effectiveness of regression in our approach. For discrete type parameters, as there is imbalance in the dataset for different parameters, we compute the F1 score \cite{forman2010apples, jeni2013facing} for classification performance. We also report classification accuracy for discrete-type parameters. For all cross-validation experiments, we report the averaged results across all images and all GMs.

\subsection{Model parsing results}
As we are the first to attempt GM parsing, there are no prior works for comparison. To provide a baseline, we, therefore, draw \VA{an} analogy with the image attribution task, where each model is represented as a one-hot vector and different models have equal inter-model distances in the high-dimensional space defined by these one-hot vectors. In model parsing, we represent each model as a $25$-D vector consisting of network architectures ($15$-D) and training loss functions ($10$-D). Thus, these models are not of equal distance in the $25$-D space. 

Based on the aforementioned observation, we define a baseline, referred to here as {\it random ground-truth}. Specifically, for each parameter, we shuffle the values/classes across all $116$ GMs to ensure that the assigned ground-truth is different from the actual ground-truth but also preserves the actual distribution of each parameter\VA{, which means that the random ground-truth baseline is not based on random chance.}
These random ground-truth vectors have the same properties as our ground-truth vectors in terms of non-equal distances. 
But the shuffled ground truths are meaningless and are not corresponding to their true model hyperparameters. We train and test our proposed approach on this randomly shuffled ground-truth.
Due to the random nature of this baseline, we perform three random shuffling and then report the average performance.
\VA{We also evaluate a baseline of always predicting the mean for continuous hyperparameters, and always predicting the mode for discrete hyperparameters across the four sets. These mean/mode values of the hyperparameters are both measures of central tendency to represent the data, and they might result in a good enough performance for model parsing.}   
%Due to the imbalance in our data for some parameters, we observe that even after shuffling, some models may end up getting their correct value/class for some parameters. Therefore, to report the results, we conduct the three experiments with different shuffling every time and report the mean performance. 
%The substantially superior performance of our method over this baseline demonstrates that, there are indeed non-negligible {\it correlation} between the generated images and their ground truth network architectures, which is the foundation of why model parsing of GMs can be a valid research task.

%the representation ability of our ground-truth model hyperparameters.
%Based on the above observation, we define a simple baseline, referred to as {\it random ground-truth}. Specifically, we randomly generate a $27$-D vector, which follows similar distributions as the ground-truth vectors for each model. These random ground-truth vectors have the same properties as our ground-truth vectors in terms of non-equal distances. But the elements of these random vectors are meaningless. Comparison with this baseline demonstrates the representation ability of our ground-truth model hyperparameters. 

To validate the effects of our proposed fingerprint estimation constraints, we conduct an ablation study and train our framework end-to-end with only the model parsing objective in Eqn.~\ref{eq:J_final}. This results in the {\it no fingerprint} baseline. Finally, to show the importance of our clustering and deviation parser, we estimate the network architecture and loss functions using just one parser, which estimates the parameters directly instead of a mean and deviation. 
We refer \VA{to} this as {\it using one parser} baseline.

\begin{figure}[t]
\begin{center}
\includegraphics[width=\linewidth]{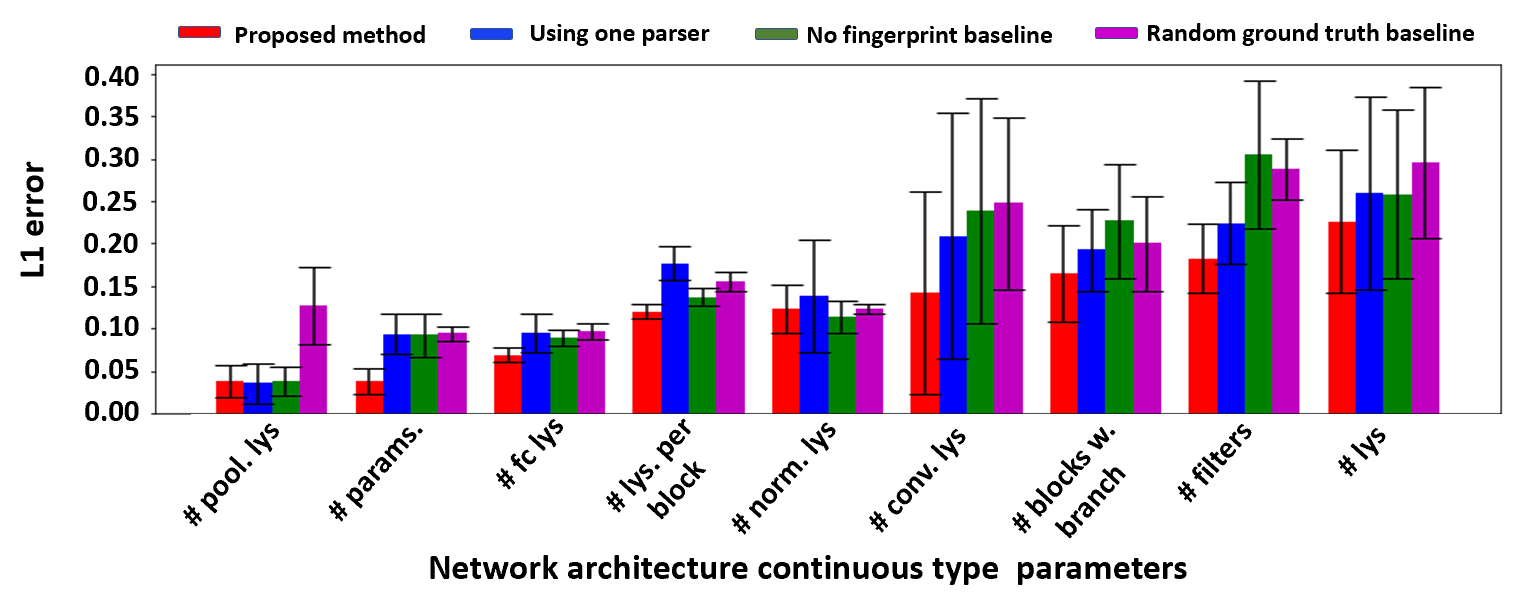}
\includegraphics[width=\linewidth]{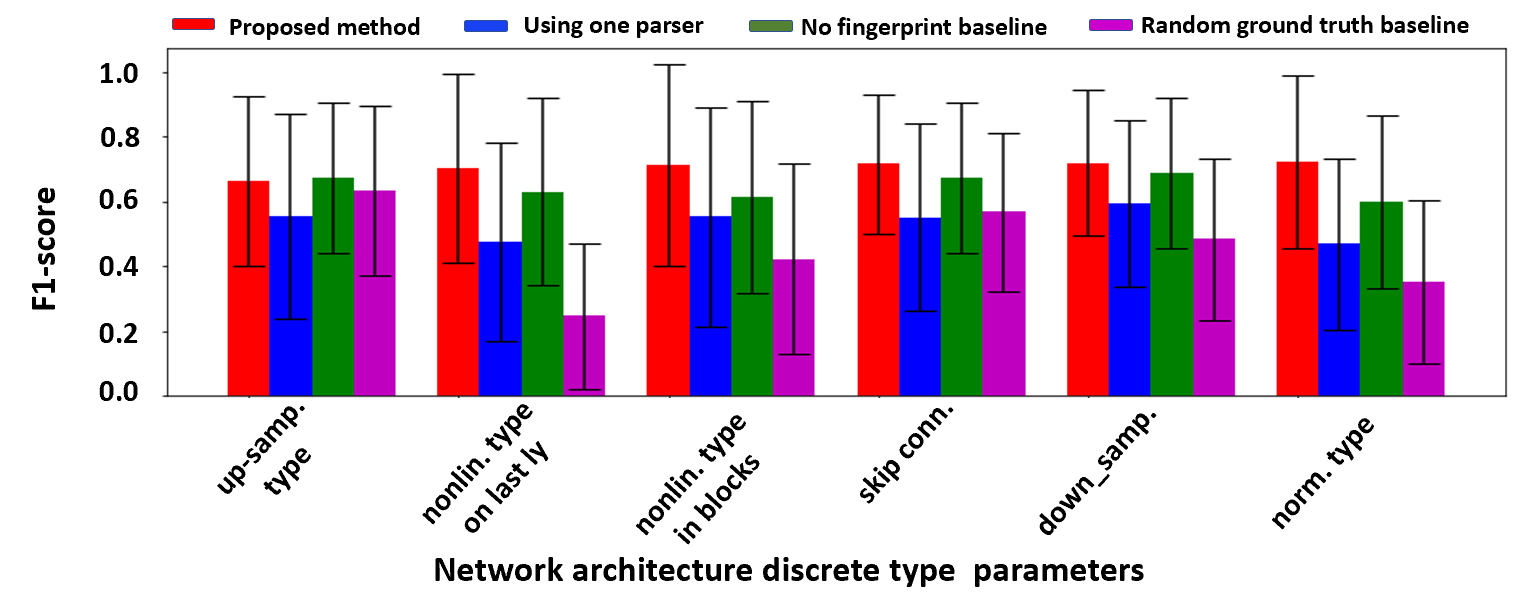}
\caption{\small $L_1$ error and F1 score for continuous and discrete parameters respectively of network architecture averaged across all images of all models in the $4$ test sets. Not only we have better average performance, but also our standard deviations are smaller.}
\label{fig:err_bar}
\end{center}
\end{figure}

\begin{table*}[t]
\begin{center}
\caption{\small Performance of network architecture prediction. We use $L_1$ error, p-value, correlation coefficient, coefficient of determination and slope of RANSAC regression line for continuous type parameters. For discrete parameters, we use F1 score and classification accuracy. \VA{We also show the standard deviation over all the test samples for $L_1$ error. The first value is the standard deviation across sets, while the second one is across the samples. The p-value would be estimated for every ours-baseline pair.} Our method performs better for both types of variables compared to the three baselines. [KEYS: corr.: correlation, coef.: coefficient, det.: determination]}
\scalebox{0.9}{
\begin{tabular}{l|c|c|c|c|c|c|c}
\hline
\multirow{2}{*}{Method} & \multicolumn{5}{c|}{Continuous type} & \multicolumn{2}{c}{Discrete type}\\\cline{2-8}
 & $L_1$ error $\bf \downarrow$ & \VA{P-value $\bf \downarrow$}& Corr. coef. $\bf \uparrow$& Coef. of det. $\bf \uparrow$& Slope $\bf \uparrow$& F1 score $\bf \uparrow$ & Accuracy $\bf \uparrow$\\ \hline \hline
Random ground-truth & $0.184 \pm 0.019\VA{/0.036}$ &\VA{$0.006\pm0.001$} & $0.261\pm0.181$ &$0.315\pm0.095$  &$0.592\pm0.041$ & $0.529 \pm 0.078$ & $0.575 \pm 0.097$\\
\VA{Mean/mode} & \VA{$0.164\pm 0.011/0.016$} & \VA{$0.035\pm0.005$} & \VA{$0.326\pm 0.112$} & \VA{$0.467\pm0.015$} & \VA{$0.632\pm 0.024$} & \VA{$0.612\pm 0.048$} & \VA{$0.604\pm 0.046$}\\
No fingerprint & $0.170 \pm 0.035\VA{/0.012}$ & \VA{$0.017\pm0.004$} & $0.738\pm0.014$ &$0.605\pm0.152$ & $0.892\pm0.021$& $0.700 \pm 0.032$ & $0.663\pm0.104$\\
Using one parser & $0.161 \pm 0.028\VA{/0.035}$ & $\VA{0.032\pm0.002}$& $0.226\pm0.030$ &$0.512\pm0.116$ & $-0.529\pm0.075$& $0.607 \pm 0.034$ &$0.593\pm0.104$\\
Ours &$ \bf 0.149 \pm  0.019\VA{/0.014}$  & \VA{-} & $\bf 0.744\pm0.098$ &$\bf 0.612\pm0.161$ &$\bf 0.921\pm0.021$ & $\bf 0.718 \pm 0.036$ &$\bf 0.706\pm0.040$\\ \hline \hline
\end{tabular}}
\label{tab:res_net}
\end{center}
\end{table*}

\begin{figure}[t]
\begin{center}
\includegraphics[width=\linewidth]{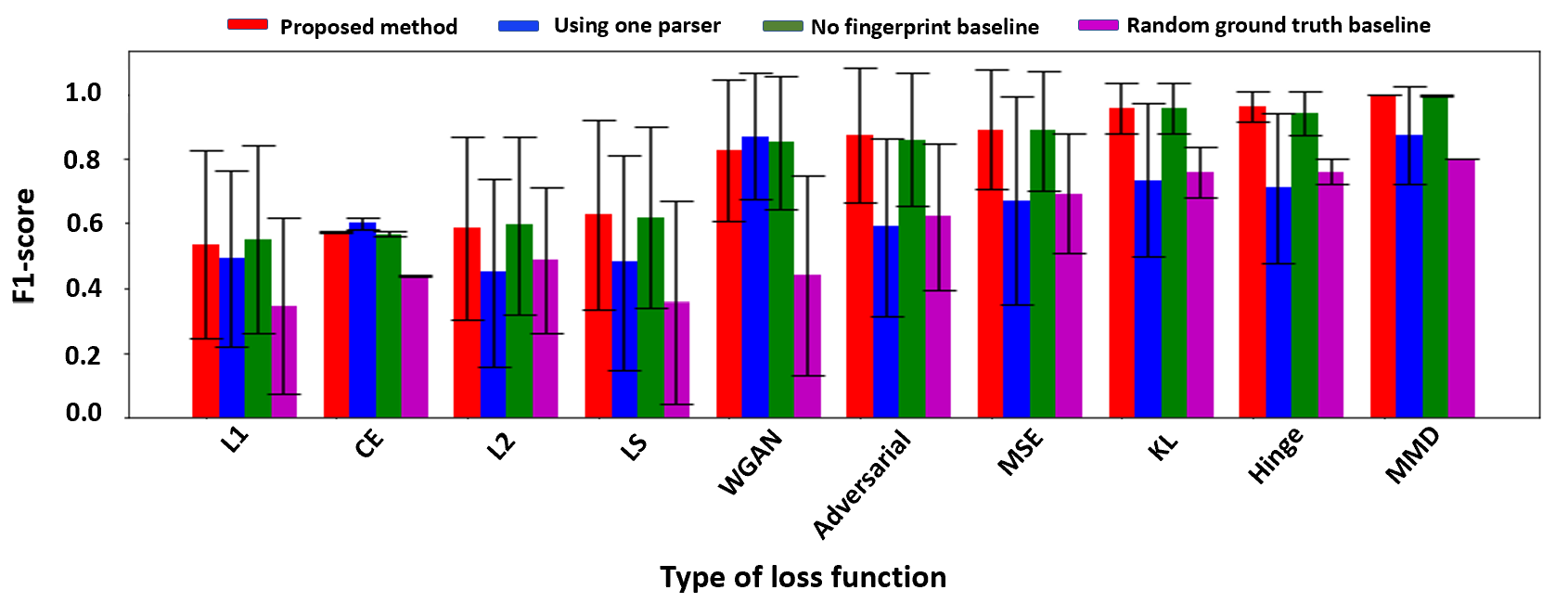}
\caption{\small F1 score for each loss function type at coarse and fine levels averaged across all images of all models in the $4$ test sets. We also show the standard deviation of performance across different sets.}
\label{fig:loss_plot}
\end{center}
\end{figure}

\begin{table}[t]
\begin{center}
\caption{\small F1 score and classification accuracy for loss type prediction. Our method performs better than all the three baselines.}
\scalebox{1}{
\begin{tabular}{l|c|c}
\hline
%\multirow{2}{*}{Method} & \multicolumn{2}{c|}{Coarse-level} & \multicolumn{2}{c}{Fine-level}\\\cline{2-5}
\multirow{2}{*}{Method} & \multicolumn{2}{c}{Loss function prediction}\\\cline{2-3}
 & F1 score $\bf \uparrow$ & Classification accuracy $\bf \uparrow$ \\
\hline \hline
Random ground-truth & $0.636\pm0.017$ & $0.716\pm 0.028$\\
Mean/mode & $0.751\pm0.027$ & $0.736\pm0.056$\\
No fingerprint & $0.800\pm0.116$ & $0.763\pm0.079$\\
Using one parser & $0.687\pm0.036$ & $0.633\pm0.052$\\
%\rowcolor{LightGray}
Ours & $\bf0.813\pm0.019$ & $\bf 0.792\pm0.021$ \\ \hline \hline
\end{tabular}}
\label{tab:res_loss}
\end{center}
\end{table}

\begin{figure*}[t!]
\begin{center}
\includegraphics[width=\linewidth]{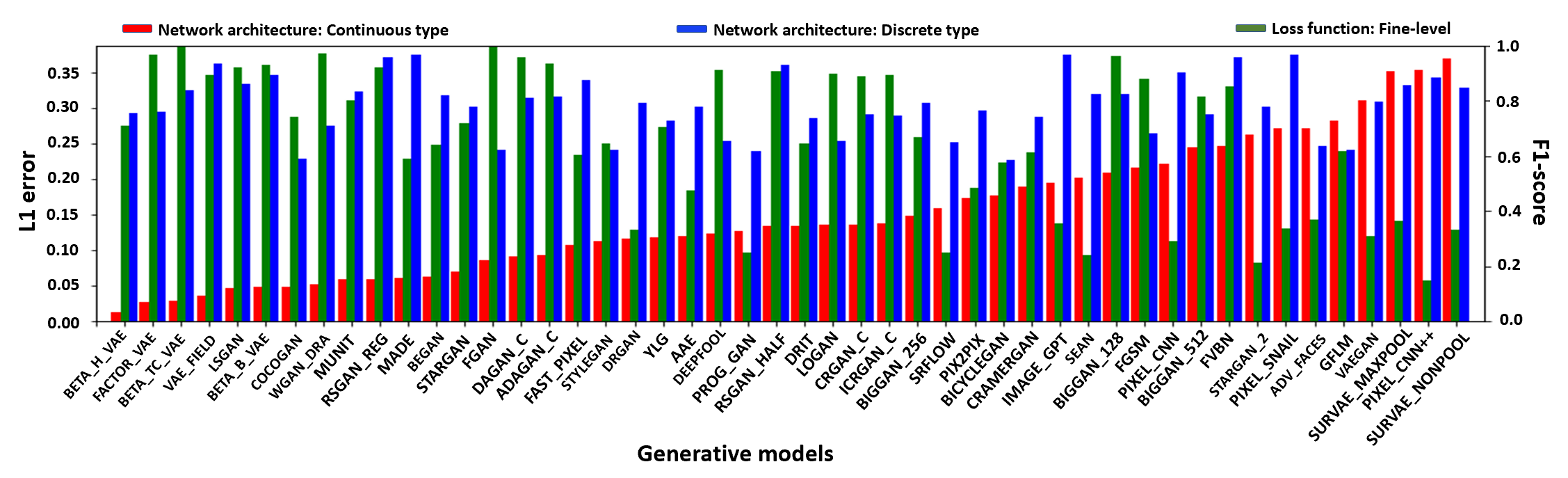}
\caption{\small Performance of all GMs in our $4$  testing sets. Similar performance trends are observed for network architecture and loss functions, \ie, if the $L_1$ error is small for continuous type parameters in network architecture, the high F1 score is also observed for discrete type parameters in network architecture and loss function. In other words, the abilities to reverse engineer the network architecture and loss \VA{function} types for one GM are reasonably consistent.}
\label{fig:gan_err_acc}
\end{center}
\end{figure*}

\begin{table*}[t]
\begin{center}
\caption{\small Performance comparison by varying the training and testing data for face and non-face GMs. Testing performance on non-face GMs is better compared to face GMs. Training and testing on the same content produces better results than on the different contents.\VA{We also show the standard deviation over all the test samples for $L_1$ error. The first value is the standard deviation across sets, while the second one is across the samples.}}
\scalebox{1}{
\begin{tabular}{l|c|cc|c}
\hline
\multirow{3}{*}{Test GMs (\# GMs)} & \multirow{3}{*}{Train GMs (\# GMs)} &\multicolumn{2}{c|}{Network architecture} &Loss function \\ \cline{3-5} 
& &Continuous type &Discrete type &\multirow{2}{*}{F1 score $\bf \uparrow$  } \\
&  & $L_1$ error $\bf \downarrow$ & F1 score $\bf \uparrow$ & \\
\hline\hline
\multirow{3}{*}{Face ($6$)}&Face ($41$) & $0.139\pm0.042\VA{/0.015}$ & $\bf 0.729\pm 0.106$ & $0.788\pm0.146$\\
&Non-face ($69$) & $0.213\pm0.066\VA{/0.136}$ & $0.688\pm0.125$ & $0.759\pm0.100$  \\
&Full ($110$) & $\bf 0.118\pm0.046\VA{/0.040}$ & $0.712\pm0.129$ & $\bf0.833\pm0.136$ \\ \hline
\multirow{3}{*}{Non-face ($6$)}&Non-face ($63$)& $0.118\pm0.021\VA{/0.049}$ & $0.794\pm0.110$ & $0.864\pm0.094$ \\
&Face ($47$)& $0.125\pm0.031\VA{/0.028}$ & $0.667\pm0.099$ & $ 0.858\pm0.115$ \\
&Full ($110$)& $\bf0.082\pm0.045\VA{/0.049}$ & $\bf0.832\pm0.046$ & $\bf 0.886\pm0.061$ \\ \hline 
\multicolumn{2}{c|}{Random guess} & $0.393$& $0.500$ & $0.500$  \\\hline\hline
\end{tabular}}
\label{tab:face-nonface}
\end{center}
\end{table*}

\begin{figure*}[t]
\begin{center}
\includegraphics[width=\linewidth]{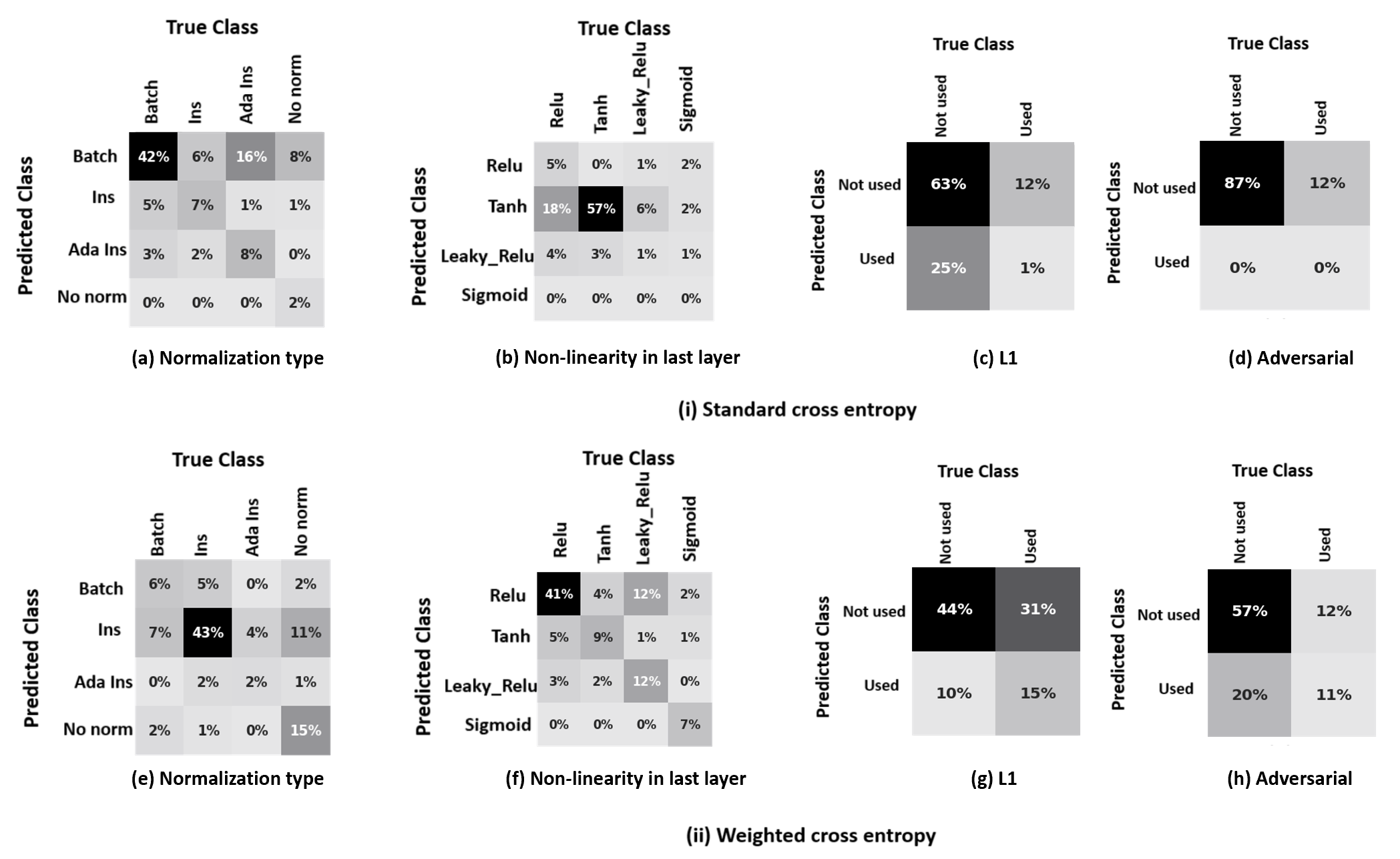}
\caption{\small Confusion matrix in the estimation of four parameters in the network architecture and loss function.  (a)-(d): Standard cross-entropy and (e)-(f): Weighted cross entropy. Weighted cross entropy handles imbalance data much better than the standard cross entropy which usually predicts one class.}
\label{fig:con_mat}
\end{center}
\end{figure*}

\begin{figure*}[t]
\centering
\resizebox{\columnwidth}{!}{
\includegraphics[]{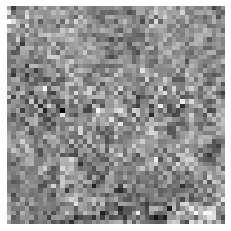}\!
\includegraphics[]{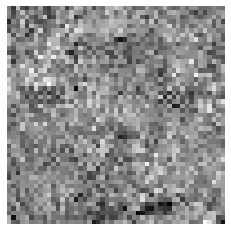}\!
\includegraphics[]{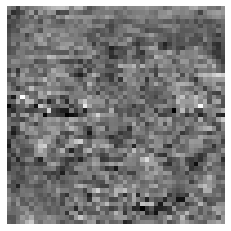}\!
\includegraphics[]{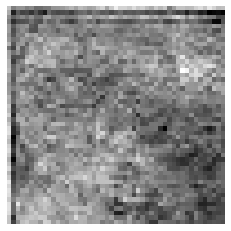}\!
\includegraphics[]{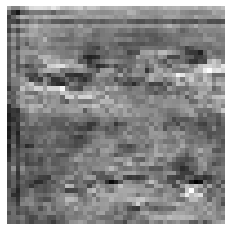}\!
\includegraphics[]{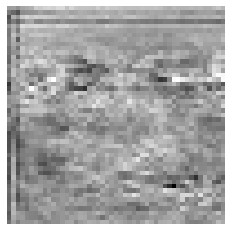}\!
\includegraphics[]{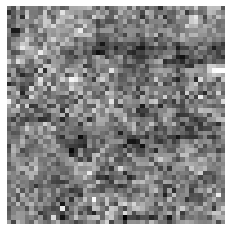}\!
\includegraphics[]{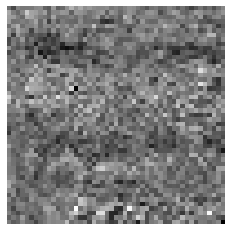}\!
\includegraphics[]{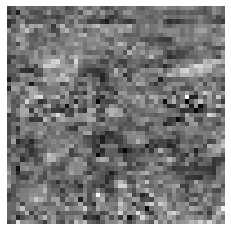}\!
\includegraphics[]{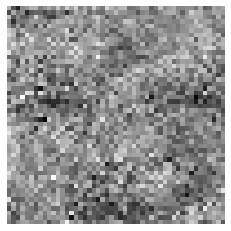}
}
\resizebox{\columnwidth}{!}{
\includegraphics[]{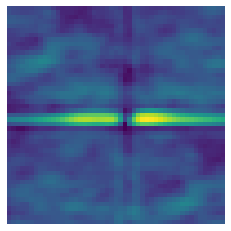}\!
\includegraphics[]{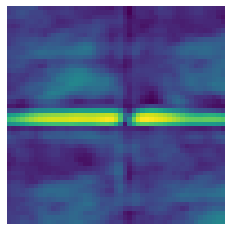}\!
\includegraphics[]{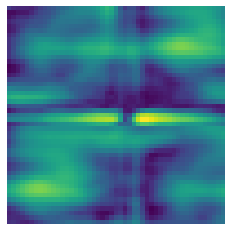}\!
\includegraphics[]{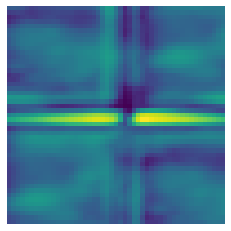}\!
\includegraphics[]{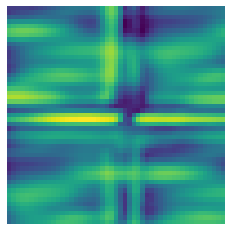}\!
\includegraphics[]{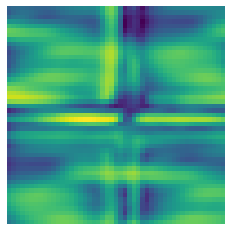}\!
\includegraphics[]{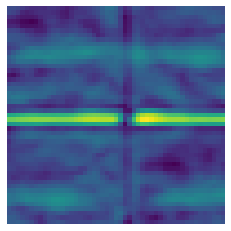}\!
\includegraphics[]{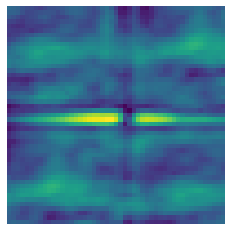}\!
\includegraphics[]{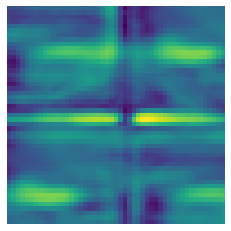}\!
\includegraphics[]{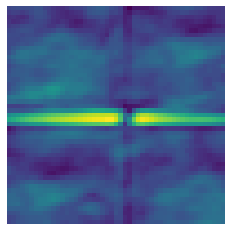}
}

\resizebox{\columnwidth}{!}{
\includegraphics[]{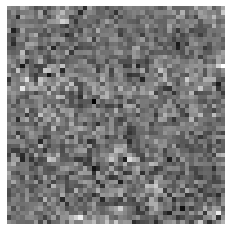}\!
\includegraphics[]{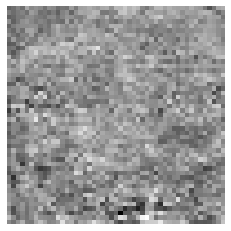}\!
\includegraphics[]{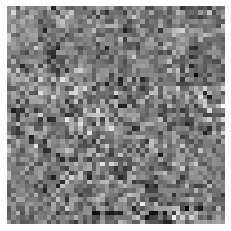}\!
\includegraphics[]{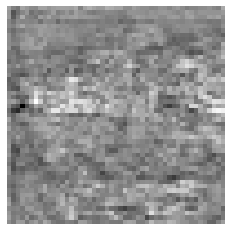}\!
\includegraphics[]{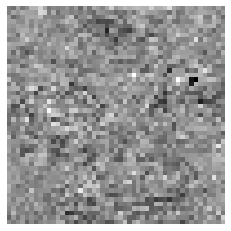}\!
\includegraphics[]{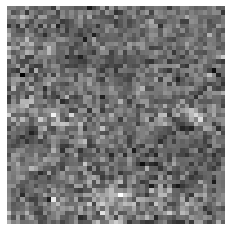}\!
\includegraphics[]{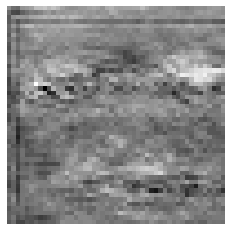}\!
\includegraphics[]{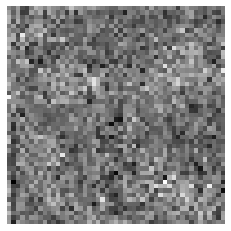}\!
\includegraphics[]{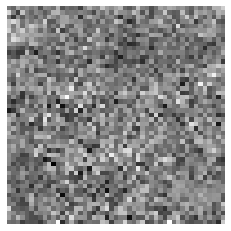}\!
\includegraphics[]{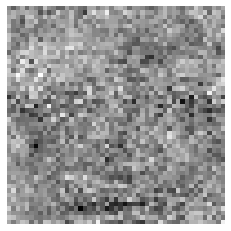}
}
\resizebox{\columnwidth}{!}{
\includegraphics[]{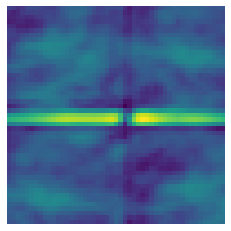}\!
\includegraphics[]{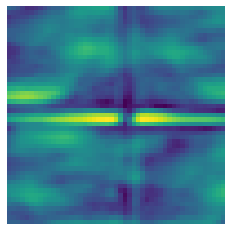}\!
\includegraphics[]{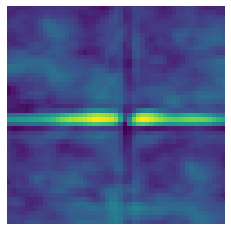}\!
\includegraphics[]{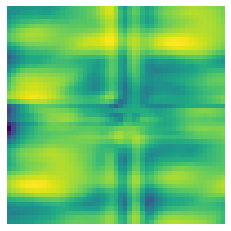}\!
\includegraphics[]{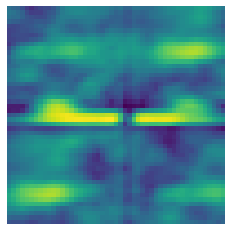}\!
\includegraphics[]{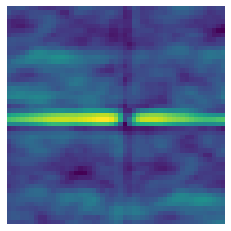}\!
\includegraphics[]{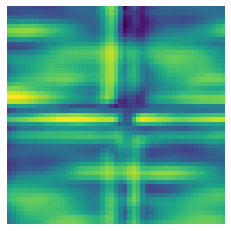}\!
\includegraphics[]{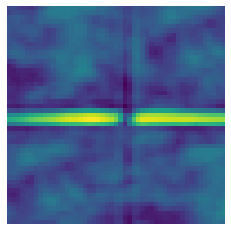}\!
\includegraphics[]{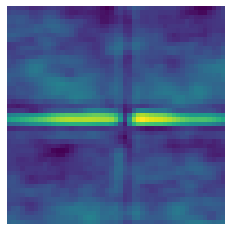}\!
\includegraphics[]{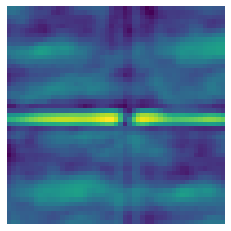}
}

\resizebox{\columnwidth}{!}{
\includegraphics[]{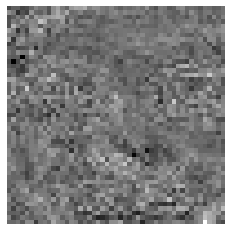}\!
\includegraphics[]{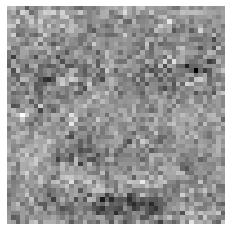}\!
\includegraphics[]{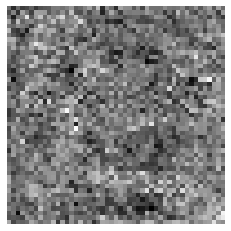}\!
\includegraphics[]{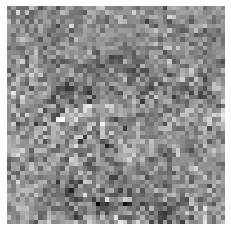}\!
\includegraphics[]{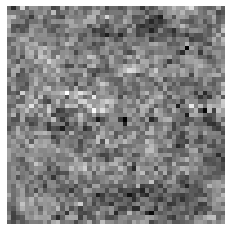}\!
\includegraphics[]{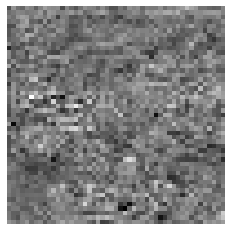}\!
\includegraphics[]{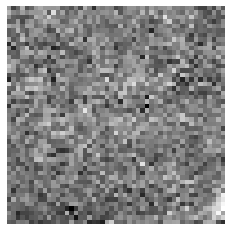}\!
\includegraphics[]{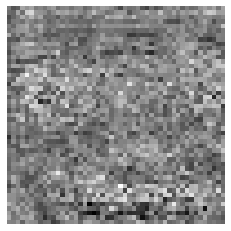}\!
\includegraphics[]{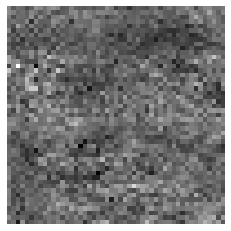}\!
\includegraphics[]{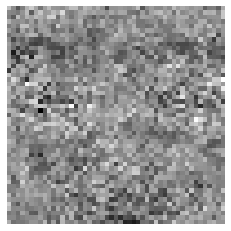}
}
\resizebox{\columnwidth}{!}{
\includegraphics[]{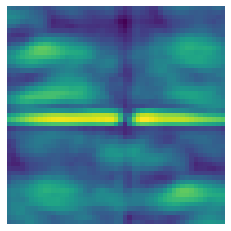}\!
\includegraphics[]{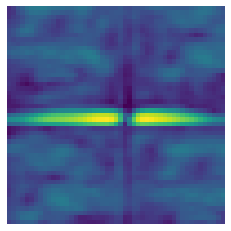}\!
\includegraphics[]{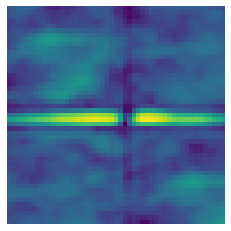}\!
\includegraphics[]{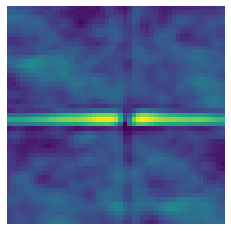}\!
\includegraphics[]{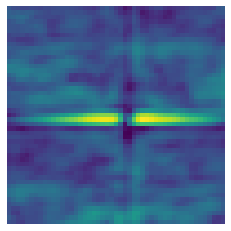}\!
\includegraphics[]{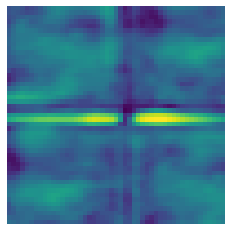}\!
\includegraphics[]{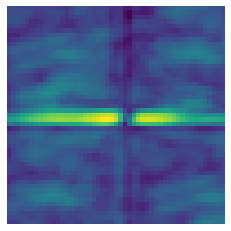}\!
\includegraphics[]{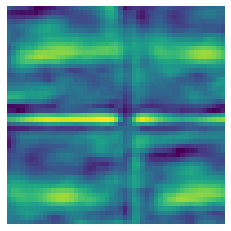}\!
\includegraphics[]{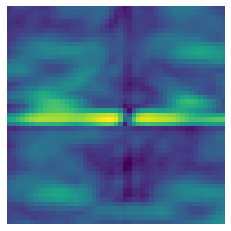}\!
\includegraphics[]{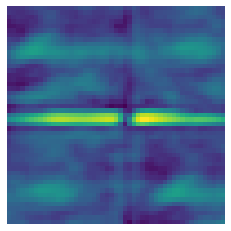}
}

\resizebox{\columnwidth}{!}{
\includegraphics[]{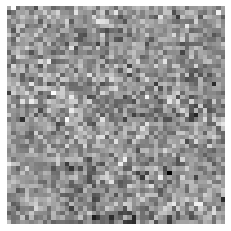}\!
\includegraphics[]{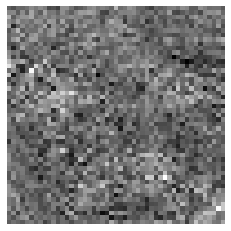}\!
\includegraphics[]{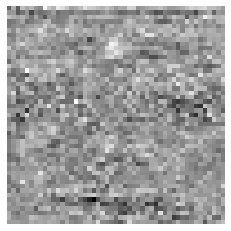}\!
\includegraphics[]{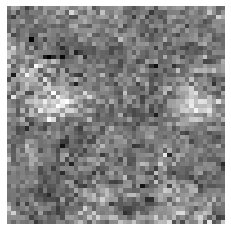}\!
\includegraphics[]{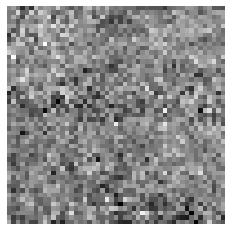}\!
\includegraphics[]{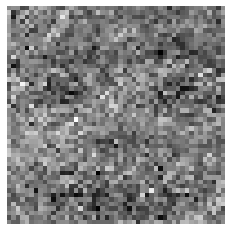}\!
\includegraphics[]{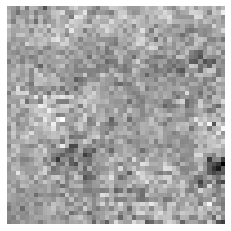}\!
\includegraphics[]{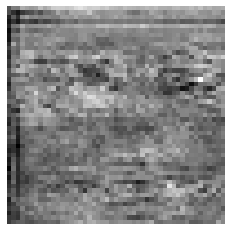}\!
\includegraphics[]{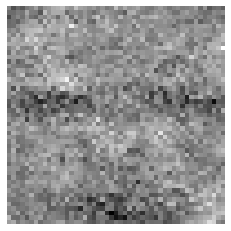}\!
\includegraphics[]{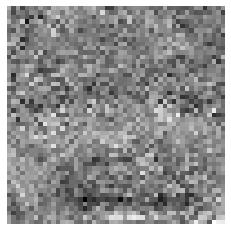}
}
\resizebox{\columnwidth}{!}{
\includegraphics[]{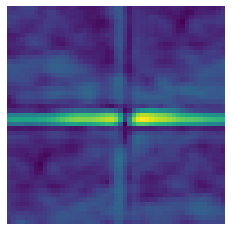}\!
\includegraphics[]{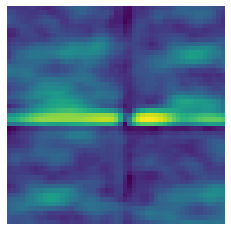}\!
\includegraphics[]{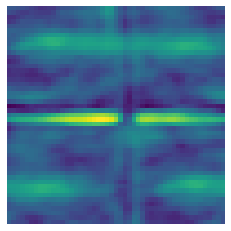}\!
\includegraphics[]{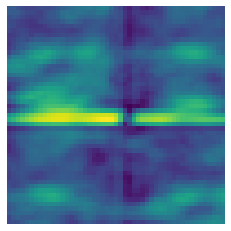}\!
\includegraphics[]{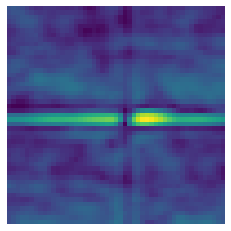}\!
\includegraphics[]{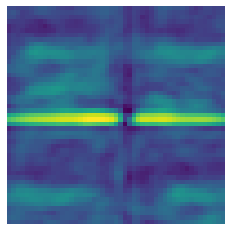}\!
\includegraphics[]{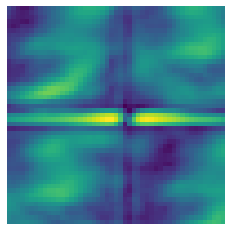}\!
\includegraphics[]{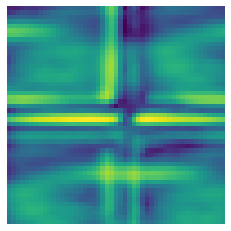}\!
\includegraphics[]{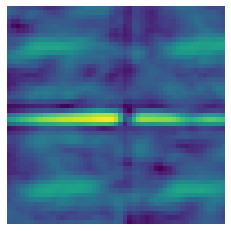}\!
\includegraphics[]{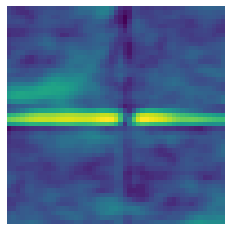}
}

\resizebox{\columnwidth}{!}{
\includegraphics[]{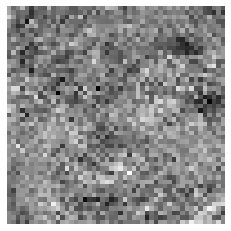}\!
\includegraphics[]{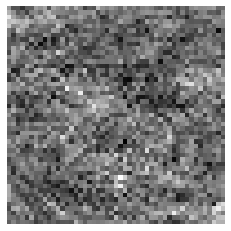}\!
\includegraphics[]{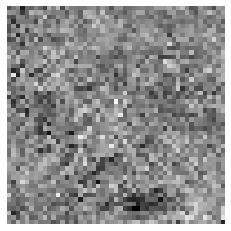}\!
\includegraphics[]{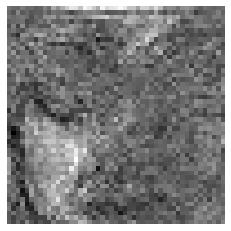}\!
\includegraphics[]{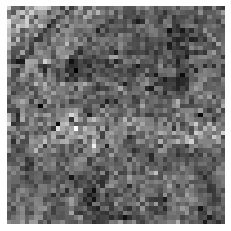}\!
\includegraphics[]{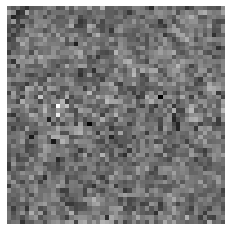}\!
\includegraphics[]{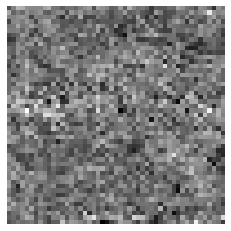}\!
\includegraphics[]{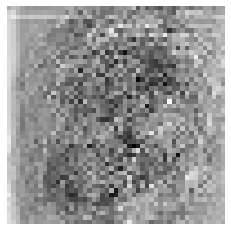}\!
\includegraphics[]{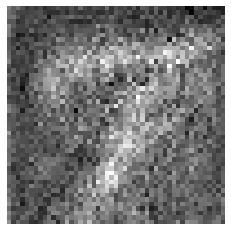}\!
\includegraphics[]{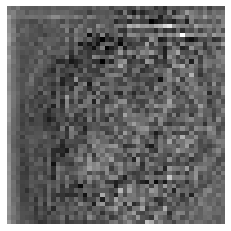}
}
\resizebox{\columnwidth}{!}{
\includegraphics[]{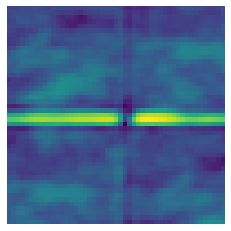}\!
\includegraphics[]{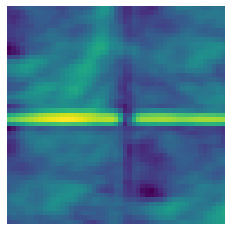}\!
\includegraphics[]{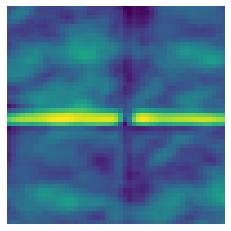}\!
\includegraphics[]{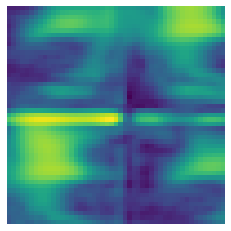}\!
\includegraphics[]{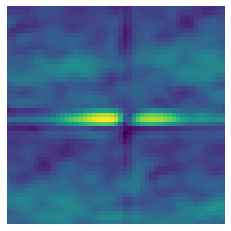}\!
\includegraphics[]{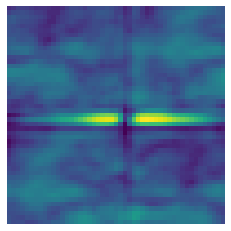}\!
\includegraphics[]{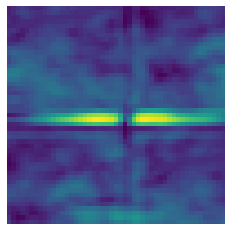}\!
\includegraphics[]{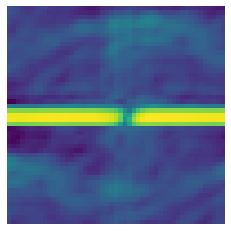}\!
\includegraphics[]{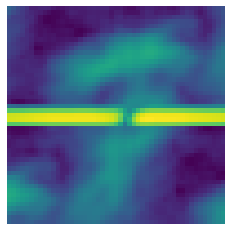}\!
\includegraphics[]{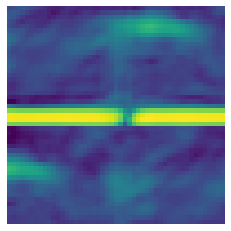}
}

\resizebox{\columnwidth}{!}{
\includegraphics[]{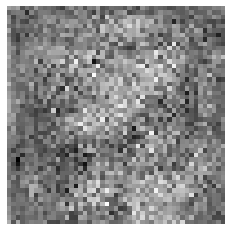}\!
\includegraphics[]{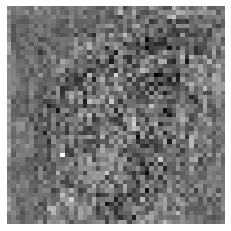}\!
\includegraphics[]{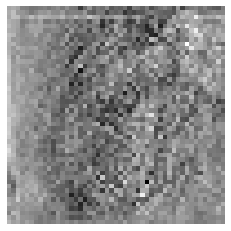}\!
\includegraphics[]{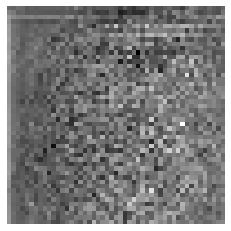}\!
\includegraphics[]{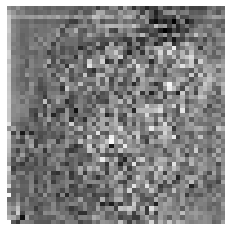}\!
\includegraphics[]{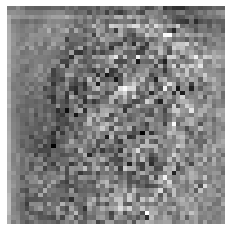}\!
\includegraphics[]{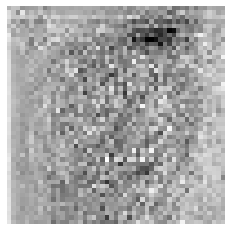}\!
\includegraphics[]{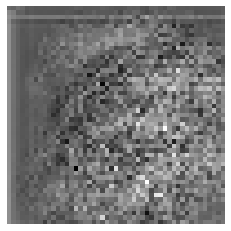}\!
\includegraphics[]{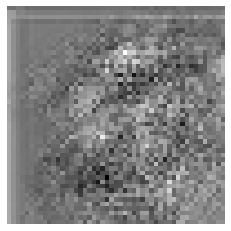}\!
\includegraphics[]{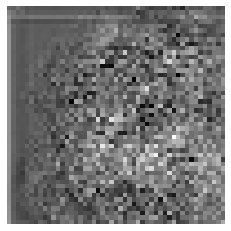}
}
\resizebox{\columnwidth}{!}{
\includegraphics[]{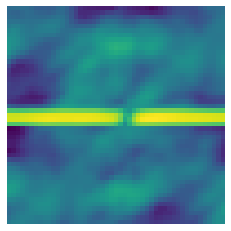}\!
\includegraphics[]{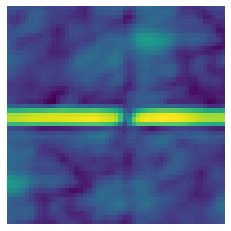}\!
\includegraphics[]{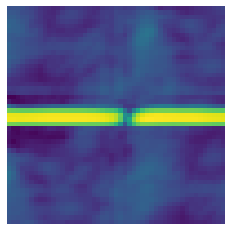}\!
\includegraphics[]{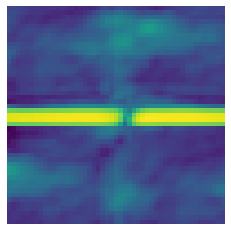}\!
\includegraphics[]{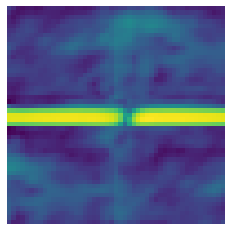}\!
\includegraphics[]{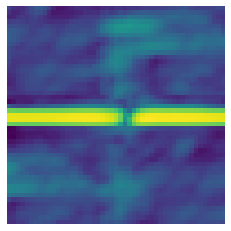}\!
\includegraphics[]{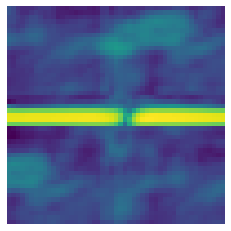}\!
\includegraphics[]{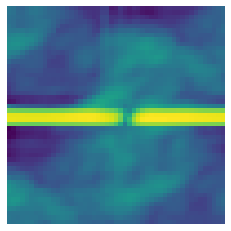}\!
\includegraphics[]{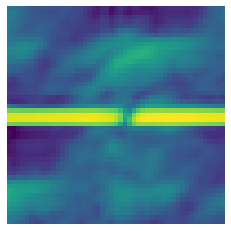}\!
\includegraphics[]{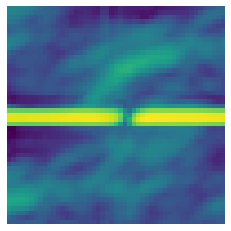}
}

\resizebox{\columnwidth}{!}{
\includegraphics[]{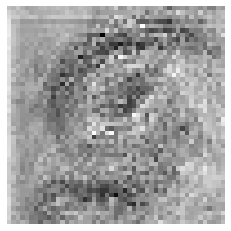}\!
\includegraphics[]{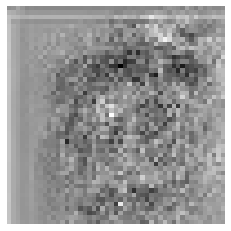}\!
\includegraphics[]{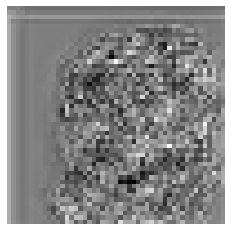}\!
\includegraphics[]{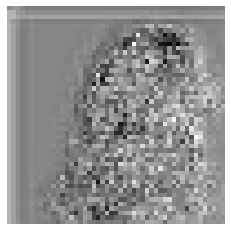}\!
\includegraphics[]{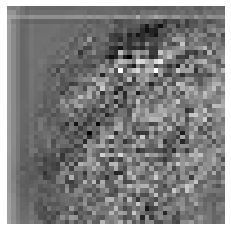}\!
\includegraphics[]{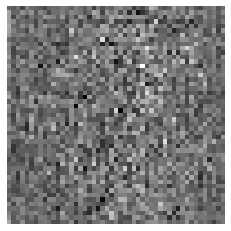}\!
\includegraphics[]{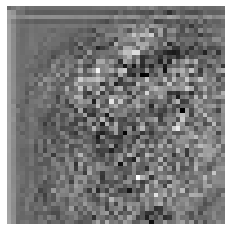}\!
\includegraphics[]{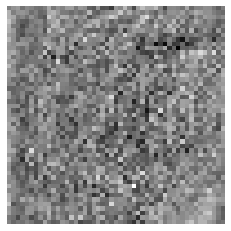}\!
\includegraphics[]{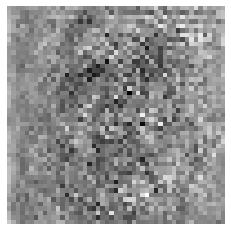}\!
\includegraphics[]{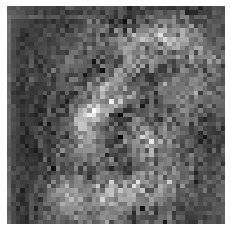}
}
\resizebox{\columnwidth}{!}{
\includegraphics[]{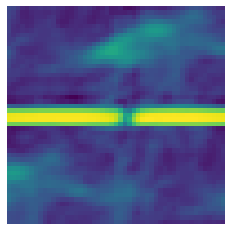}\!
\includegraphics[]{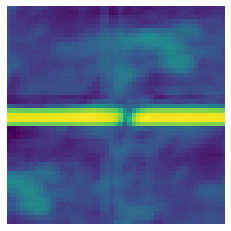}\!
\includegraphics[]{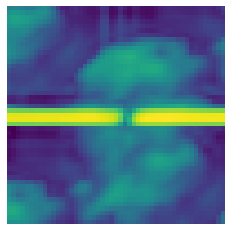}\!
\includegraphics[]{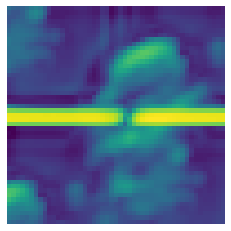}\!
\includegraphics[]{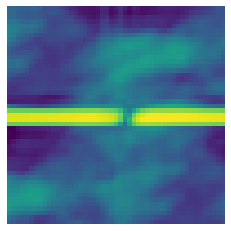}\!
\includegraphics[]{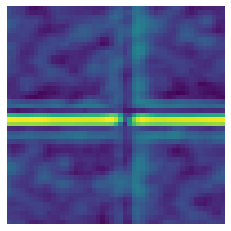}\!
\includegraphics[]{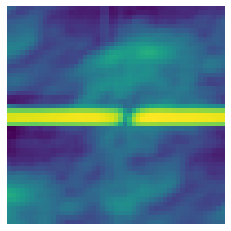}\!
\includegraphics[]{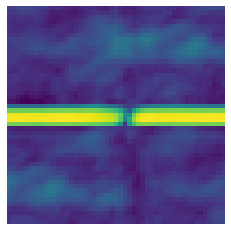}\!
\includegraphics[]{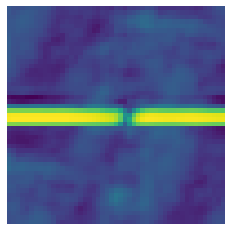}\!
\includegraphics[]{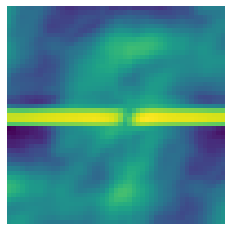}
}

\resizebox{\columnwidth}{!}{
\includegraphics[]{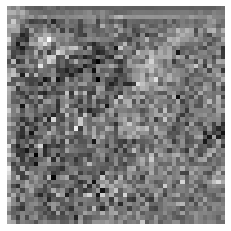}\!
\includegraphics[]{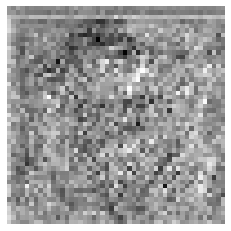}\!
\includegraphics[]{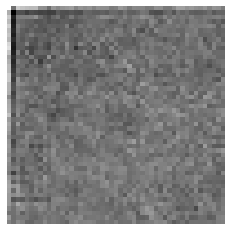}\!
\includegraphics[]{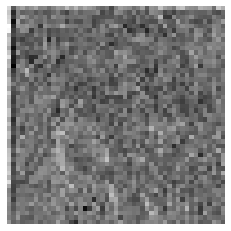}\!
\includegraphics[]{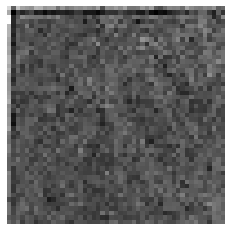}\!
\includegraphics[]{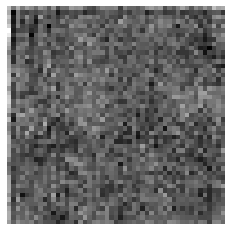}\!
\includegraphics[]{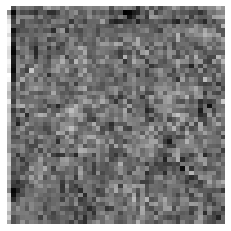}\!
\includegraphics[]{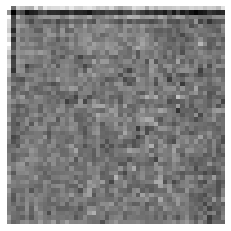}\!
\includegraphics[]{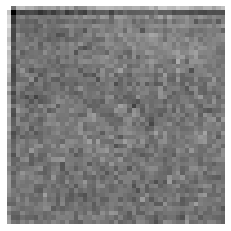}\!
\includegraphics[]{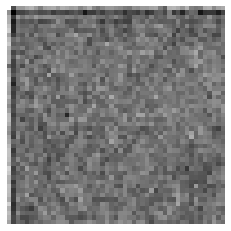}
}
\resizebox{\columnwidth}{!}{
\includegraphics[]{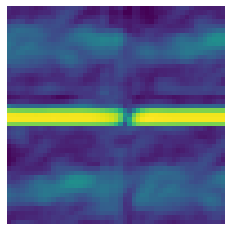}\!
\includegraphics[]{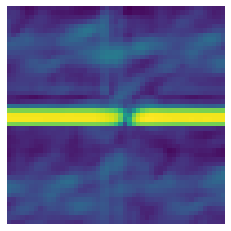}\!
\includegraphics[]{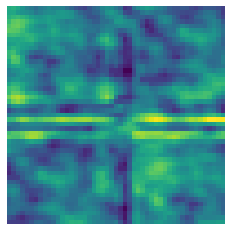}\!
\includegraphics[]{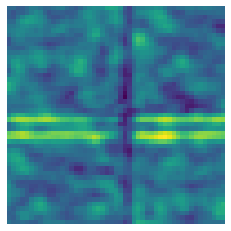}\!
\includegraphics[]{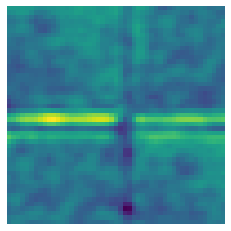}\!
\includegraphics[]{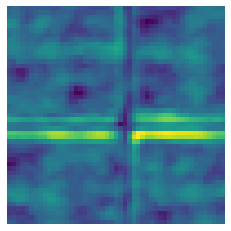}\!
\includegraphics[]{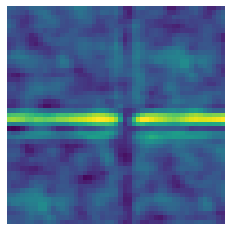}\!
\includegraphics[]{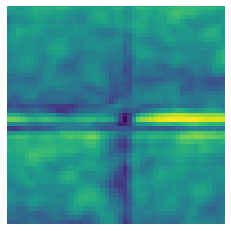}\!
\includegraphics[]{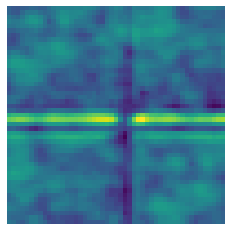}\!
\includegraphics[]{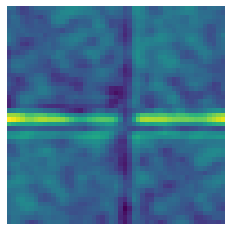}
}

\resizebox{\columnwidth}{!}{
\includegraphics[]{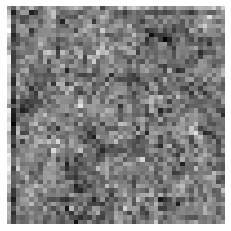}\!
\includegraphics[]{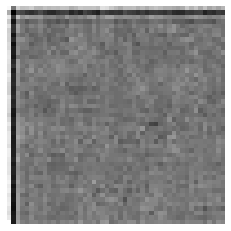}\!
\includegraphics[]{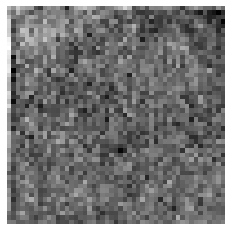}\!
\includegraphics[]{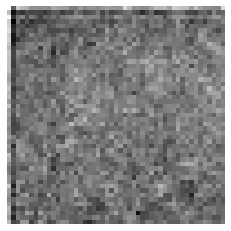}\!
\includegraphics[]{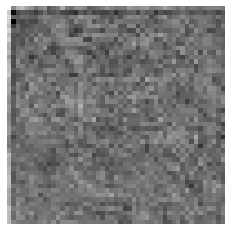}\!
\includegraphics[]{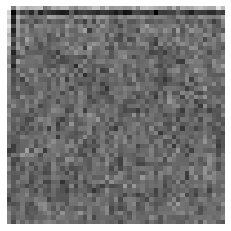}\!
\includegraphics[]{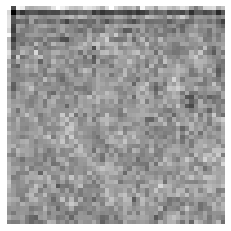}\!
\includegraphics[]{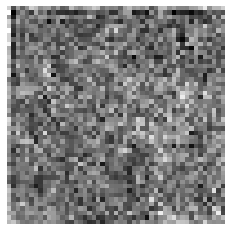}\!
\includegraphics[]{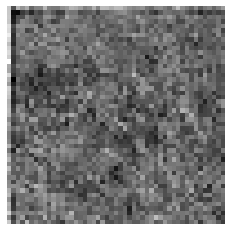}\!
\includegraphics[]{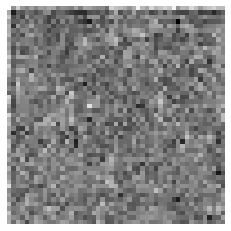}
}
\resizebox{\columnwidth}{!}{
\includegraphics[]{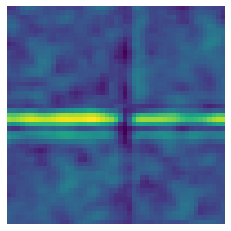}\!
\includegraphics[]{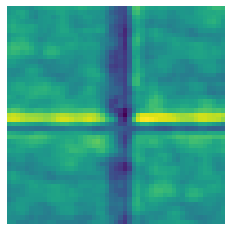}\!
\includegraphics[]{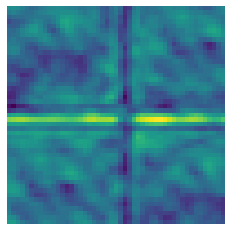}\!
\includegraphics[]{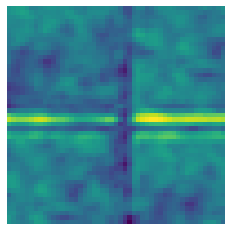}\!
\includegraphics[]{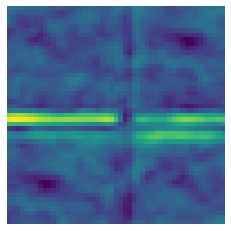}\!
\includegraphics[]{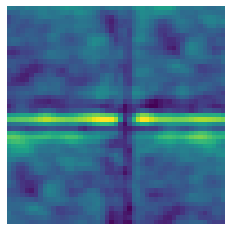}\!
\includegraphics[]{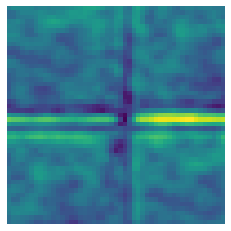}\!
\includegraphics[]{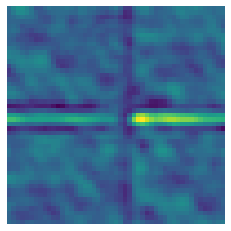}\!
\includegraphics[]{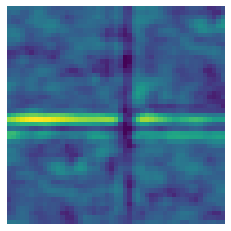}\!
\includegraphics[]{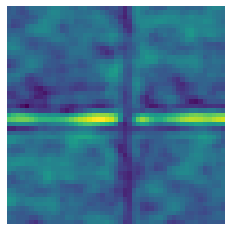}
}

\resizebox{\columnwidth}{!}{
\includegraphics[]{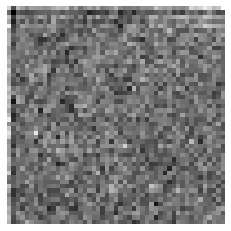}\!
\includegraphics[]{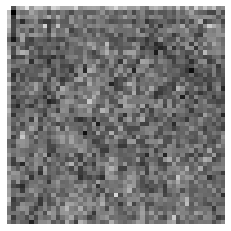}\!
\includegraphics[]{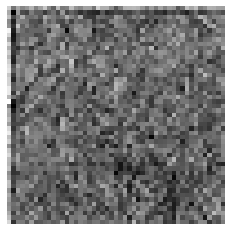}\!
\includegraphics[]{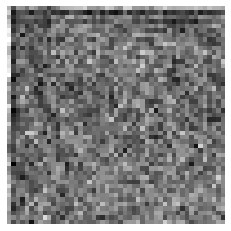}\!
\includegraphics[]{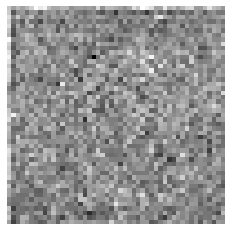}\!
\includegraphics[]{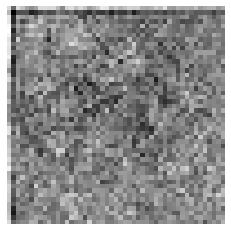}\!
\includegraphics[]{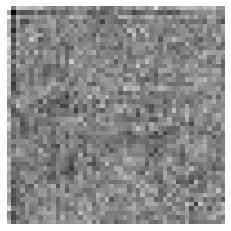}\!
\includegraphics[]{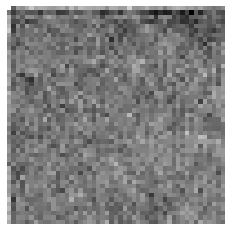}\!
\includegraphics[]{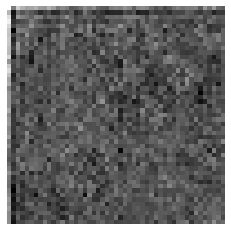}\!
\includegraphics[]{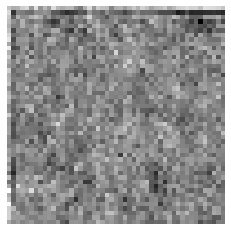}

}
\resizebox{\columnwidth}{!}{
\includegraphics[]{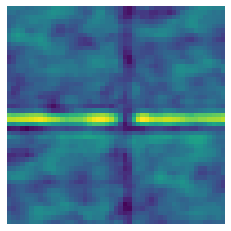}\!
\includegraphics[]{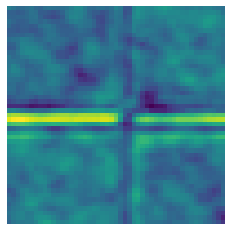}\!
\includegraphics[]{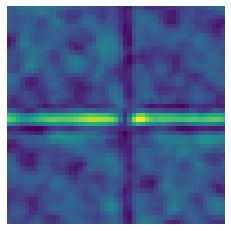}\!
\includegraphics[]{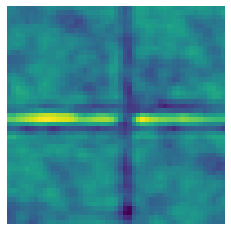}\!
\includegraphics[]{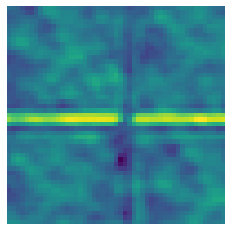}\!
\includegraphics[]{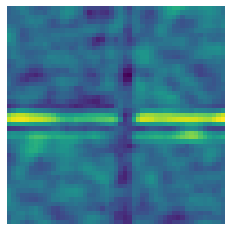}\!
\includegraphics[]{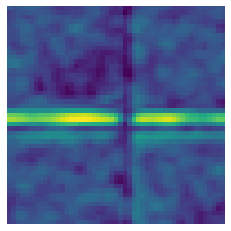}\!
\includegraphics[]{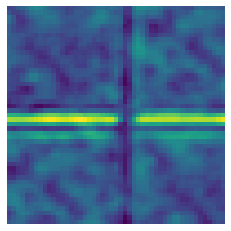}\!
\includegraphics[]{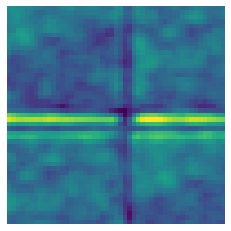}\!
\includegraphics[]{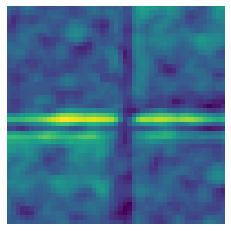}

}

\resizebox{\columnwidth}{!}{
\includegraphics[]{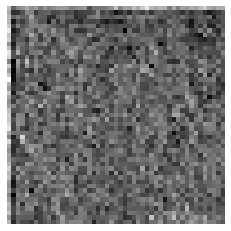}\!
\includegraphics[]{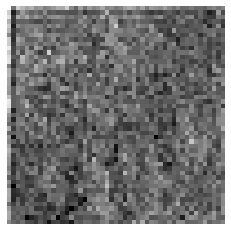}\!
\includegraphics[]{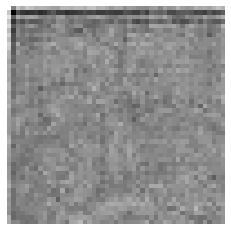}\!
\includegraphics[]{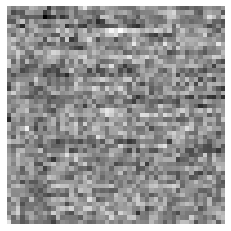}\!
\includegraphics[]{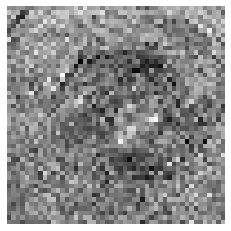}\!
\includegraphics[]{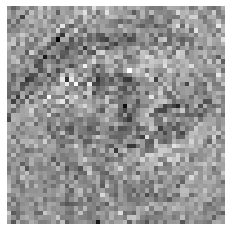}\!
\includegraphics[]{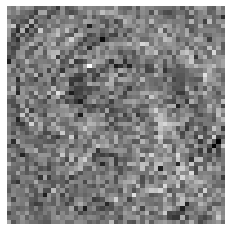}\!
\includegraphics[]{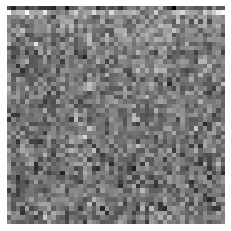}\!
\includegraphics[]{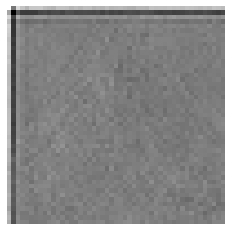}\!
\includegraphics[]{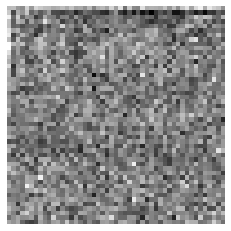}

}
\resizebox{\columnwidth}{!}{
\includegraphics[]{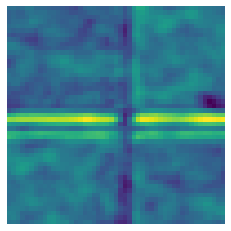}\!
\includegraphics[]{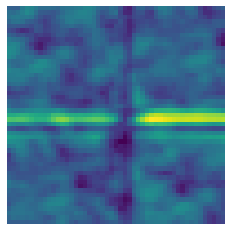}\!
\includegraphics[]{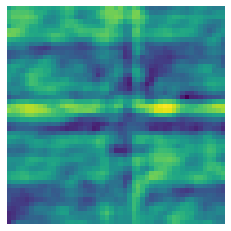}\!
\includegraphics[]{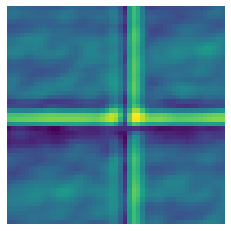}\!
\includegraphics[]{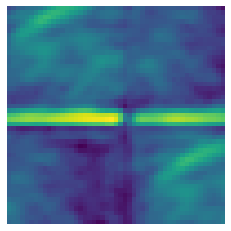}\!
\includegraphics[]{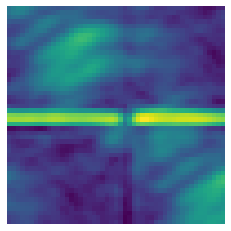}\!
\includegraphics[]{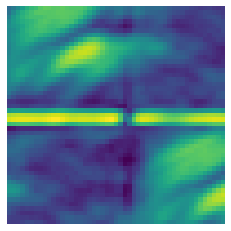}\!
\includegraphics[]{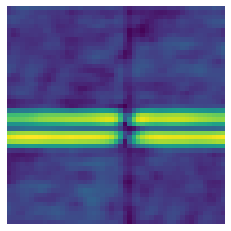}\!
\includegraphics[]{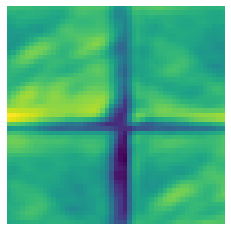}\!
\includegraphics[]{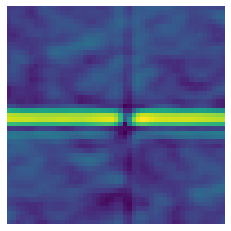}

}

\resizebox{\columnwidth}{!}{
\includegraphics[]{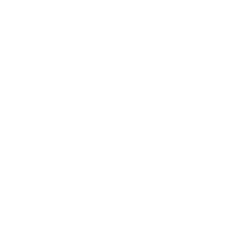}\!
\includegraphics[]{figures_revision/fingerprint/white.png}\!
\includegraphics[]{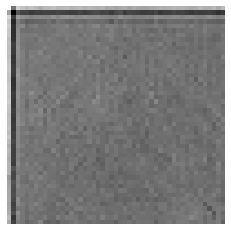}\!
\includegraphics[]{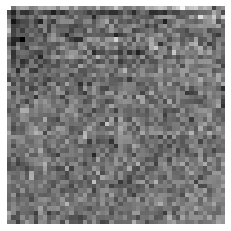}\!
\includegraphics[]{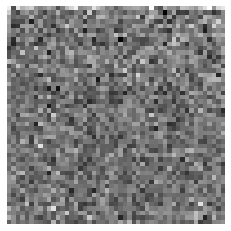}\!
\includegraphics[]{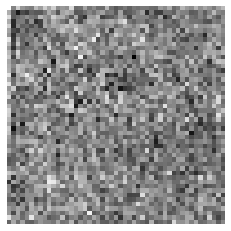}\!
\includegraphics[]{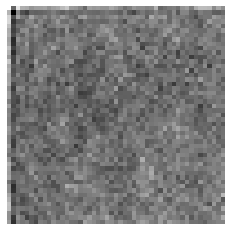}\!
\includegraphics[]{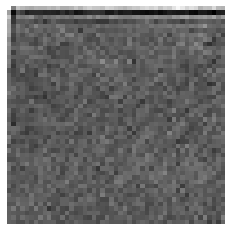}\!
\includegraphics[]{figures_revision/fingerprint/white.png}\!
\includegraphics[]{figures_revision/fingerprint/white.png}
}
\resizebox{\columnwidth}{!}{
\includegraphics[]{figures_revision/fingerprint/white.png}\!
\includegraphics[]{figures_revision/fingerprint/white.png}\!
\includegraphics[]{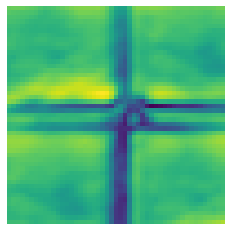}\!
\includegraphics[]{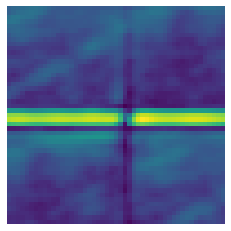}\!
\includegraphics[]{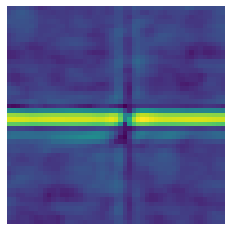}\!
\includegraphics[]{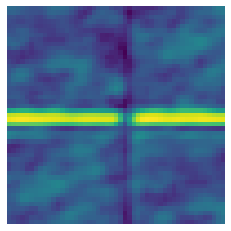}\!
\includegraphics[]{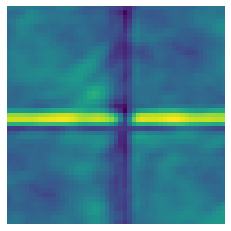}\!
\includegraphics[]{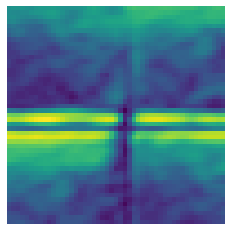}\!
\includegraphics[]{figures_revision/fingerprint/white.png}\!
\includegraphics[]{figures_revision/fingerprint/white.png}
}

\vspace{2mm}
\caption{\small Estimated fingerprints (left) and corresponding frequency spectrum (right) from one generated image of each of 116 GMs. Many frequency spectrums show distinct \VA{high-frequency} signals, while some appear to be similar to each other.}
\label{fingerprints}
\end{figure*}

\begin{table}[t]
\begin{center}
\caption{\small Ablation study of  the $4$ loss terms in fingerprint estimation. Removing any one loss for fingerprint estimation deteriorates the performance with the \VA{worst} results in the case of removing all losses. [KEYS: fing.: fingerprint]. \VA{We also show the standard deviation over all the test samples for $L_1$ error. The first value is the standard deviation across sets, while the second one is across the samples.}}
\scalebox{0.9}{
\begin{tabular}{@{}l|cc|c@{}}
\hline
\multirow{3}{*}{Loss removed} &\multicolumn{2}{c|}{Network architecture} &Loss function \\ \cline{2-4} 
&Continuous type &Discrete type &\multirow{3}{*}{F1 score $\bf \uparrow$} \\
& $L_1$ error $\bf \downarrow$ & F1 score $\bf \uparrow$ & \\
\hline\hline
Magnitude loss & $0.156\pm0.007\VA{/0.009}$ & $0.674\pm0.012$ & $0.755\pm0.046$ \\
Spectrum loss & $\bf0.149\pm0.022\VA{/0.016}$ & $0.676\pm0.034$ & $0.786\pm0.042$ \\
Repetitive loss& $0.150\pm0.018\VA{/0.026}$ & $0.708\pm0.031$ & $0.794\pm0.031$ \\ 
Energy loss & $0.162\pm0.032\VA{/0.038}$ & $0.703\pm0.045$ & $0.785\pm0.028$ \\
All (no fing.) & $0.170\pm0.035\VA{/0.037}$ & $0.700\pm0.032$ & $0.800\pm0.016$ \\
Nothing (ours)& $\bf0.149\pm0.019\VA{/0.014}$ & $\bf0.718\pm 0.036$  &$\bf0.813\pm0.019$ \\\hline \hline
\end{tabular}}
\label{tab:loss_ablation}
\end{center}
\end{table}

\begin{figure*}[t]
\begin{center}
%\vspace{-3mm}
\includegraphics[width=\linewidth]{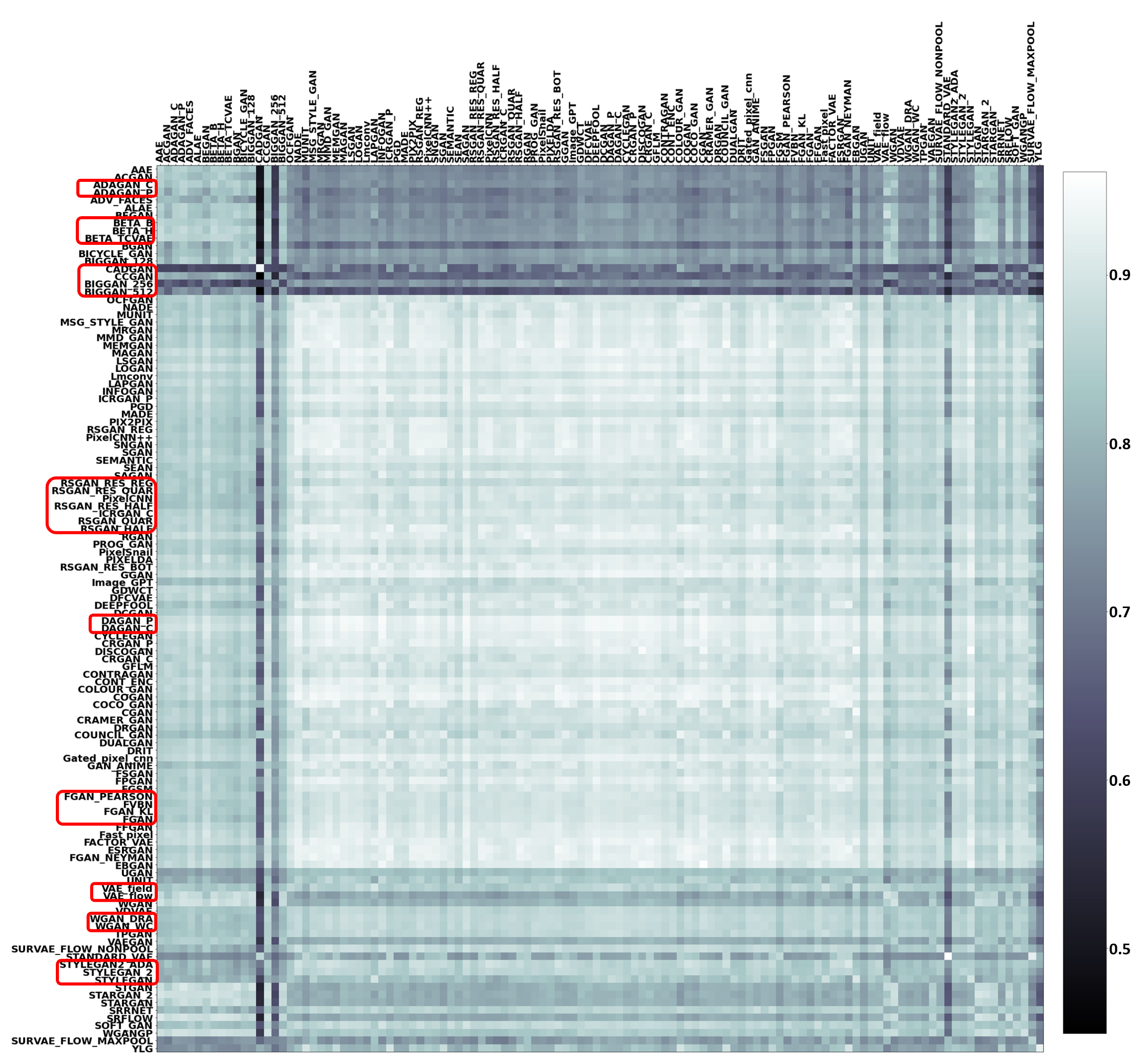}
%\vspace{-5mm}
\caption{\small Cosine similarity matrix for pairs of $116$ GM's fingerprints. Each element of this matrix is the average Cosine similarities of $50$ pairs of fingerprints from two GMs. We see the higher intra-GM and lower inter-GM similarities. We can also see GMs with similar network architecture or loss function are clustered together, as shown in the red boxes on the left.}
\label{fig:corr_gan}
%\vspace{-4mm}
\end{center}
\end{figure*}

\minisection{Network architecture prediction} We report the results of network architecture prediction in Tab.~\ref{tab:res_net} for the $4$ testing sets, as defined in Sec.~\ref{sec:exp:setting}. % containing GM trained on face datasets. 
%We report mean performance across all six evaluation sets. 
Our method achieves a much lower $L_1$ error compared to the random ground-truth baseline for continuous type parameters and higher classification accuracy and F1 score for discrete type parameters. 
This result indicates that there is indeed a much stronger and generalized correlation between generated images and the embedding space of meaningful architecture hyper-parameters and loss function types, compared to a random vector of the same length and distribution. This correlation is the foundation of why model parsing of GMs can be a valid and feasible task.
\VA{Our approach also outperforms the mean/mode baseline, proving that always predicting the mean of the data for continuous parameters is not good enough.} 
Removing fingerprint estimation objectives leads to worse results showing the importance of the fingerprint estimation in model parsing. We demonstrate the effectiveness of estimating mean and deviation by evaluating the performance of using just one parser. Our method clearly outperforms the approach of using one parser.

Figure~\ref{fig:err_bar} shows the detailed $L_1$ error and F1 score for all network architecture parameters. 
We observe that our method performs substantially better than the random ground-truth baseline for almost all parameters. 
As for the no fingerprint and using one parser baselines, our method is still better in most cases with a few parameters showing similar results. We also show the standard deviation of every estimated parameter for all the methods. Our proposed approach in general has smaller standard deviations than the two baselines.
%Even though for some parameters ({\it e.g.}, \# parameters), the performance of random ground-truth is similar, the standard deviation is reducing for our proposed approach
%showing our consistent estimation with less variability across different testing  sets. 
For continuous type parameters, we further show the effectiveness of regression prediction by evaluating three metrics namely, correlation coefficient, coefficient of determination and slope of RANSAC regression line. These metrics are evaluated between prediction and ground-truth. \VA{Further, we also estimate a p-value of a t-test, where the null hypothesis is as follows: the sequence of sample-wise $L_1$ error differences between our method and the baseline method is sampled from zero-mean Gaussian. This p-value would be estimated for every ours-baseline pair. We report the mean and the standard deviation across all four sets. The p-value of our approach when compared to all the three baselines is less than $0.05$, thereby rejecting the null hypothesis and proving our improvement is statistically significant.} For other three metrics, the values closer to $1$ shows effective regression. For our method, we have slope of $0.921$, correlation coefficient of $0.744$ and coefficient of determination as $0.612$ which shows the effectiveness of our approach. Further, our approach outperforms all the baselines for all three metrics.

%We further calculate the averaged $L_1$ error for each parameter in the network architecture vector defined in Tab.~\ref{tab:net}. As shown in Fig.~\ref{fig:err_bar}, the most easy-to-estimate parameters include whether attention is used and the number of pooling layers. The hardest parameters include down-sampling type and skip connections. Fig.~\ref{fig:gan_err_acc} shows the $L_1$ error for each GM averaged across all parameters. The VAEs are much easier to estimate than GAN models potentially due to VAEs have simpler and similar network architectures. We also attempted to use a self-attention mechanism to select and estimate a subset of easy-to-estimate network parameters. But we did not observe a significant performance difference. How these parameters will affect the generated images is an interesting future direction to explore. 

\begin{table}[t]
\begin{center}
\caption{\small Network architecture estimation  and loss function prediction when given multiple images of one GM. Performance increases when enlarging the number of images for evaluation from $1$ to $10$. Performance becomes stable for more than $10$ images. \VA{We also show the standard deviation over all the test samples for $L_1$ error. The first value is the standard deviation across sets, while the second one is across the samples.} }
\scalebox{0.9}{
\begin{tabular}{l|cc|c}
\hline
\multirow{3}{*}{$\#$ images} &\multicolumn{2}{c|}{Network architecture} &Loss function \\ \cline{2-4} 
&Continuous type &Discrete type & \multirow{3}{*}{F1 score $\bf \uparrow$ }\\
& $L_1$ error $\bf \downarrow$ & F1 score $\bf \uparrow$ &  \\
\hline\hline
$1$ & $0.215\pm0.054\VA{/0.067}$ & $0.696\pm0.089$ & $ 0.798\pm0.010$\\
$10$ & $0.151\pm0.033\VA{/0.039}$ &  $\bf 0.726\pm0.075$ & $ 0.793\pm0.070$ \\
$100$ & $\bf 0.145\pm0.032\VA{/0.036}$ & $ 0.721\pm0.073$ &  $0.789\pm0.071$ \\
$500$ & $0.146\pm0.033\VA{/0.031}$ & $ 0.720\pm0.070$ & $\bf 0.808\pm0.007$ \\\hline \hline
\end{tabular}}
\label{tab:eval_size}
\end{center}\vspace{-3mm}
\end{table}

\minisection{Loss function prediction} We calculate the F1 score and classification accuracy for loss function parameters. The performance are shown in Tab.~\ref{tab:res_loss}. For the random ground-truth baseline, the performance is close to a random guess. Our approach performs much better than all the baselines. 
Figure~\ref{fig:loss_plot} shows the detailed F1 score for all loss function parameters. Apparently our method works better than all the baselines for almost all parameters.  We also show the standard deviation of every estimated parameter for all the methods. Similar behaviour of standard deviation for different methods was observed as in the network architecture.
Figure~\ref{fig:gan_err_acc} provides another perspective of model parsing by showing the performance in terms of $48$ unique GMs across our $4$ testing sets.

%We observe that NLL, KL, MMD and MSE loss are easier to estimate as shown in Fig.~\ref{fig:loss_plot}. Pixel-level loss is more challenging probably because the similarity between the fine-level losses make it hard to distinguish. As shown in Fig.~\ref{fig:gan_err_acc}, the performance for VAEs is higher than GAN models. We also observe high performance at fine-level when coarse-level performance is high, indicating the positive effect of our hierarchical learning for loss prediction.

\minisection{Practical Usage of Model Parsing.}
As our work is the first one to propose the task of model parsing, it's beneficial to ask the question: \textit{what is the performance desired for practical usage of model parsing in the real world?} To answer this question, we can expect that an error less than $10\%$ can be considered useful for the practical application of model parsing. The rationale is the following. We consider two of the most similar generative models, RSGAN\_HALF and RSGAN\_QUAR, in our dataset. Upon further analysis, we observe that these models differ in only $2$ out of $15$ parameters. Therefore, we argue that an error rate below $10\%$ is reasonable for practical purposes as this error is less than the difference between the two most similar generative models. Therefore, for the task of model parsing, we expect $L_1$ error of less than $0.1$ and an $F1$ score of over $90\%$ for practical usage. Our proposed approach achieves an $L_1$ error slightly above $10\%$ ($0.14$) and an $F1$ score of $80\%$, both of which have reasonable margins toward the above mentioned thresholds.

\subsection{Ablation study}
\minisection{Face \vs non-face GMs}
Our dataset consists of $47$ GMs trained on face datasets and $69$ GMs trained on non-face datasets. 
%In other words, the contents of the  images generated by these GMs are quite diverse. 
Let's denote these GMs as face GMs and non-face GMs, respectively. 
All aforementioned experiments are conducted by training on $104$ GMs and evaluating on $12$ GMs. 
Here we conduct an ablation study to train and evaluate on different types of GMs. 
We study the performance on face and non-face testing GMs when training on three different training sets, including only face GMs, only non-face GMs and all GMs. 
Note that all testing GMs are excluded during training each time.
We also add a baseline where both regression and classification make a random guess on their estimation.

The results are shown in Tab.~\ref{tab:face-nonface}. We have three observations. First, model parsing for non-face GMs are easier than face GMs. This might be partially due to the generally lower-quality images generated by non-face GMs compared to those by face GMs, thus more traces are remained for model parsing.
Second, training and testing on the same content can generate better results than on different contents. 
Third, training on the full datasets improves some parameter estimation but may hurt other parameters slightly.

\minisection{Weighted cross-entropy loss}
As mentioned before, the ground truth of many network hyperparameters have biased distributions.
For example, the ``normalization type" parameter in Tab.~\ref{tab:net} has uneven distribution among its $4$ possible types. 
With this biased distribution, our classifier might make a constant prediction to the type with the highest probability in the ground truth, as this could minimize the loss especially for severe biasness.
This degenerate classifier clearly has no value to model parsing.
%out dataset has imbalance for some classification parameters which can result for the individual classifier to prefer predicting one class than others. 
To address this issue, we propose to use the weighted cross-entropy loss with different loss weights for each class.
These weights are calculated using the ground-truth distribution of every parameter in the full dataset. 
To validate if the above approach is able to remedy this issue, we compare it with the standard cross-entropy loss.

Figure~\ref{fig:con_mat} shows the confusion matrix for discrete type parameters in network architecture prediction and coarse/fine level parameters in loss function prediction. 
The rows in the confusion matrix are represented by predicted classes and columns are represented by the ground-truth classes. 
We clearly see that the classifier is mostly biased towards more frequent classes in all $4$ examples, when the standard cross-entropy loss is used. 
However, this problem is remedied when using the weighted cross-entropy loss, and the classifiers make meaningful predictions.
%which considers the distribution of data across various parameters to handle data imbalance issues.

\minisection{Fingerprint losses}
We proposed four loss terms in Sec.~\ref{sec:fen} to guide the training of the fingerprint estimation including magnitude loss, spectrum loss, repetitive loss and energy loss. 
We conduct an ablation study to demonstrate the importance of these four losses in our proposed method. 
This includes four experiments, each removing one of the loss terms and comparing the performance with our proposed method (remove nothing) and no fingerprint baseline (remove all). 
As shown in Tab.~\ref{tab:loss_ablation},  removing any loss for fingerprint estimation hurts the performance. 
Our ``no fingerprint" baseline, for which we remove all losses, performs worst of all. 
Therefore, each loss clearly has a positive effect on the fingerprint estimation and model parsing. % of hyperparameters. 

\minisection{Model parsing with multiple images}
We evaluate model parsing when varying the number of test images. For each GM, we randomly select $1$, $10$, $100$, and $500$ images per GM from different face GMs sets for evaluation.
With multiple images per GM, we average the prediction for continuous type parameters and take majority voting for discrete type parameters and loss function parameters. We compute the $L_1$ error and F1 score for the continuous and discrete type parameters respectively and average the result across different sets. We repeat the above experiment multiple times, each time randomly selecting the number of images.  %we average the $L_1$ error for continuous type parameters and calculate the F1-score for every set of images.  
We compare the $L_1$ error and F1 score for respective parameters. Tab.~\ref{tab:eval_size} shows noticeable gains with $10$ images and minor gains with $100$ images. There is \VA{not} much performance difference when evaluating on $100$ or $500$ images, which suggests that our framework is robust in generating consistent results when tested on different numbers of generated images by the same GM.

\minisection{Content-independent fingerprint}
%To verify the generalizability of our fingerprint, we try to do the classification on the dataset type for all models. 
Ideally our estimated fingerprint should be independent of the content of the image.
That is, the fingerprint only includes the trace left by the GM while not indicating the content in any way.
To validate this, we partition all GMs into four classes based on their contents: FACES ($47$ GMs), MNIST ($25$), CIFAR10 ($31$), and OTHER ($13$). 
Every class  has images generated by the GMs belong to this class. 
%We train our FEN separately and then 
We feed these images to a pre-trained FEN and obtain their fingerprints. 
Then we train a shallow network consisting of five convolutional layers and two fully connected layers for a $4$-way classification.
However, we observe the training cannot converge. 
This means that our estimated fingerprint from FEN doesn't have any content-specific properties for content classification. 
As a result, the model parsing of the hyperparameters doesn't leverage the content information across different GMs, which is a desirable property.

\minisection{\VA{Evaluation on diffusion models}} 
\VA{Due to the recent advancement of diffusion models for fake media generation, we evaluate our approach for these generative models. Specifically, we collect $7$ diffusion models with $1K$ images each. We create $4$ different test set splits, each set containing $3$ diffusion models selected randomly. The remaining diffusion models, along with the full dataset is used for training. The result for our approach along with all the baselines is shown in Tab.~\ref{tab:diff_models}. Our method clearly outperforms all the baselines, indicating the effectiveness of our approach for unseen models proposed in future. \VA{We also show the standard deviation over all the test samples for $L_1$ error. The first value is the standard deviation across sets, while the second one is across the samples.} }

%{@{}l|cc|c@{}}
\begin{table}[t]
\begin{center}
\caption{\VA{\small Evaluation on diffusion models. We also show the standard deviation over all the test samples for $L_1$ error. The first value is the standard deviation across sets, while the second one is across the samples.}}
\begin{adjustbox}{width=1\columnwidth}
\begin{tabular}{@{}l|cc|c@{}}
\hline
 &\multicolumn{2}{c|}{\VA{Network architecture}} &\VA{Loss function} \\ \cline{2-4} 
&\VA{Continuous type} &\VA{Discrete type} &\multirow{3}{*}{\VA{F1 score $\bf \uparrow$}} \\
\multirow{-3}{*}{\VA{Method}}& \VA{$L_1$ error $\bf \downarrow$} & \VA{F1 score $\bf \uparrow$} & \\
\hline\hline
\VA{Random ground-truth}  & \VA{$0.240\pm0.065/0.069$} & \VA{$0.664\pm 0.105$} & \VA{$0.619\pm0.083$}  \\
\VA{No fingerprint} & \VA{$0.211\pm0.080/0.078$} & \VA{$0.764\pm0.112$} & \VA{$0.711\pm 0.085$}  \\
\VA{Using one parser} & \VA{$0.201\pm 0.045/0.041$} & \VA{$0.564\pm0.101$} & \VA{$0.654\pm 0.054$}  \\
\VA{Ours} & \VA{$\bf 0.189\pm 0.051/0.049$} & \VA{$\bf0.787\pm 0.099$} & \VA{$\bf0.724\pm 0.076$} \\\hline \hline
\end{tabular}
\end{adjustbox}
\label{tab:diff_models}
\end{center}
\end{table}

\subsection{Visualization}
Figure~\ref{fingerprints} shows an estimated fingerprint image and its frequency spectrum averaged over $25$ randomly selected images per GM. 
We observe that estimated fingerprints have the desired properties defined by our loss terms, including low magnitude and highlights in middle and high frequencies. %This evidence validates the effectiveness of our proposed fingerprint estimation.%, which performs well for various tasks including deepfake detection, image attribution, and model parsing.

We also find that the fingerprints estimated from different generated images of the same GM are similar. To quantify this, we compute a Cosine similarity matrix ${\bf{C}}\in\mathbb{R}^{116\times116}$ where ${\bf{C}}(i, j)$ is the averaged Cosine similarity of $25$ randomly sampled fingerprint pairs from GM $i$ and $j$. The matrix $\bf{C}$ in Figure~\ref{fig:corr_gan} clearly illustrates the higher intra-GM ad lower inter-GM fingerprint similarities.

%the diagonal elements show higher values compared to others, testifying to the similarity of fingerprints generated from different images of the same model.
\begin{table}
\centering{
%begin{center}
\caption{\small Binary classification performance for coordinated misinformation attack.}
\scalebox{1}{
\begin{tabular}{c|c|c}\hline
Method & AUC (\%) & Classification accuracy (\%) \\ \hline \hline
FEN & $83.5$ & $76.85$\\
FEN + PN & $\bf 87.3$ & $\bf 80.6$\\  \hline
\end{tabular}
}
\label{tab:cma} 
%\vspace{-3mm}
}
%\end{center}
\end{table}

\begin{table}[t]
\begin{center}
\caption{\small AUC for deepfake detection on the Celeb-DF dataset~\cite{21}. }
%Most previous methods only train on binary, genuine \vs fake labels. '*' denotes training with pixel-level supervision; these results are not directly comparable to all olthers.}
\scalebox{1}{
\begin{tabular}{c|c|c}
\hline
Method & Training Data & AUC ($\%$)  \\ \hline \hline
\multicolumn{3}{c}{Methods training with {\textit{pixel-level}} supervision}\\ \hline
Xception+Reg \cite{7} & DFFD & $64.4$\\
Xception+Reg \cite{7} & DFFD, UADFV & $71.2$\\ \hline
\multicolumn{3}{c}{Methods training with {\textit{image-level}} supervision}\\ \hline
Two-stream \cite{22} &\multirow{4}{*}{Private}& $53.8$   \\
Meso4 \cite{23} && $54.8$  \\ 
VA-LogReg \cite{26} && $55.1$  \\
DSP-FWA \cite{29} && $64.6$ \\ \hline
Multi-task \cite{27} &FF& $54.3$   \\\hline
Capsule \cite{28} &\multirow{9}{*}{FF++}& $57.5$\\
Xception-c40 \cite{1}& & $ 65.5$\\
Two-branch \cite{masi2020two}& & $73.4$\\
SPSL \cite{liu2021spatial}& & $\bf76.8$\\
SPSL \cite{liu2021spatial} (reproduced) & & $73.2$\\
Ours (fingerprint) & & $69.6$\\
Ours (image+fingerprint) & & $71.1$\\
Ours (image+fingerprint+phase) & & $74.6$\\
Ours (model parsing) & & $64.3$\\\hline
HeadPose \cite{24} & \multirow{5}{*}{UADFV}&$54.6$  \\
FWA \cite{25} & & $56.9$\\
Xception \cite{7}& & $52.2$\\
Xception+Reg \cite{7} & & $57.1$\\
%\rowcolor{LightGray}
Ours & & $\bf 64.7$\\ \hline
Xception \cite{7}& \multirow{2}{*}{DFFD} & $63.9$\\
%\rowcolor{LightGray}
Ours && $\bf 65.3$ \\ \hline
Xception \cite{7} & \multirow{2}{*}{DFFD, UADFV} & $67.6$\\
%\rowcolor{LightGray}
Ours & & $\bf 70.2$ \\
\hline \hline
\end{tabular}
}
\label{tab:deepfake}
\end{center}
%\vspace{-1mm}
\end{table}

\begin{table}
\centering{
%begin{center}
\caption{\small Classification rates of image attribution. The baseline results are cited from~\cite{11}.}
\scalebox{1}{
\begin{tabular}{l|cc}\hline
Method & CelebA & LSUN \\ \hline \hline
kNN & $28.00$ & $36.30$\\ 
Eigenface~\cite{71} & $53.28$ & -\\ 
PRNU~\cite{10} & $86.61$ & $67.84$ \\
\VA{Yu~\etal~\cite{11}} & $99.43$ & $98.58$ \\ \hline
%\rowcolor{LightGray}
Ours & $\bf99.66$ & $\bf 99.84$\\ 
%Ours$^+$ & ? & ?\\ 
\hline \hline
\end{tabular}
}
\label{tab:attr} 
%\vspace{-3mm}
}
%\end{center}
\end{table}

\subsection{Applications}

\minisection{Coordinated misinformation attack}
Our model parsing framework can be leveraged to estimate whether there exists a coordinated misinformation attack. 
That is, given two fake images, we hope to classify whether they are generated from the same GM or not.
We do so by computing the Cosine similarity between the hyperparameters parsed from the given two images.
First, we train our framework on $101$ GMs, and test on $15$ seen GMs and $15$ unseen GMs. \VA{The list of GMs are mentioned in the supplementary.}
%Then, we calculate the cosine similarity between the pair of images and use this to decide whether the two test images come from the same GM or not. 
To evaluate this task, we report the Area Under Curve (AUC) and the classification accuracy at the optimum threshold. 
The results are shown in Tab.~\ref{tab:cma} comparing two methods, just using FEN network and using both FEN and PN. 
We conclude that our framework using FEN and PN can identify whether two images came from the same source with around $80\%$ accuracy. Using only FEN network to compare the similarities of the fingerprint performs worse. This justifies the benefit of using parsed parameters for coordinated misinformation attack. 

\begin{figure}[t!]
\begin{center}
\includegraphics[width=\linewidth]{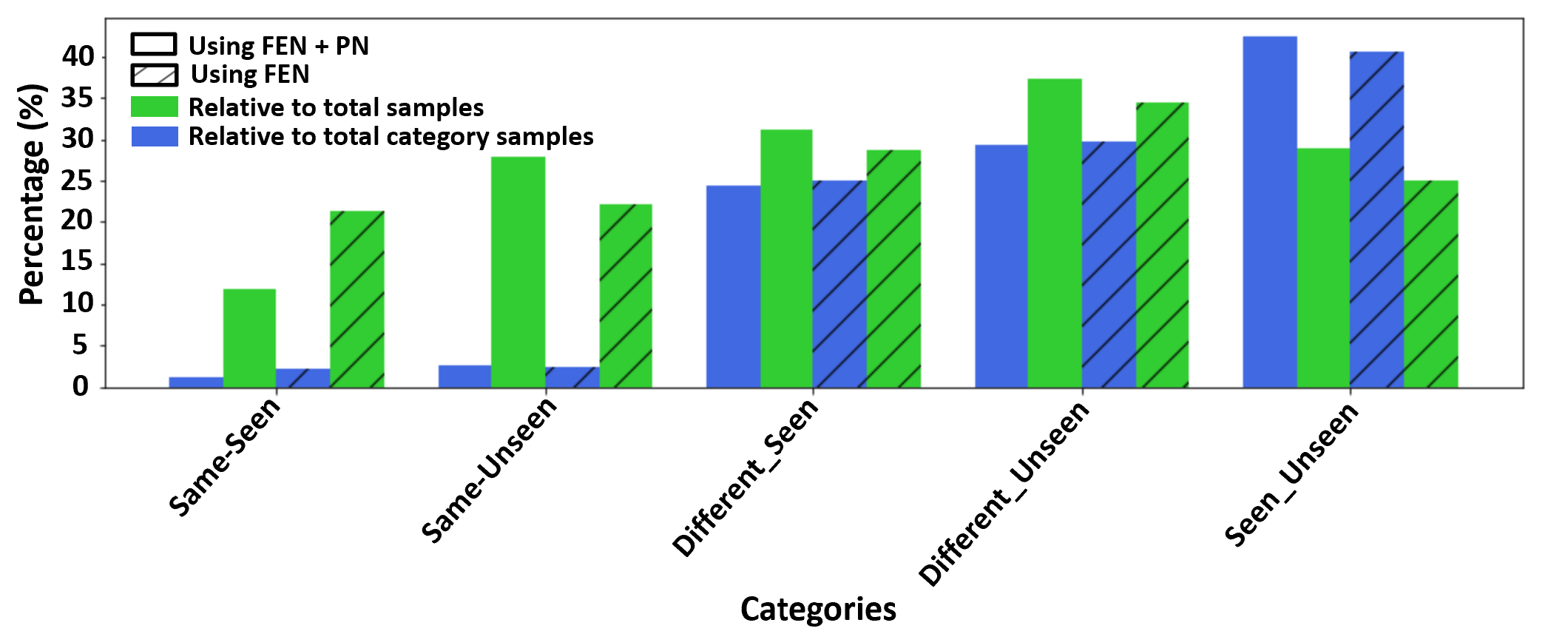}
\caption{\small Percentage of wrongly classified samples for five different categories of test sample pair. A larger number of sample pairs are wrongly classified if the pair of images come from same unseen GMs.}
\label{fig:seen_unseen_graph}
\end{center}
\end{figure}

In fact, due to the nature of our test set, each pair of test samples can come from five different categories, namely, $1.$ Same seen GM, $2.$ Same unseen GM, $3.$ Different seen GMs, $4.$ Different unseen GMs, and $5.$ One seen and one unseen GM. We show an analysis of the wrongly classified samples in Figure~\ref{fig:seen_unseen_graph} with respect to total number of samples and total number of samples in each category. Around $70\%$ of the wrongly classified samples \VA{belong to} the category of images coming from categories having atleast one GM unseen in training which is expected. However, if one of the test GM was seen in training, the number of wrongly classified samples decreased. This can be advantageous in detecting a manipulated image from an unknown GM.    

%\subsection{Deepfake detection and image attribution}
\minisection{Deepfake detection} Our FEN can be adopted for deepfake detection by adding a shallow network for binary classification. We evaluate our method on the recently introduced Celeb-DF dataset~\cite{21}. We experiment with three training sets, UADFV, DFFD, and FF++, in order to compare with previous results. We follow the same training protocols used in~\cite{7} for UADFV and DFFD and~\cite{liu2021spatial} for FF++. 
%; for Celeb-df we use the protocol defined for that benchmark~\cite{21}.

We report the AUC in Tab.~\ref{tab:deepfake}. Compared with methods trained on UADFV, our approach achieves a significantly better result, despite the more advanced backbones used by others. Our results when trained on DFFD and UADFV fall only slightly behind the best performance reported by Xception+Reg~\cite{7}. Importantly, however, they trained with pixel-level supervision which is typically unavailable. These results are provided for completeness, but are not directly comparable to all other methods trained with only image-level supervision for binary classification. Compared to all other methods, our method achieves the highest deepfake detection AUC.

Finally, we compare the performance of our method when trained on FF++ dataset. \cite{liu2021spatial} performs the best by using the phase information as an additional channel to the \VA{Xception} classifier. However, as the pre-trained models were not released for~\cite{liu2021spatial}, we reproduce their method and report the performance shown in Tab.~\ref{tab:deepfake}. We observe a performance gap between the reproduced and reported performance which should be further investigated in the future. Following~\cite{liu2021spatial}, we concatenate the fingerprint information with the RGB image and phase channels which are passed through a Xception classifier. Our method outperforms the reproduced performance of~\cite{liu2021spatial} showing the additional benefit of our fingerprint. Finally, we also perform the classification based on the pre-trained model parsing network and fine-tune it using the classification loss. The performance deteriorated compared to using the fingerprint. This shows that although the model parsing network have some deepfake detection abilities, they are less informative to perform deepfake detection well. 

% supervision during training whereas we do not. Moreover, the network used by Xception+Reg is more advanced, containing 23 million parameters vs. our 10 million. Importantly, our method was not directly optimized for fake detection, yet it generalize well to genuine vs. fake images as evident by these results.

% Despite their additional supervision and their larger network (23 million parameters vs. our 10 million) the gap in their favor is very small. 

%As described in Sec.~\ref{sec:other_app}, our method can be extended to deepfake detection. Specifically, we add a shallow network on the estimated fingerprint for binary classification. This shallow network is composed of two convolutional and seven fully-connected layers. We test our approach on the recently introduced Celeb-df dataset~\cite{21}, following the same protocols as they prescribe: We select key frames from all training and testing videos, and report accuracy using the Area Under Curve (AUC). Tab.~\ref{tab:deepfake} shows that our approach achieve results comparable to the SOTA. Importantly, our fingerprint estimation is designed to detect patterns from generated images. Our method is able generalize well to genuine images as evident by these results. 

\minisection{Image attribution}
Similar to deepfake detection, we use a shallow network for image attribution. The only difference is that image attribution is a multi-class task and depends on the number of GMs during training. Following~\cite{11}, we train our model on $100K$ genuine and $100K$ fake face images each from four GMs: SNGAN~\cite{60}, MMDGAN~\cite{54}, CRAMERGAN~\cite{48} and ProGAN~\cite{57}, for five-class classification. Tab.~\ref{tab:attr} reports the performance. Our result on CelebA~\cite{21} and LSUN~\cite{73} outperform the performance in~\cite{11}. This again validates the generalization ability of the proposed fingerprint estimation.

%%
%% =======================================================
%%

\section{Conclusion}

In this paper, we define the model parsing problem as inferring the network architectures and training loss functions of a GM from the generative images. We make the first attempt to tackle this \VA{challenging} problem. The main idea is to estimate the fingerprint for each image and use it for model parsing. Four constraints are developed for fingerprint estimation. We propose \VA{hierarchical} learning to parse the hyperparameters in coarse-level and fine-level that can leverage the similarities between different GMs. Our fingerprint estimation framework can not only perform model parsing, but also extend to detecting coordinated misinformation attack, deepfake detection and image attribution. We have collected a large-scale fake image dataset from $116$ different GMs. Various experiments have validated the effects of different components in our approach. 

\section*{Acknowledgement}

This work was partially supported by Facebook AI. This material, except Section 4.5 and related efforts, is based upon work partially supported by the
Defense Advanced Research Projects Agency (DARPA) under Agreement No.~HR00112090131 to Xiaoming Liu at Michigan State University.

% or


\ifCLASSOPTIONcaptionsoff
  \newpage
\fi



%\bibliographystyle{IEEEtran}
%\begin{thebibliography}{1}

%H.~Kopka and P.~W. Daly, \emph{A Guide to \LaTeX}, %3rd~ed.\hskip 1em plus

%\end{thebibliography}
\nocite{*}
{\small
\bibliographystyle{IEEEtran}
\bibliography{egbib}
}
%\iffalse

\begin{IEEEbiography}[{\includegraphics[width=1in,height=1.25in,clip,keepaspectratio]{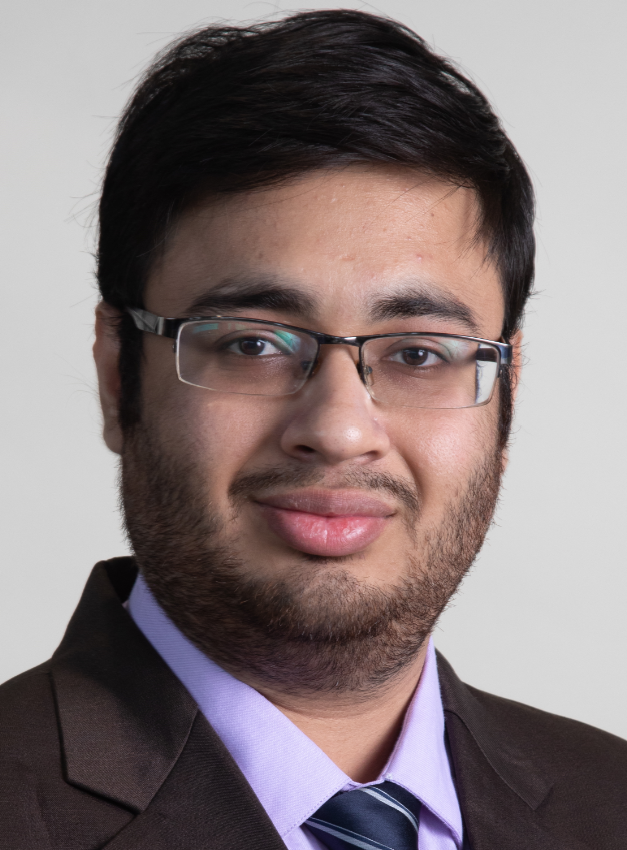}}]{Vishal Asnani}
is pursuing his Ph. D.~degree in  the Computer Science and Engineering department from Michigan State University since $2021$. He received his Bachelor's degree in Electrical and Instrumentation Engineering from Birla Institute of technology and Science, Pilani, India in $2019$. His research interests include computer vision and machine learning with a focus on the studying of generative models and deepfake detection. 
\end{IEEEbiography}

\begin{IEEEbiography}[{\includegraphics[width=1in,height=1.25in,clip,keepaspectratio]{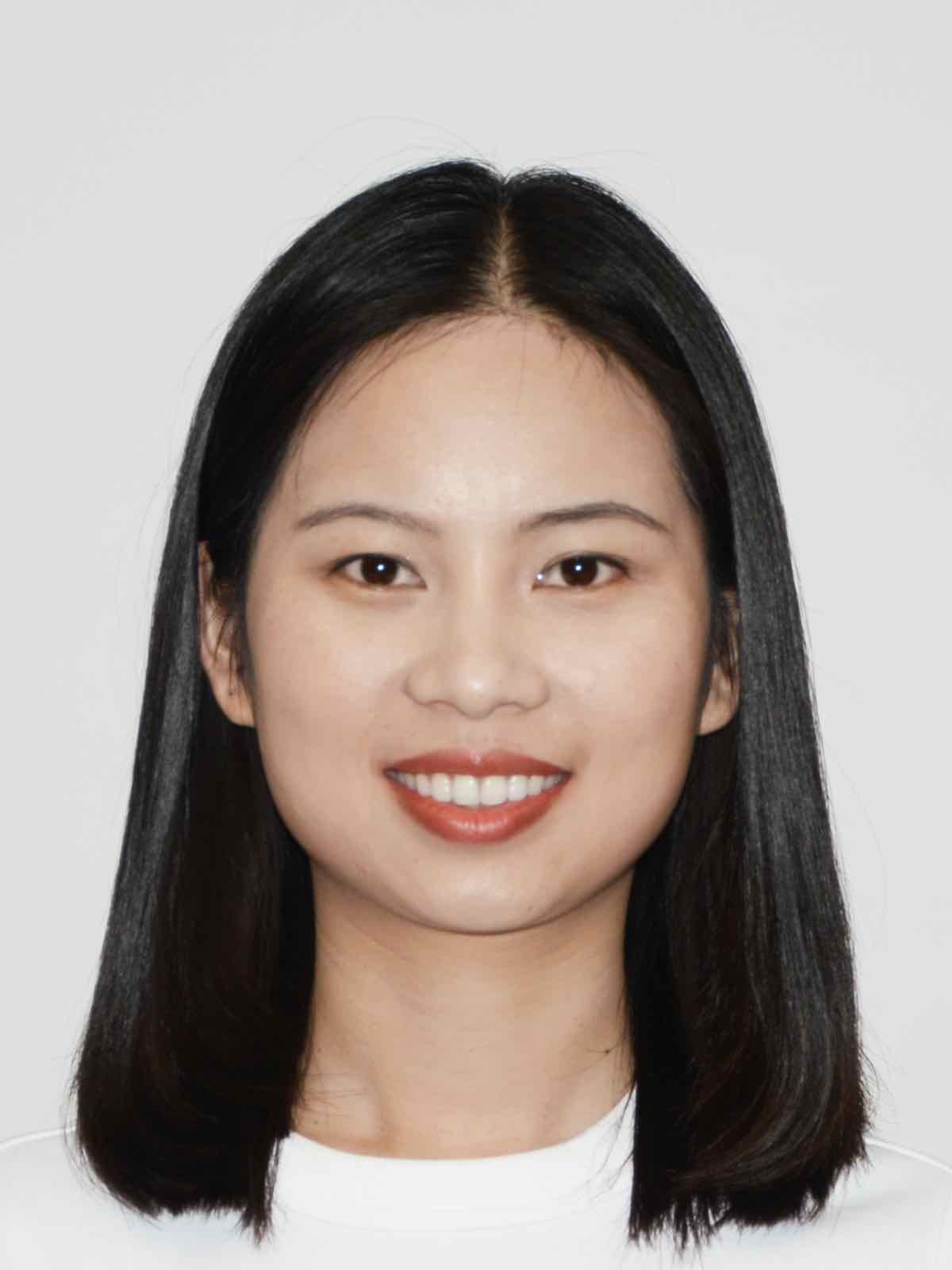}}]{Xi Yin}
    is a Research Scientist at Facebook AI Applied Research team. She received her Ph.D.~degree in Computer Science and Engineering from Michigan State University in $2018$. Before joining Facebook AI, she was an Senior Applied Scientist at Microsoft Cloud and AI. Her research is focused on computer vision, deep learning, vision and language. She has co-authored $18$ papers in top vision conferences and journals, and filed $3$ U.S.~patents. She has received Best Student Paper Award at WACV $2014$. She is an Area Chair for IJCB $2021$ and ICCV $2021$. 
\end{IEEEbiography}

\begin{IEEEbiography}[{\includegraphics[width=1in,height=1.25in,clip,keepaspectratio]{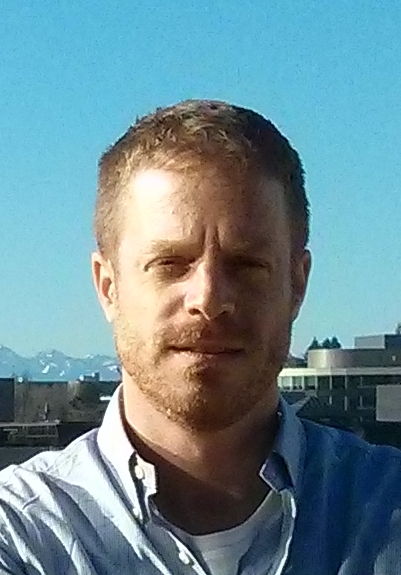}}]{Tal Hassner}
Tal Hassner received his M.Sc. and Ph.D. degrees in applied mathematics and computer science from the Weizmann Institute of Science in 2002 and 2006, respectively. In 2008 he joined the Department of Math. and Computer Science at The Open Univ. of Israel where he was an Associate Professor until 2018. From 2015 to 2018, he was a senior computer scientist at the Information Sciences Institute (ISI) and a Visiting Research Associate Professor at the Institute for Robotics and Intelligent Systems, Viterbi School of Engineering, both at USC, CA, USA. From 2018 to 2019, he was a principal applied scientist at AWS Rekognition. Since 2019 he is a research manager at Meta (formally Facebook). He served as a program chair at WACV'18, ICCV'21, and ECCV'22, a general chair for WACV'24, a workshop chair at CVPER'20, tutorial chair at ICCV'17, and area chair in CVPR, ECCV, AAAI, and others. Finally, he is an associate editor at IEEE-TPAMI and IEEE-TBIOM.  
\end{IEEEbiography}

\begin{IEEEbiography}[{\includegraphics[width=1in,height=1.25in,clip,keepaspectratio]{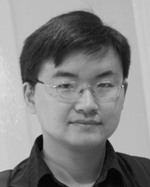}}]{Xiaoming Liu}
    is a MSU Foundation Professor at the Department of Computer Science and Engineering of Michigan State University. He received the Ph.D. degree in Electrical and Computer Engineering from Carnegie Mellon University in 2004. Before joining MSU in Fall $2012$, he was a research scientist at General Electric (GE) Global Research. His research interests include computer vision, machine learning, and biometrics. As a co-author, he is a recipient of Best Industry Related Paper Award runner-up at ICPR $2014$, Best Student Paper Award at WACV $2012$ and $2014$, Best Poster Award at BMVC $2015$, and Michigan State University College of Engineering Withrow Endowed Distinguished Scholar Award. He has been the Area Chair for numerous conferences, including CVPR, ICCV, ECCV, ICLR, NeurIPS, the Program CO-Chair of WACV'$18$, BTAS'$18$, AVSS'$22$ conferences, and General Co-Chair of FG'$23$ conference. He is an Associate Editor of Pattern Recognition Letters, Pattern Recognition, and IEEE Transactions on Image Processing. He has authored more than $150$ scientific publications, and has filed $29$ U.S.~patents. He is a fellow of IAPR.
\end{IEEEbiography}
    
\clearpage
\setcounter{equation}{0}
\setcounter{figure}{0}
\setcounter{table}{0}
\setcounter{section}{0}
\twocolumn[\centering \section*{\Large \textbf{Reverse Engineering of Generative Models:Inferring Model Hyperparameters from Generated Images \\ -- Supplementary material --\\[1cm]}}]

\begin{table*}[!t]
\begin{center}
\caption{\small Test sets used for evaluation. Each set contains six GANs, two VAEs, two ARs, one AA and one NF. }
\scalebox{1}{
\begin{tabular}{l|c|c|c|c}
\hline
GM & Set $1$ & Set $2$ & Set $3$ & Set $4$\\\hline
GM $1$ & ADV\_FACES & AAE & BICYCLE\_GAN & GFLM \\
GM $2$ & BETA\_B & ADAGAN\_C & BIGGAN\_512 & IMAGE\_GPT \\
GM $3$ & BETA\_TCVAE & BEGAN & CRGAN\_C & LSGAN \\
GM $4$ & BIGGAN\_128 & BETA\_H & FACTOR\_VAE & MADE \\
GM $5$ & DAGAN\_C & BIGGAN\_256 & FGSM & PIX2PIX \\
GM $6$ & DRGAN & COCOGAN & ICRGAN\_C & PROG\_GAN \\
GM $7$ & FGAN & CRAMERGAN & LOGAN & RSGAN\_REG \\
GM $8$ & PIXEL\_CNN & DEEPFOOL & MUNIT & SEAN \\
GM $9$ & PIXEL\_CNN++ & DRIT & PIXEL\_SNAIL & STYLE\_GAN \\
GM $10$ & RSGAN\_HALF & FAST\_PIXEL & STARGAN\_2 & SURVAE\_FLOW\_NONPOOL \\
GM $11$ & STARGAN & FVBN & SURVAE\_FLOW\_MAXPOOL & WGAN\_DRA \\
GM $12$ & VAEGAN & SRFLOW & VAE\_FIELD & YLG\\

\hline \hline
\end{tabular}}
\label{tab:test_sets}
\end{center}\vspace{-3mm}
\end{table*}

\section{Test sets for evaluation}
The experiments described in the text were performed on four different test sets, each set containing twelve different GMs for the leave out testing. For test sets, we follow the distribution of GMs as follows: six GANs, two VAEs, two ARs, one NF and one AA model. We select this distribution because of the number of GMs of each type in our dataset which has $81$ GANs, $13$ VAEs, 11 ARs, $5$ NFs and $6$ AAs. The sets considered are shown in Table~\ref{tab:test_sets}.

\begin{table*}
\begin{center}
\caption{\small Ground truth feature vector used for prediction of network architecture for all GMs. F$1$: \# layers, F$2$: \# convolutional layers, F$3$: \# fully connected layers, F$4$: \# pooling layers, F$5$: \# normalization layers, F$6$: \#filters, F$7$: \# blocks, F$8$:\# layers per block, F$9$: \# parameters, F$10$: normalization type, F$11$: non-linearity type in last layer, F$12$: non\-linearity type in blocks, F$13$: up-sampling type, F$14$: skip connection, F$15$: down\-sampling }
\label{tab:net_gt}
\scalebox{0.65}{
\begin{tabular}{|P{3.5cm}|P{0.7cm}|P{0.7cm}|P{0.7cm}|P{0.7cm}|P{0.7cm}|P{1cm}|P{0.7cm}|P{0.7cm}|P{1.5cm}|P{0.7cm}|P{0.7cm}|P{0.7cm}|P{0.7cm}|P{0.7cm}|P{0.7cm}|}\hline
% $\#$ lys & $\#$ conv. lys & $\#$ fc lys & $\#$ pool. lys & $\#$ norm. lys & $\#$ norm. lys & $\#$ params & $\#$ blocks & $\#$ lys. per block & $\#$ blocks w. branch & $\#$ branches & norm. type & up-samp. type & nonlin. type in blocks & nonlin. type in last ly. & skip conn. & down-samp. & attention\\
GM & F$1$ & F$2$ & F$3$ & F$4$ & F$5$ & F$6$ & F$7$ & F$8$ & F$9$ & F$10$ & F$11$ & F$12$ & F$13$ & F$14$ & F$15$  \\\hline
AAE & $9$ & $0$ & $7$ & $0$ & $2$ & $0$ & $0$ & $0$ & $1593378$ & $0$ & $1$ & $0$ & $0$ & $1$ & $0$ \\
ACGAN & $18$ & $10$ & $1$ & $0$ & $7$ & $2307$ & $5$ & $3$ & $4276739$ & $0$ & $1$ & $1$ & $0$ & $1$ & $0$ \\
ADAGAN\_C & $35$ & $14$ & $13$ & $1$ & $7$ & $4131$ & $9$ & $3$ & $9416196$ & $0$ & $1$ & $1$ & $0$ & $1$ & $0$ \\
ADAGAN\_P & $35$ & $14$ & $13$ & $1$ & $7$ & $4131$ & $9$ & $3$ & $9416196$ & $0$ & $1$ & $1$ & $0$ & $1$ & $0$ \\
ADV\_FACES & $45$ & $23$ & $1$ & $1$ & $20$ & $2627$ & $4$ & $6$ & $30000000$ & $1$ & $1$ & $1$ & $0$ & $1$ & $0$ \\
ALAE & $33$ & $25$ & $8$ & $0$ & $0$ & $4094$ & $3$ & $8$ & $50200000$ & $1$ & $2$ & $2$ & $1$ & $0$ & $1$ \\
BEGAN & $10$ & $9$ & $1$ & $0$ & $0$ & $515$ & $2$ & $4$ & $7278472$ & $0$ & $1$ & $0$ & $0$ & $0$ & $0$ \\
BETA\_B & $7$ & $4$ & $3$ & $0$ & $0$ & $99$ & $1$ & $3$ & $469173$ & $3$ & $3$ & $1$ & $0$ & $1$ & $1$ \\
BETA\_H & $7$ & $4$ & $3$ & $0$ & $0$ & $99$ & $1$ & $3$ & $469173$ & $3$ & $3$ & $1$ & $0$ & $1$ & $1$ \\
BETA\_TCVAE & $7$ & $4$ & $3$ & $0$ & $0$ & $99$ & $1$ & $3$ & $469173$ & $3$ & $3$ & $1$ & $0$ & $1$ & $1$ \\
BGAN & $8$ & $0$ & $5$ & $0$ & $3$ & $0$ & $2$ & $3$ & $1757412$ & $0$ & $1$ & $2$ & $0$ & $0$ & $0$ \\
BICYCLE\_GAN & $25$ & $14$ & $1$ & $0$ & $10$ & $4483$ & $2$ & $10$ & $23680256$ & $0$ & $1$ & $1$ & $0$ & $0$ & $0$ \\
BIGGAN\_128 & $63$ & $21$ & $1$ & $0$ & $41$ & $6123$ & $6$ & $10$ & $50400000$ & $0$ & $1$ & $1$ & $1$ & $1$ & $1$ \\
BIGGAN\_256 & $75$ & $25$ & $1$ & $0$ & $49$ & $7215$ & $6$ & $12$ & $55900000$ & $0$ & $1$ & $1$ & $1$ & $1$ & $1$ \\
BIGGAN\_512 & $87$ & $29$ & $1$ & $0$ & $57$ & $8365$ & $6$ & $14$ & $56200000$ & $0$ & $1$ & $1$ & $1$ & $1$ & $1$ \\
CADGAN & $8$ & $4$ & $1$ & $0$ & $3$ & $451$ & $3$ & $2$ & $3812355$ & $0$ & $1$ & $1$ & $0$ & $1$ & $1$ \\
CCGAN & $22$ & $12$ & $0$ & $0$ & $10$ & $3203$ & $2$ & $9$ & $29257731$ & $0$ & $1$ & $1$ & $1$ & $1$ & $1$ \\
CGAN & $8$ & $0$ & $5$ & $0$ & $3$ & $0$ & $2$ & $3$ & $1757412$ & $0$ & $1$ & $2$ & $0$ & $0$ & $0$ \\
COCO\_GAN & $19$ & $9$ & $1$ & $0$ & $9$ & $2883$ & $3$ & $4$ & $50000000$ & $0$ & $1$ & $1$ & $0$ & $0$ & $0$ \\
COGAN & $9$ & $5$ & $0$ & $0$ & $4$ & $259$ & $2$ & $2$ & $1126790$ & $0$ & $1$ & $2$ & $0$ & $1$ & $1$ \\
COLOUR\_GAN & $19$ & $10$ & $0$ & $0$ & $9$ & $2435$ & $2$ & $9$ & $19422404$ & $0$ & $1$ & $1$ & $0$ & $1$ & $1$ \\
CONT\_ENC & $19$ & $11$ & $0$ & $0$ & $8$ & $5987$ & $2$ & $8$ & $40401187$ & $0$ & $1$ & $2$ & $0$ & $1$ & $1$ \\
CONTRAGAN & $35$ & $14$ & $13$ & $1$ & $7$ & $4131$ & $9$ & $3$ & $9416196$ & $0$ & $1$ & $1$ & $0$ & $1$ & $0$ \\
COUNCIL\_GAN & $62$ & $30$ & $3$ & $0$ & $29$ & $6214$ & $2$ & $10$ & $69616944$ & $1$ & $1$ & $1$ & $0$ & $1$ & $0$ \\
CRAMER\_GAN & $9$ & $4$ & $1$ & $0$ & $4$ & $454$ & $2$ & $3$ & $9681284$ & $0$ & $1$ & $1$ & $0$ & $1$ & $0$ \\
CRGAN\_C & $35$ & $14$ & $13$ & $1$ & $7$ & $4131$ & $9$ & $3$ & $9416196$ & $0$ & $1$ & $1$ & $0$ & $1$ & $0$ \\
CRGAN\_P & $35$ & $14$ & $13$ & $1$ & $7$ & $4131$ & $9$ & $3$ & $9416196$ & $0$ & $1$ & $1$ & $0$ & $1$ & $0$ \\
CYCLEGAN & $47$ & $24$ & $0$ & $0$ & $23$ & $2947$ & $4$ & $9$ & $11378179$ & $1$ & $1$ & $1$ & $1$ & $1$ & $1$ \\
DAGAN\_C & $35$ & $14$ & $13$ & $1$ & $7$ & $4131$ & $9$ & $3$ & $9416196$ & $0$ & $1$ & $1$ & $0$ & $1$ & $0$ \\
DAGAN\_P & $35$ & $14$ & $13$ & $1$ & $7$ & $4131$ & $9$ & $3$ & $9416196$ & $0$ & $1$ & $1$ & $0$ & $1$ & $0$ \\
DCGAN & $9$ & $4$ & $1$ & $0$ & $4$ & $454$ & $2$ & $3$ & $9681284$ & $0$ & $1$ & $1$ & $0$ & $1$ & $0$ \\
DEEPFOOL & $95$ & $92$ & $1$ & $2$ & $0$ & $7236$ & $4$ & $10$ & $22000000$ & $2$ & $0$ & $1$ & $1$ & $0$ & $0$ \\
DFCVAE & $45$ & $22$ & $2$ & $0$ & $21$ & $4227$ & $4$ & $7$ & $2546234$ & $0$ & $3$ & $2$ & $0$ & $0$ & $1$ \\
DISCOGAN & $21$ & $12$ & $0$ & $0$ & $9$ & $3459$ & $2$ & $9$ & $29241731$ & $1$ & $1$ & $2$ & $1$ & $1$ & $1$ \\
DRGAN & $44$ & $28$ & $1$ & $1$ & $14$ & $4481$ & $3$ & $8$ & $18885068$ & $0$ & $1$ & $0$ & $0$ & $1$ & $1$ \\
DRIT & $19$ & $10$ & $0$ & $0$ & $9$ & $1793$ & $4$ & $3$ & $9564170$ & $1$ & $1$ & $1$ & $1$ & $1$ & $1$ \\
DUALGAN & $25$ & $14$ & $1$ & $0$ & $10$ & $4483$ & $2$ & $10$ & $23680256$ & $0$ & $1$ & $1$ & $0$ & $0$ & $0$ \\
EBGAN & $6$ & $3$ & $1$ & $0$ & $2$ & $195$ & $2$ & $2$ & $738433$ & $0$ & $1$ & $2$ & $0$ & $0$ & $1$ \\
ESRGAN & $66$ & $66$ & $0$ & $0$ & $0$ & $4547$ & $5$ & $4$ & $7012163$ & $2$ & $2$ & $2$ & $1$ & $0$ & $0$ \\
FACTOR\_VAE & $7$ & $4$ & $3$ & $0$ & $0$ & $99$ & $1$ & $3$ & $469173$ & $3$ & $3$ & $1$ & $0$ & $1$ & $1$ \\
Fast pixel & $17$ & $9$ & $0$ & $0$ & $8$ & $768$ & $2$ & $8$ & $4600000$ & $0$ & $3$ & $0$ & $0$ & $1$ & $0$ \\
FFGAN & $39$ & $19$ & $1$ & $1$ & $19$ & $3261$ & $0$ & $0$ & $50000000$ & $0$ & $1$ & $1$ & $1$ & $1$ & $1$ \\
FGAN & $5$ & $0$ & $3$ & $0$ & $2$ & $0$ & $2$ & $2$ & $2256401$ & $0$ & $3$ & $1$ & $0$ & $1$ & $0$ \\
FGAN\_KL & $5$ & $0$ & $3$ & $0$ & $2$ & $0$ & $2$ & $2$ & $2256401$ & $0$ & $3$ & $1$ & $0$ & $1$ & $0$ \\
FGAN\_NEYMAN & $5$ & $0$ & $3$ & $0$ & $2$ & $0$ & $2$ & $2$ & $2256401$ & $0$ & $3$ & $1$ & $0$ & $1$ & $0$ \\
FGAN\_PEARSON & $5$ & $0$ & $3$ & $0$ & $2$ & $0$ & $2$ & $2$ & $2256401$ & $0$ & $3$ & $1$ & $0$ & $1$ & $0$ \\
FGSM & $95$ & $92$ & $1$ & $2$ & $0$ & $7236$ & $4$ & $10$ & $22000000$ & $2$ & $0$ & $1$ & $1$ & $0$ & $0$ \\
FPGAN & $23$ & $12$ & $0$ & $0$ & $11$ & $2179$ & $2$ & $6$ & $53192576$ & $1$ & $1$ & $1$ & $0$ & $0$ & $1$ \\
FSGAN & $37$ & $20$ & $0$ & $1$ & $16$ & $2863$ & $4$ & $8$ & $94669184$ & $0$ & $0$ & $1$ & $1$ & $1$ & $1$ \\
FVBN & $28$ & $0$ & $28$ & $0$ & $0$ & $0$ & $1$ & $1$ & $307721$ & $2$ & $3$ & $0$ & $0$ & $1$ & $0$ \\
GAN\_ANIME & $25$ & $18$ & $0$ & $0$ & $7$ & $2179$ & $4$ & $6$ & $8467854$ & $1$ & $1$ & $1$ & $0$ & $1$ & $1$ \\
Gated\_pixel\_cnn & $32$ & $32$ & $0$ & $0$ & $0$ & $5433$ & $3$ & $10$ & $3364161$ & $2$ & $3$ & $2$ & $1$ & $1$ & $0$ \\
GDWCT & $79$ & $27$ & $40$ & $1$ & $11$ & $5699$ & $2$ & $4$ & $51965832$ & $1$ & $1$ & $1$ & $0$ & $0$ & $1$ \\
GFLM & $95$ & $92$ & $1$ & $2$ & $0$ & $7236$ & $4$ & $10$ & $22000000$ & $2$ & $0$ & $1$ & $1$ & $0$ & $0$ \\
GGAN & $8$ & $4$ & $1$ & $0$ & $3$ & $451$ & $3$ & $2$ & $3812355$ & $0$ & $1$ & $1$ & $0$ & $1$ & $1$ \\
ICRGAN\_C & $35$ & $14$ & $13$ & $1$ & $7$ & $4131$ & $9$ & $3$ & $9416196$ & $0$ & $1$ & $1$ & $0$ & $1$ & $0$ \\
ICRGAN\_P & $35$ & $14$ & $13$ & $1$ & $7$ & $4131$ & $9$ & $3$ & $9416196$ & $0$ & $1$ & $1$ & $0$ & $1$ & $0$ \\
Image\_GPT & $59$ & $42$ & $0$ & $0$ & $17$ & $4673$ & $7$ & $8$ & $401489$ & $0$ & $3$ & $2$ & $1$ & $1$ & $1$ \\
INFOGAN & $7$ & $3$ & $1$ & $0$ & $3$ & $195$ & $2$ & $2$ & $1049985$ & $0$ & $1$ & $2$ & $0$ & $0$ & $1$ \\
LAPGAN & $11$ & $6$ & $5$ & $0$ & $0$ & $262$ & $4$ & $2$ & $2182857$ & $2$ & $1$ & $1$ & $1$ & $1$ & $0$ \\
Lmconv & $105$ & $60$ & $10$ & $35$ & $0$ & $7156$ & $15$ & $5$ & $46000000$ & $2$ & $3$ & $0$ & $1$ & $1$ & $1$ \\
LOGAN & $35$ & $14$ & $13$ & $1$ & $7$ & $4131$ & $9$ & $3$ & $9416196$ & $0$ & $1$ & $1$ & $0$ & $1$ & $0$ \\
LSGAN & $9$ & $5$ & $0$ & $0$ & $4$ & $1923$ & $2$ & $4$ & $23909265$ & $0$ & $1$ & $1$ & $0$ & $0$ & $0$ \\
MADE & $2$ & $0$ & $2$ & $0$ & $0$ & $0$ & $1$ & $2$ & $12552784$ & $2$ & $3$ & $0$ & $0$ & $1$ & $0$ \\
MAGAN & $9$ & $5$ & $0$ & $0$ & $4$ & $963$ & $2$ & $3$ & $11140934$ & $0$ & $1$ & $1$ & $0$ & $1$ & $0$ \\
MEMGAN & $14$ & $7$ & $1$ & $0$ & $6$ & $1155$ & $3$ & $4$ & $4128515$ & $0$ & $1$ & $1$ & $0$ & $1$ & $0$ \\
MMD\_GAN & $9$ & $4$ & $1$ & $0$ & $4$ & $454$ & $2$ & $3$ & $9681284$ & $0$ & $1$ & $1$ & $0$ & $1$ & $0$ \\
MRGAN & $9$ & $4$ & $1$ & $0$ & $4$ & $451$ & $3$ & $2$ & $15038350$ & $0$ & $1$ & $1$ & $0$ & $1$ & $0$ \\
MSG\_STYLE\_GAN & $33$ & $25$ & $8$ & $0$ & $0$ & $4094$ & $3$ & $8$ & $50200000$ & $1$ & $2$ & $2$ & $1$ & $0$ & $1$ \\
MUNIT & $18$ & $15$ & $0$ & $0$ & $3$ & $3715$ & $2$ & $6$ & $10305035$ & $1$ & $0$ & $1$ & $1$ & $1$ & $1$ \\
NADE & $1$ & $0$ & $1$ & $0$ & $0$ & $0$ & $1$ & $1$ & $785284$ & $2$ & $3$ & $0$ & $0$ & $1$ & $0$ \\
OCFGAN & $9$ & $4$ & $1$ & $0$ & $4$ & $454$ & $2$ & $3$ & $9681284$ & $0$ & $1$ & $1$ & $0$ & $1$ & $0$ \\
PGD & $95$ & $92$ & $1$ & $2$ & $0$ & $7236$ & $4$ & $10$ & $22000000$ & $2$ & $0$ & $1$ & $1$ & $0$ & $0$ \\
PIX2PIX & $29$ & $16$ & $0$ & $0$ & $13$ & $5507$ & $2$ & $13$ & $54404099$ & $1$ & $1$ & $2$ & $1$ & $1$ & $1$ \\
PixelCNN & $17$ & $9$ & $0$ & $0$ & $8$ & $768$ & $2$ & $8$ & $4600000$ & $0$ & $3$ & $0$ & $0$ & $1$ & $0$ \\
PixelCNN++ & $105$ & $60$ & $10$ & $35$ & $0$ & $7156$ & $15$ & $5$ & $46000000$ & $2$ & $3$ & $0$ & $1$ & $1$ & $1$ \\
PIXELDA & $27$ & $14$ & $1$ & $0$ & $12$ & $835$ & $4$ & $6$ & $483715$ & $0$ & $1$ & $1$ & $1$ & $0$ & $0$ \\
PixelSnail & $90$ & $90$ & $0$ & $0$ & $0$ & $4051$ & $3$ & $10$ & $40000000$ & $2$ & $0$ & $3$ & $0$ & $1$ & $0$ \\
PROG\_GAN & $26$ & $25$ & $1$ & $0$ & $0$ & $4600$ & $3$ & $8$ & $46200000$ & $0$ & $3$ & $3$ & $0$ & $0$ & $1$ \\
RGAN & $7$ & $3$ & $1$ & $0$ & $3$ & $195$ & $2$ & $2$ & $1049985$ & $0$ & $1$ & $2$ & $0$ & $0$ & $1$ \\
RSGAN\_HALF & $8$ & $4$ & $1$ & $0$ & $3$ & $899$ & $3$ & $2$ & $13129731$ & $0$ & $1$ & $1$ & $0$ & $1$ & $0$ \\
RSGAN\_QUAR & $8$ & $4$ & $1$ & $0$ & $3$ & $451$ & $3$ & $2$ & $3812355$ & $0$ & $1$ & $1$ & $0$ & $1$ & $0$ \\
RSGAN\_REG & $8$ & $4$ & $1$ & $0$ & $3$ & $1795$ & $3$ & $2$ & $48279555$ & $0$ & $1$ & $1$ & $0$ & $1$ & $0$ \\
RSGAN\_RES\_BOT & $15$ & $7$ & $1$ & $0$ & $7$ & $963$ & $3$ & $4$ & $758467$ & $0$ & $1$ & $1$ & $1$ & $1$ & $0$ \\
RSGAN\_RES\_HALF & $15$ & $7$ & $1$ & $0$ & $7$ & $1155$ & $3$ & $4$ & $1201411$ & $0$ & $1$ & $1$ & $1$ & $1$ & $0$ \\
RSGAN\_RES\_QUAR & $15$ & $7$ & $1$ & $0$ & $7$ & $579$ & $3$ & $4$ & $367235$ & $0$ & $1$ & $1$ & $1$ & $1$ & $0$ \\
RSGAN\_RES\_REG & $15$ & $7$ & $1$ & $0$ & $7$ & $2307$ & $3$ & $4$ & $4270595$ & $0$ & $1$ & $1$ & $1$ & $1$ & $0$ \\
SAGAN & $11$ & $6$ & $1$ & $0$ & $4$ & $139$ & $2$ & $4$ & $16665286$ & $0$ & $1$ & $2$ & $0$ & $0$ & $0$ \\
SEAN & $19$ & $16$ & $0$ & $0$ & $0$ & $5062$ & $2$ & $7$ & $266907367$ & $3$ & $1$ & $1$ & $0$ & $1$ & $0$ \\
SEMANTIC & $23$ & $12$ & $0$ & $0$ & $11$ & $2179$ & $2$ & $6$ & $53192576$ & $1$ & $1$ & $1$ & $0$ & $0$ & $1$ \\
SGAN & $7$ & $3$ & $1$ & $0$ & $3$ & $195$ & $2$ & $2$ & $1049985$ & $0$ & $1$ & $2$ & $0$ & $0$ & $1$ \\
SNGAN & $23$ & $11$ & $1$ & $0$ & $11$ & $3871$ & $4$ & $5$ & $10000000$ & $0$ & $1$ & $1$ & $0$ & $1$ & $0$ \\
SOFT\_GAN & $8$ & $0$ & $5$ & $0$ & $3$ & $0$ & $2$ & $3$ & $1757412$ & $0$ & $1$ & $2$ & $0$ & $0$ & $0$ \\
SRFLOW & $66$ & $66$ & $0$ & $0$ & $2$ & $4547$ & $5$ & $4$ & $7012163$ & $2$ & $2$ & $0$ & $1$ & $0$ & $0$ \\
SRRNET & $74$ & $36$ & $1$ & $0$ & $37$ & $2819$ & $4$ & $16$ & $4069955$ & $0$ & $1$ & $1$ & $0$ & $1$ & $1$ \\
STANDARD\_VAE & $7$ & $4$ & $3$ & $0$ & $0$ & $99$ & $1$ & $3$ & $469173$ & $3$ & $3$ & $1$ & $0$ & $1$ & $1$ \\
STARGAN & $23$ & $12$ & $0$ & $0$ & $11$ & $2179$ & $2$ & $6$ & $53192576$ & $1$ & $1$ & $1$ & $0$ & $0$ & $1$ \\
STARGAN\_2 & $67$ & $26$ & $12$ & $4$ & $25$ & $4188$ & $4$ & $12$ & $94008488$ & $1$ & $2$ & $2$ & $0$ & $0$ & $1$ \\
STGAN & $19$ & $10$ & $0$ & $0$ & $9$ & $2953$ & $2$ & $5$ & $25000000$ & $0$ & $1$ & $2$ & $1$ & $1$ & $1$ \\
STYLEGAN & $33$ & $25$ & $8$ & $0$ & $0$ & $4094$ & $3$ & $8$ & $50200000$ & $1$ & $2$ & $2$ & $1$ & $0$ & $1$ \\
STYLEGAN\_2 & $33$ & $25$ & $8$ & $0$ & $0$ & $4094$ & $3$ & $8$ & $59000000$ & $1$ & $2$ & $2$ & $1$ & $0$ & $1$ \\
STYLEGAN2\_ADA & $33$ & $25$ & $8$ & $0$ & $0$ & $4094$ & $3$ & $8$ & $59000000$ & $1$ & $2$ & $2$ & $1$ & $0$ & $1$ \\
SURVAE\_FLOW\_MAXPOOL & $95$ & $90$ & $0$ & $5$ & $0$ & $6542$ & $2$ & $20$ & $25000000$ & $2$ & $0$ & $0$ & $0$ & $0$ & $0$ \\
SURVAE\_FLOW\_NONPOOL & $90$ & $90$ & $0$ & $0$ & $0$ & $6542$ & $2$ & $20$ & $25000000$ & $2$ & $0$ & $0$ & $0$ & $0$ & $0$ \\
TPGAN & $45$ & $31$ & $2$ & $1$ & $11$ & $5275$ & $0$ & $0$ & $27233200$ & $0$ & $3$ & $3$ & $0$ & $1$ & $1$ \\
UGAN & $9$ & $4$ & $1$ & $0$ & $4$ & $771$ & $2$ & $3$ & $4850692$ & $0$ & $3$ & $1$ & $0$ & $1$ & $0$ \\
UNIT & $43$ & $22$ & $0$ & $0$ & $21$ & $4739$ & $4$ & $8$ & $13131779$ & $1$ & $1$ & $1$ & $1$ & $1$ & $1$ \\
VAE\_field & $6$ & $0$ & $6$ & $0$ & $0$ & $0$ & $1$ & $3$ & $300304$ & $2$ & $3$ & $0$ & $0$ & $0$ & $0$ \\
VAE\_flow & $14$ & $0$ & $14$ & $0$ & $0$ & $0$ & $2$ & $4$ & $760448$ & $2$ & $3$ & $0$ & $0$ & $0$ & $0$ \\
VAEGAN & $17$ & $7$ & $2$ & $0$ & $8$ & $867$ & $2$ & $6$ & $26396740$ & $0$ & $1$ & $1$ & $0$ & $0$ & $1$ \\
VDVAE & $48$ & $42$ & $0$ & $6$ & $0$ & $3502$ & $3$ & $13$ & $41000000$ & $2$ & $0$ & $2$ & $1$ & $1$ & $1$ \\
WGAN & $9$ & $5$ & $0$ & $0$ & $4$ & $1923$ & $2$ & $4$ & $23909265$ & $0$ & $1$ & $1$ & $0$ & $0$ & $0$ \\
WGAN\_DRA & $18$ & $10$ & $1$ & $0$ & $7$ & $2307$ & $5$ & $3$ & $4276739$ & $0$ & $1$ & $1$ & $0$ & $1$ & $0$ \\
WGAN\_WC & $18$ & $10$ & $1$ & $0$ & $7$ & $2307$ & $5$ & $3$ & $4276739$ & $0$ & $1$ & $1$ & $0$ & $1$ & $0$ \\
WGANGP & $9$ & $5$ & $0$ & $0$ & $4$ & $1923$ & $2$ & $4$ & $23905841$ & $0$ & $1$ & $1$ & $0$ & $0$ & $0$ \\
YLG & $33$ & $20$ & $1$ & $2$ & $10$ & $5155$ & $5$ & $5$ & $42078852$ & $0$ & $1$ & $1$ & $1$ & $1$ & $1$ \\

\hline
\end{tabular}
}
\end{center}
\end{table*}

\iffalse
\begin{table}[t]

\begin{center}
\caption{Feature value for different labels of multi-class and binary features.}
\tiny
\label{tab:net_gt_labels}
\begin{adjustbox}{width=1\columnwidth}
\begin{tabular}{|c|c|c|}\hline
Feature & Label & Value \\\hline
\multirow{4}{*}{F$10$}& $0$ & Batch Normalization \\
& $1$ & Instance Normalization \\
& $2$ & Adaptive Instance Normalization \\
& $3$ & No Normalization \\\hline
\multirow{4}{*}{F$11$}& $0$ & ReLU \\
& $1$ & Tanh \\
& $2$ & Leaky\_ReLU  \\
& $3$ & Sigmoid  \\\hline
\multirow{4}{*}{F$12$}& $0$ & ELU \\
& $1$ & ReLU \\
& $2$ & Leaky\_ReLU  \\
& $3$ & Sigmoid  \\\hline
\multirow{2}{*}{F$13$}& $0$ & Nearest Neighbour \\
& $1$ & Deconvolution \\\hline
\multirow{2}{*}{F$14$ and F$15$}& $0$ & Feature used \\
& $1$ & Feature not used \\\hline
    
\end{tabular}
\end{adjustbox}
\end{center}
\end{table}
\fi

\begin{table}[t]

\begin{center}
\caption{Feature value for different labels of multi-class and binary features.}
\tiny
\label{tab:net_gt_labels}
\begin{adjustbox}{width=1\columnwidth}
\begin{tabular}{|c|c|c|}\hline
Feature & Label & Value \\\hline
\multirow{4}{*}{Normalization type}& $0$ & Batch Normalization \\
& $1$ & Instance Normalization \\
& $2$ & Adaptive Instance Normalization \\
& $3$ & No Normalization \\\hline
\multirow{4}{*}{Non-linearity type in last layer}& $0$ & ReLU \\
& $1$ & Tanh \\
& $2$ & Leaky\_ReLU  \\
& $3$ & Sigmoid  \\\hline
\multirow{4}{*}{Non-linearity type in blocks}& $0$ & ELU \\
& $1$ & ReLU \\
& $2$ & Leaky\_ReLU  \\
& $3$ & Sigmoid  \\\hline
\multirow{2}{*}{Upsampling type}& $0$ & Nearest Neighbour \\
& $1$ & Deconvolution \\\hline
\multirow{2}{*}{Skip connection and downsampling}& $0$ & Feature used \\
& $1$ & Feature not used \\\hline
    
\end{tabular}
\end{adjustbox}
\end{center}
\end{table}

\begin{table*}[t]
\begin{center}
\caption{Ground truth feature vector used for prediction of loss type for all GMs.}
\label{tab:loss_gt}
%\begin{adjustbox}{width=1.5\columnwidth}
\scalebox{0.68}{
\begin{tabular}{|P{3.5cm}|P{1cm}|P{1cm}|P{1cm}|P{1cm}|P{1cm}|P{1cm}|P{1cm}|P{1cm}|P{1cm}|P{1cm}|}\hline
GM&$L_1$&$L_2$&MSE&MMD&LS&WGAN&KL&Adversarial&Hinge&CE\\\hline
AAE & $1$ & $0$ & $0$ & $0$ & $0$ & $0$ & $0$ & $0$ & $0$ & $1$ \\
ACGAN & $1$ & $0$ & $0$ & $0$ & $0$ & $0$ & $0$ & $0$ & $0$ & $1$ \\
ADAGAN\_C & $0$ & $0$ & $0$ & $0$ & $1$ & $0$ & $0$ & $0$ & $0$ & $1$ \\
ADAGAN\_P & $0$ & $0$ & $0$ & $0$ & $1$ & $0$ & $0$ & $0$ & $0$ & $0$ \\
ADV\_FACES & $1$ & $0$ & $1$ & $0$ & $1$ & $0$ & $0$ & $0$ & $0$ & $0$ \\
ALAE & $0$ & $0$ & $1$ & $0$ & $1$ & $0$ & $0$ & $0$ & $0$ & $0$ \\
BEGAN & $1$ & $0$ & $0$ & $0$ & $0$ & $0$ & $0$ & $0$ & $0$ & $0$ \\
BETA\_B & $0$ & $0$ & $0$ & $0$ & $0$ & $0$ & $1$ & $0$ & $0$ & $1$ \\
BETA\_H & $0$ & $0$ & $0$ & $0$ & $0$ & $0$ & $1$ & $0$ & $0$ & $1$ \\
BETA\_TCVAE & $1$ & $0$ & $0$ & $0$ & $0$ & $0$ & $1$ & $0$ & $0$ & $1$ \\
BGAN & $0$ & $0$ & $0$ & $0$ & $1$ & $0$ & $0$ & $0$ & $0$ & $1$ \\
BICYCLE\_GAN & $1$ & $0$ & $1$ & $0$ & $0$ & $0$ & $1$ & $0$ & $0$ & $0$ \\
BIGGAN\_128 & $1$ & $0$ & $0$ & $0$ & $0$ & $0$ & $0$ & $0$ & $0$ & $0$ \\
BIGGAN\_256 & $1$ & $0$ & $0$ & $0$ & $0$ & $0$ & $0$ & $0$ & $0$ & $0$ \\
BIGGAN\_512 & $1$ & $0$ & $0$ & $0$ & $0$ & $0$ & $0$ & $0$ & $0$ & $0$ \\
CADGAN & $0$ & $0$ & $0$ & $1$ & $0$ & $0$ & $0$ & $0$ & $0$ & $0$ \\
CCGAN & $0$ & $0$ & $0$ & $0$ & $1$ & $0$ & $0$ & $1$ & $0$ & $0$ \\
CGAN & $0$ & $0$ & $1$ & $0$ & $1$ & $0$ & $0$ & $0$ & $0$ & $0$ \\
COCO\_GAN & $1$ & $1$ & $0$ & $0$ & $0$ & $1$ & $0$ & $0$ & $1$ & $0$ \\
COGAN & $0$ & $0$ & $0$ & $0$ & $1$ & $0$ & $0$ & $0$ & $0$ & $0$ \\
COLOUR\_GAN & $1$ & $0$ & $0$ & $0$ & $1$ & $0$ & $0$ & $0$ & $0$ & $0$ \\
CONT\_ENC & $0$ & $1$ & $0$ & $0$ & $1$ & $0$ & $0$ & $0$ & $0$ & $0$ \\
CONTRAGAN & $1$ & $0$ & $0$ & $0$ & $0$ & $0$ & $0$ & $1$ & $0$ & $1$ \\
COUNCIL\_GAN & $1$ & $0$ & $1$ & $0$ & $1$ & $0$ & $0$ & $0$ & $0$ & $0$ \\
CRAMER\_GAN & $0$ & $0$ & $0$ & $0$ & $0$ & $1$ & $0$ & $0$ & $0$ & $0$ \\
CRGAN\_C & $1$ & $1$ & $0$ & $0$ & $0$ & $0$ & $0$ & $0$ & $0$ & $1$ \\
CRGAN\_P & $1$ & $1$ & $0$ & $0$ & $0$ & $0$ & $0$ & $0$ & $0$ & $0$ \\
CYCLEGAN & $1$ & $0$ & $0$ & $0$ & $1$ & $0$ & $0$ & $0$ & $0$ & $0$ \\
DAGAN\_C & $1$ & $0$ & $0$ & $0$ & $0$ & $0$ & $0$ & $0$ & $0$ & $1$ \\
DAGAN\_P & $1$ & $0$ & $0$ & $0$ & $0$ & $0$ & $0$ & $0$ & $0$ & $0$ \\
DCGAN & $0$ & $0$ & $0$ & $0$ & $0$ & $0$ & $0$ & $0$ & $0$ & $1$ \\
DEEPFOOL & $1$ & $1$ & $0$ & $0$ & $0$ & $0$ & $0$ & $0$ & $0$ & $0$ \\
DFCVAE & $0$ & $1$ & $0$ & $0$ & $0$ & $0$ & $1$ & $0$ & $0$ & $1$ \\
DISCOGAN & $1$ & $0$ & $0$ & $0$ & $1$ & $0$ & $0$ & $0$ & $0$ & $0$ \\
DRGAN & $0$ & $0$ & $0$ & $0$ & $1$ & $0$ & $0$ & $0$ & $0$ & $1$ \\
DRIT & $1$ & $0$ & $0$ & $0$ & $1$ & $0$ & $0$ & $0$ & $0$ & $1$ \\
DUALGAN & $1$ & $0$ & $0$ & $0$ & $0$ & $1$ & $0$ & $0$ & $0$ & $0$ \\
EBGAN & $0$ & $1$ & $0$ & $0$ & $1$ & $0$ & $0$ & $1$ & $1$ & $0$ \\
ESRGAN & $1$ & $0$ & $0$ & $0$ & $1$ & $0$ & $0$ & $0$ & $0$ & $0$ \\
FACTOR\_VAE & $1$ & $0$ & $0$ & $0$ & $0$ & $0$ & $1$ & $0$ & $0$ & $1$ \\
Fast pixel & $0$ & $0$ & $0$ & $0$ & $0$ & $0$ & $0$ & $0$ & $0$ & $1$ \\
FFGAN & $1$ & $1$ & $0$ & $0$ & $1$ & $0$ & $0$ & $0$ & $0$ & $1$ \\
FGAN & $0$ & $0$ & $0$ & $0$ & $1$ & $0$ & $0$ & $1$ & $0$ & $0$ \\
FGAN\_KL & $1$ & $0$ & $0$ & $0$ & $0$ & $0$ & $0$ & $0$ & $0$ & $0$ \\
FGAN\_NEYMAN & $0$ & $1$ & $0$ & $0$ & $0$ & $0$ & $0$ & $0$ & $0$ & $0$ \\
FGAN\_PEARSON & $0$ & $0$ & $1$ & $0$ & $0$ & $0$ & $0$ & $0$ & $1$ & $0$ \\
FGSM & $0$ & $0$ & $0$ & $0$ & $1$ & $0$ & $0$ & $0$ & $0$ & $0$ \\
FPGAN & $1$ & $1$ & $0$ & $0$ & $1$ & $0$ & $0$ & $0$ & $0$ & $1$ \\
FSGAN & $1$ & $0$ & $0$ & $0$ & $1$ & $0$ & $0$ & $0$ & $0$ & $1$ \\
FVBN & $0$ & $0$ & $0$ & $0$ & $0$ & $0$ & $0$ & $0$ & $0$ & $1$ \\
GAN\_ANIME & $1$ & $1$ & $0$ & $0$ & $0$ & $1$ & $0$ & $0$ & $1$ & $0$ \\
Gated\_pixel\_cnn & $0$ & $0$ & $0$ & $0$ & $0$ & $0$ & $0$ & $0$ & $0$ & $1$ \\
GDWCT & $1$ & $0$ & $1$ & $0$ & $0$ & $0$ & $0$ & $0$ & $1$ & $0$ \\
GFLM & $0$ & $0$ & $1$ & $0$ & $0$ & $0$ & $0$ & $0$ & $0$ & $1$ \\
GGAN & $1$ & $0$ & $0$ & $0$ & $0$ & $0$ & $0$ & $0$ & $0$ & $0$ \\
ICRGAN\_C & $1$ & $1$ & $0$ & $0$ & $0$ & $0$ & $0$ & $0$ & $0$ & $1$ \\
ICRGAN\_P & $1$ & $1$ & $0$ & $0$ & $0$ & $0$ & $0$ & $0$ & $0$ & $0$ \\
Image\_GPT & $0$ & $0$ & $0$ & $0$ & $0$ & $0$ & $0$ & $0$ & $0$ & $1$ \\
INFOGAN & $0$ & $0$ & $1$ & $0$ & $1$ & $0$ & $0$ & $0$ & $0$ & $1$ \\
LAPGAN & $0$ & $0$ & $0$ & $0$ & $1$ & $0$ & $0$ & $0$ & $0$ & $0$ \\
Lmconv & $0$ & $0$ & $0$ & $0$ & $0$ & $0$ & $0$ & $0$ & $0$ & $1$ \\
LOGAN & $1$ & $1$ & $0$ & $0$ & $0$ & $0$ & $0$ & $1$ & $0$ & $0$ \\
LSGAN & $0$ & $0$ & $1$ & $0$ & $0$ & $0$ & $0$ & $0$ & $1$ & $0$ \\
MADE & $0$ & $0$ & $0$ & $0$ & $0$ & $0$ & $0$ & $0$ & $0$ & $1$ \\
MAGAN & $0$ & $0$ & $1$ & $0$ & $0$ & $0$ & $0$ & $0$ & $0$ & $0$ \\
MEMGAN & $0$ & $0$ & $0$ & $0$ & $1$ & $0$ & $0$ & $0$ & $0$ & $0$ \\
MMD\_GAN & $1$ & $0$ & $0$ & $1$ & $0$ & $0$ & $0$ & $0$ & $0$ & $0$ \\
MRGAN & $0$ & $0$ & $1$ & $0$ & $1$ & $0$ & $0$ & $0$ & $0$ & $0$ \\
MSG\_STYLE\_GAN & $0$ & $0$ & $0$ & $0$ & $1$ & $0$ & $0$ & $0$ & $0$ & $0$ \\
MUNIT & $1$ & $0$ & $0$ & $0$ & $1$ & $0$ & $0$ & $0$ & $0$ & $0$ \\
NADE & $0$ & $0$ & $0$ & $0$ & $0$ & $0$ & $0$ & $0$ & $0$ & $1$ \\
OCFGAN & $0$ & $0$ & $0$ & $1$ & $0$ & $0$ & $0$ & $0$ & $1$ & $0$ \\
PGD & $1$ & $1$ & $0$ & $0$ & $0$ & $0$ & $0$ & $0$ & $0$ & $0$ \\
PIX2PIX & $1$ & $0$ & $0$ & $0$ & $1$ & $0$ & $0$ & $0$ & $0$ & $0$ \\
PixelCNN & $0$ & $0$ & $0$ & $0$ & $0$ & $0$ & $0$ & $0$ & $0$ & $1$ \\
PixelCNN++ & $0$ & $0$ & $0$ & $0$ & $0$ & $0$ & $0$ & $0$ & $0$ & $1$ \\
PIXELDA & $0$ & $0$ & $0$ & $0$ & $1$ & $0$ & $0$ & $0$ & $1$ & $1$ \\
PixelSnail & $0$ & $0$ & $0$ & $0$ & $0$ & $0$ & $0$ & $0$ & $0$ & $1$ \\
PROG\_GAN & $0$ & $0$ & $0$ & $0$ & $0$ & $1$ & $0$ & $0$ & $1$ & $0$ \\
RGAN & $0$ & $0$ & $0$ & $0$ & $0$ & $1$ & $0$ & $0$ & $0$ & $0$ \\
RSGAN\_HALF & $0$ & $0$ & $0$ & $0$ & $0$ & $0$ & $0$ & $0$ & $0$ & $1$ \\
RSGAN\_QUAR & $0$ & $0$ & $0$ & $0$ & $0$ & $0$ & $0$ & $0$ & $0$ & $1$ \\
RSGAN\_REG & $0$ & $0$ & $0$ & $0$ & $0$ & $0$ & $0$ & $0$ & $0$ & $1$ \\
RSGAN\_RES\_BOT & $0$ & $0$ & $0$ & $0$ & $0$ & $0$ & $0$ & $0$ & $0$ & $1$ \\
RSGAN\_RES\_HALF & $0$ & $0$ & $0$ & $0$ & $0$ & $0$ & $0$ & $0$ & $0$ & $1$ \\
RSGAN\_RES\_QUAR & $0$ & $0$ & $0$ & $0$ & $0$ & $0$ & $0$ & $0$ & $0$ & $1$ \\
RSGAN\_RES\_REG & $0$ & $0$ & $0$ & $0$ & $0$ & $0$ & $0$ & $0$ & $0$ & $1$ \\
SAGAN & $0$ & $0$ & $0$ & $0$ & $1$ & $0$ & $0$ & $0$ & $0$ & $0$ \\
SEAN & $1$ & $0$ & $0$ & $0$ & $1$ & $0$ & $0$ & $0$ & $0$ & $0$ \\
SEMANTIC & $0$ & $1$ & $0$ & $0$ & $1$ & $0$ & $0$ & $0$ & $0$ & $0$ \\
SGAN & $0$ & $0$ & $0$ & $0$ & $1$ & $0$ & $0$ & $0$ & $0$ & $1$ \\
SNGAN & $0$ & $0$ & $0$ & $0$ & $1$ & $0$ & $0$ & $1$ & $0$ & $0$ \\
SOFT\_GAN & $0$ & $0$ & $0$ & $0$ & $1$ & $0$ & $0$ & $0$ & $0$ & $0$ \\
SRFLOW & $1$ & $0$ & $0$ & $0$ & $0$ & $0$ & $0$ & $0$ & $0$ & $1$ \\
SRRNET & $0$ & $1$ & $1$ & $0$ & $1$ & $0$ & $0$ & $0$ & $0$ & $1$ \\
STANDARD\_VAE & $0$ & $0$ & $0$ & $0$ & $0$ & $0$ & $1$ & $0$ & $0$ & $1$ \\
STARGAN & $1$ & $0$ & $0$ & $0$ & $1$ & $0$ & $0$ & $0$ & $0$ & $1$ \\
STARGAN\_2 & $1$ & $0$ & $0$ & $0$ & $1$ & $0$ & $0$ & $0$ & $0$ & $0$ \\
STGAN & $1$ & $0$ & $0$ & $0$ & $1$ & $1$ & $0$ & $0$ & $0$ & $0$ \\
STYLEGAN & $0$ & $1$ & $0$ & $0$ & $0$ & $1$ & $0$ & $0$ & $0$ & $0$ \\
STYLEGAN\_2 & $0$ & $1$ & $0$ & $0$ & $1$ & $0$ & $0$ & $0$ & $1$ & $0$ \\
STYLEGAN2\_ADA & $0$ & $1$ & $0$ & $1$ & $1$ & $0$ & $0$ & $0$ & $1$ & $0$ \\
SURVAE\_FLOW\_MAXPOOL & $0$ & $0$ & $0$ & $0$ & $0$ & $0$ & $1$ & $0$ & $0$ & $1$ \\
SURVAE\_FLOW\_NONPOOL & $0$ & $0$ & $0$ & $0$ & $0$ & $0$ & $1$ & $0$ & $0$ & $1$ \\
TPGAN & $1$ & $0$ & $0$ & $0$ & $0$ & $1$ & $0$ & $0$ & $0$ & $0$ \\
UGAN & $0$ & $0$ & $0$ & $0$ & $1$ & $0$ & $0$ & $0$ & $0$ & $0$ \\
UNIT & $0$ & $0$ & $0$ & $0$ & $1$ & $0$ & $1$ & $0$ & $0$ & $0$ \\
VAE\_field & $0$ & $0$ & $0$ & $0$ & $0$ & $0$ & $1$ & $0$ & $0$ & $1$ \\
VAE\_flow & $0$ & $0$ & $0$ & $0$ & $0$ & $0$ & $1$ & $0$ & $0$ & $1$ \\
VAEGAN & $1$ & $0$ & $0$ & $0$ & $1$ & $0$ & $1$ & $0$ & $0$ & $0$ \\
VDVAE & $0$ & $0$ & $0$ & $0$ & $0$ & $0$ & $1$ & $0$ & $0$ & $1$ \\
WGAN & $0$ & $0$ & $0$ & $0$ & $0$ & $1$ & $0$ & $0$ & $0$ & $0$ \\
WGAN\_DRA & $0$ & $0$ & $1$ & $0$ & $0$ & $1$ & $0$ & $0$ & $0$ & $0$ \\
WGAN\_WC & $0$ & $0$ & $0$ & $0$ & $0$ & $1$ & $0$ & $0$ & $0$ & $0$ \\
WGANGP & $0$ & $1$ & $0$ & $0$ & $0$ & $1$ & $0$ & $0$ & $0$ & $0$ \\
YLG & $0$ & $0$ & $0$ & $0$ & $0$ & $1$ & $0$ & $0$ & $0$ & $0$ \\

\hline
\end{tabular}
}
%\end{adjustbox}
\end{center}
\end{table*}

\begin{table*}
\begin{center}
\caption{\VA{\small Ground truth feature vector used for prediction of network architecture for evaluation on diffusion models. F$1$: \# layers, F$2$: \# convolutional layers, F$3$: \# fully connected layers, F$4$: \# pooling layers, F$5$: \# normalization layers, F$6$: \#filters, F$7$: \# blocks, F$8$:\# layers per block, F$9$: \# parameters, F$10$: normalization type, F$11$: non-linearity type in last layer, F$12$: non\-linearity type in blocks, F$13$: up-sampling type, F$14$: skip connection, F$15$: down\-sampling} }
\label{tab:net_gt_diff}
\scalebox{0.8}{
\begin{tabular}{|P{3.5cm}|P{0.7cm}|P{0.7cm}|P{0.7cm}|P{0.7cm}|P{0.7cm}|P{1cm}|P{0.7cm}|P{0.7cm}|P{1.5cm}|P{0.7cm}|P{0.7cm}|P{0.7cm}|P{0.7cm}|P{0.7cm}|P{0.7cm}|}\hline
\VA{GM} & \VA{F$1$} & \VA{F$2$} & \VA{F$3$} & \VA{F$4$} & \VA{F$5$} & \VA{F$6$} & \VA{F$7$} & \VA{F$8$} & \VA{F$9$} & \VA{F$10$} & \VA{F$11$} & \VA{F$12$} & \VA{F$13$} & \VA{F$14$} & \VA{F$15$}  \\\hline
\VA{ADM} & \VA{$134$} & \VA{$122$} & \VA{$12$} & \VA{$0$} & \VA{$0$} & \VA{$5000$} & \VA{$8$} & \VA{$12$} & \VA{$554000000$} & \VA{$1$} & \VA{$1$} & \VA{$1$} & \VA{$1$} & \VA{$1$} & \VA{$1$} \\
\VA{ADM-G} & \VA{$134$} & \VA{$122$} & \VA{$12$} & \VA{$0$} & \VA{$0$} & \VA{$5000$} & \VA{$8$} & \VA{$12$} & \VA{$600000000$} & \VA{$1$} & \VA{$1$} & \VA{$1$} & \VA{$1$} & \VA{$1$} & \VA{$1$} \\
\VA{DDPM} & \VA{$134$} & \VA{$122$} & \VA{$12$} & \VA{$0$} & \VA{$0$} & \VA{$5000$} & \VA{$8$} & \VA{$12$} & \VA{$554000000$} & \VA{$1$} & \VA{$1$} & \VA{$1$} & \VA{$1$} & \VA{$1$} & \VA{$1$} \\
\VA{DDIM} & \VA{$134$} & \VA{$122$} & \VA{$12$} & \VA{$0$} & \VA{$0$} & \VA{$5000$} & \VA{$8$} & \VA{$12$} & \VA{$554000000$} & \VA{$1$} & \VA{$1$} & \VA{$1$} & \VA{$1$} & \VA{$1$} & \VA{$1$} \\
\VA{LDM} & \VA{$134$} & \VA{$122$} & \VA{$12$} & \VA{$0$} & \VA{$0$} & \VA{$5000$} & \VA{$8$} & \VA{$12$} & \VA{$554000000$} & \VA{$1$} & \VA{$1$} & \VA{$1$} & \VA{$1$} & \VA{$1$} & \VA{$1$} \\
\VA{Stable-Diffusion} & \VA{$94$} & \VA{$84$} & \VA{$10$} & \VA{$0$} & \VA{$0$} & \VA{$5000$} & \VA{$8$} & \VA{$12$} & \VA{$552000000$} & \VA{$1$} & \VA{$1$} & \VA{$1$} & \VA{$1$} & \VA{$1$} & \VA{$1$} \\
\VA{GLIDE-Diffusion} & \VA{$90$} & \VA{$80$} & \VA{$10$} & \VA{$0$} & \VA{$0$} & \VA{$5000$} & \VA{$8$} & \VA{$12$} & \VA{$270000000$} & \VA{$1$} & \VA{$1$} & \VA{$1$} & \VA{$1$} & \VA{$1$} & \VA{$1$} \\
\hline
\end{tabular}
}
\end{center}
\end{table*}

\begin{table*}[t]
\begin{center}
\caption{\VA{Ground truth feature vector used for prediction of loss type for evaluation on diffusion models.}}
\label{tab:loss_gt_diff}
%\begin{adjustbox}{width=1.5\columnwidth}
\scalebox{0.8}{
\begin{tabular}{|P{3.5cm}|P{1cm}|P{1cm}|P{1cm}|P{1cm}|P{1cm}|P{1cm}|P{1cm}|P{1cm}|P{1cm}|P{1cm}|}\hline
\VA{GM}&\VA{$L_1$}&\VA{$L_2$}&\VA{MSE}&\VA{MMD}&\VA{LS}&\VA{WGAN}&\VA{KL}&\VA{Adversarial}&\VA{Hinge}&\VA{CE}\\\hline
\VA{ADM} & \VA{$0$} & \VA{$0$} & \VA{$1$} & \VA{$0$} & \VA{$0$} & \VA{$0$} & \VA{$0$} & \VA{$0$} & \VA{$0$} & \VA{$1$} \\
\VA{ADM-G}  & \VA{$0$} & \VA{$0$} & \VA{$1$} & \VA{$0$} & \VA{$0$} & \VA{$0$} & \VA{$0$} & \VA{$0$} & \VA{$0$} & \VA{$0$} \\
\VA{DDPM} & \VA{$0$} & \VA{$0$} & \VA{$1$} & \VA{$0$} & \VA{$0$} & \VA{$0$} & \VA{$0$} & \VA{$0$} & \VA{$0$} & \VA{$0$} \\
\VA{DDIM} & \VA{$0$} & \VA{$0$} & \VA{$1$} & \VA{$0$} & \VA{$0$} & \VA{$0$} & \VA{$0$} & \VA{$0$} & \VA{$0$} & \VA{$1$} \\
\VA{LDM} & \VA{$1$} & \VA{$0$} & \VA{$1$} & \VA{$0$} & \VA{$0$} & \VA{$0$} & \VA{$0$} & \VA{$0$} & \VA{$0$} & \VA{$0$} \\
\VA{Stable-Diffusion} & \VA{$0$} & \VA{$0$} & \VA{$1$} & \VA{$1$} & \VA{$0$} & \VA{$0$} & \VA{$0$} & \VA{$0$} & \VA{$0$} & \VA{$0$} \\
\VA{GLIDE-Diffusion} & \VA{$0$} & \VA{$0$} & \VA{$1$} & \VA{$1$} & \VA{$0$} & \VA{$0$} & \VA{$0$} & \VA{$0$} & \VA{$0$} & \VA{$1$} \\
\hline
\end{tabular}
}
%\end{adjustbox}
\end{center}
\end{table*}

\begin{table*}[!t]
\begin{center}
\caption{\VA{\small Test sets used for evaluation on diffusion models.} }
\scalebox{1}{
\begin{tabular}{l|c|c|c|c}
\hline
\VA{GM} & \VA{Set $1$} & \VA{Set $2$} & \VA{Set $3$} & \VA{Set $4$}\\\hline
\VA{GM $1$} & \VA{ADM} & \VA{DDPM} & \VA{Stable-diffusion} & \VA{GLIDE-Diffusion} \\
\VA{GM $2$} & \VA{ADM-G} & \VA{DDIM} & \VA{ADM-G} & \VA{DDIM} \\
\VA{GM $3$} & \VA{DDPM} & \VA{LDM} & \VA{GLIDE-Diffusion} & \VA{LDM} \\\hline \hline
\end{tabular}}
\label{tab:test_sets_diff}
\end{center}\vspace{-3mm}
\end{table*}

\begin{table*}[!t]
\begin{center}
\caption{\VA{\small Test sets used for coordinated misinformation attacks.} }
\scalebox{0.65}{
\begin{tabular}{l|c|c|c|c|c|c|c|c|c|c|c|c|c|c|c}
\hline
\VA{Type} & \VA{GM $1$} & \VA{GM $2$} & \VA{GM $3$} & \VA{GM $4$} & \VA{GM $5$} & \VA{GM $6$} & \VA{GM $7$} & \VA{GM $8$} & \VA{GM $9$} & \VA{GM $10$} & \VA{GM $11$} & \VA{GM $12$} & \VA{GM $13$} & \VA{GM $14$} & \VA{GM $15$}\\\hline
\VA{Seen GMs} & \VA{BETA\_B} & \VA{GAN\_ANIME} & \VA{RGAN} & \VA{DRIT} & \VA{PIX2PIX} & \VA{UNIT} & \VA{SAGAN} & \VA{DFCVAE} & \VA{LOGAN} & \VA{DAGAN\_C} & \VA{SRRNET} & \VA{LSGAN} & \VA{BIGGAN\_128} & \VA{RSGAN\_HALF} & \VA{BICYCLE\_GAN}\\
\VA{Unseen GMs} & \VA{DRGAN} & \VA{RSGAN\_REG} & \VA{MAGAN} & \VA{MADE} & \VA{ALAE} & \VA{ACGAN} & \VA{WGAN} & \VA{TPGAN} & \VA{LAPGAN} & \VA{BETA\_TCVAE} & \VA{BGAN} & \VA{FFGAN} & \VA{CRGAN\_C} & \VA{FGAN} & \VA{STARGAN}\\\hline \hline
\end{tabular}}
\label{tab:test_sets_diff}
\end{center}\vspace{-3mm}
\end{table*}

\begin{figure*}[t]
\begin{center}
\includegraphics[width=\linewidth]{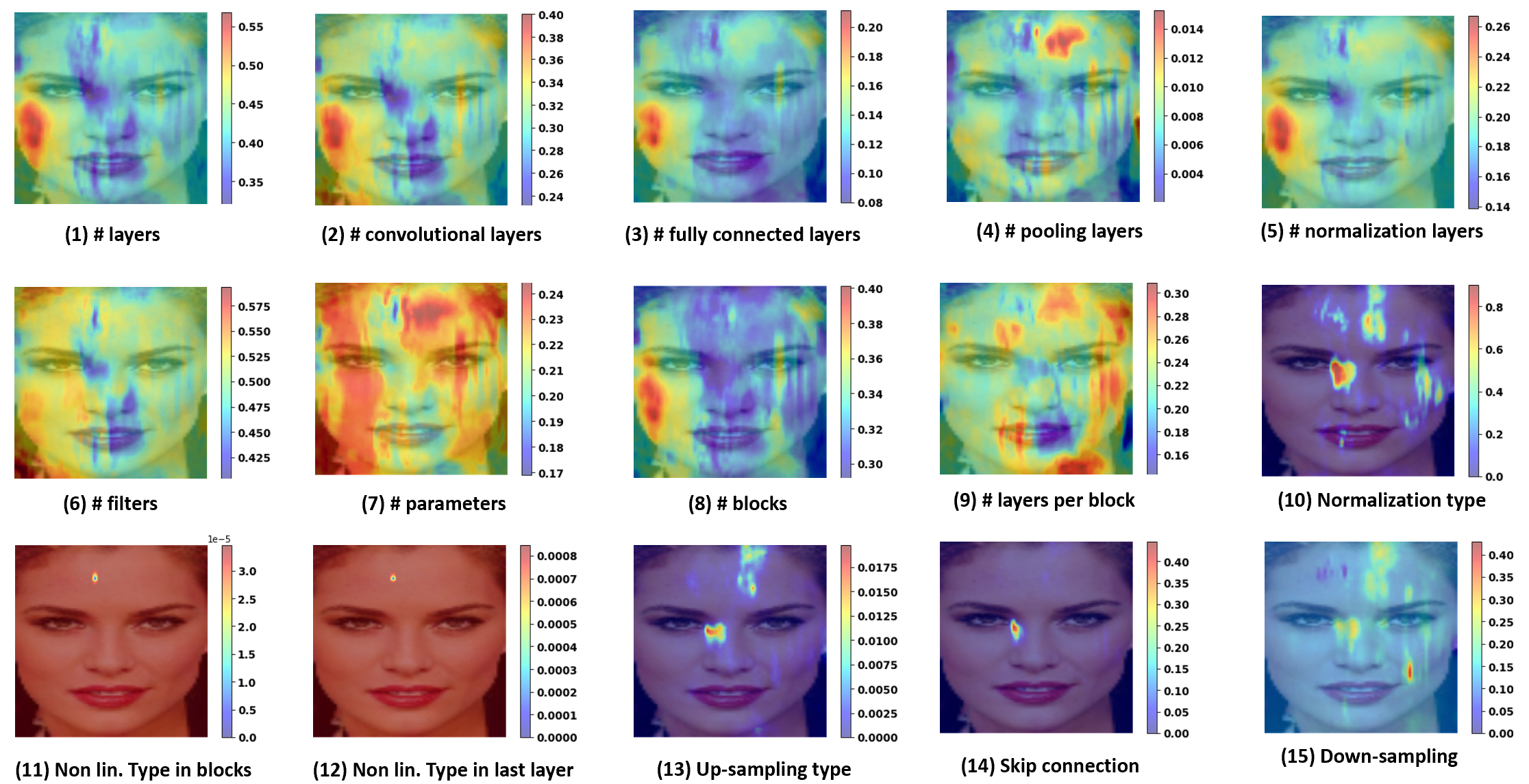}
\includegraphics[width=\linewidth]{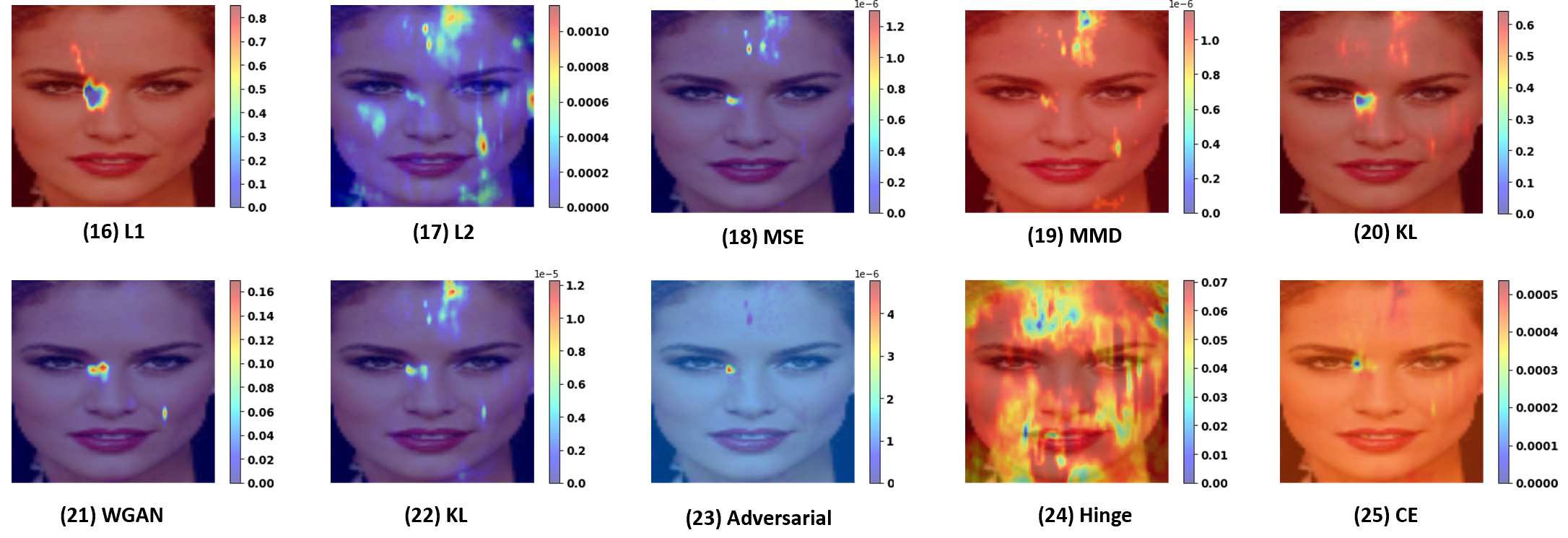}
\caption{Feature heatmap for each feature in network architecture and loss function predicted feature vector for face data. Each heatmap provides the importance of the region in the estimation of the respective parameter. }
\label{fig:heat_face}
\end{center}
\end{figure*}

\begin{figure*}[t]
\begin{center}
\includegraphics[width=\linewidth]{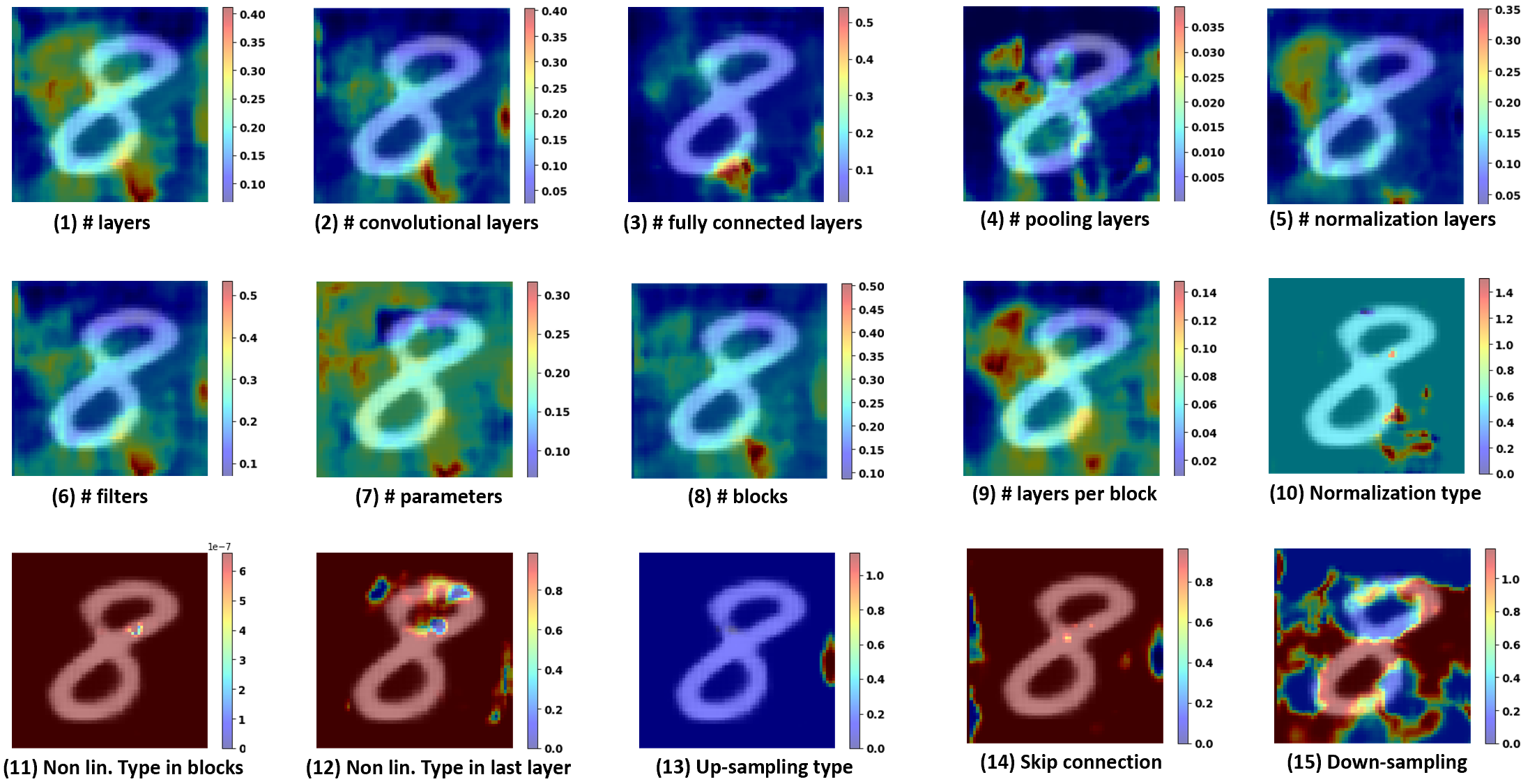}
\includegraphics[width=\linewidth]{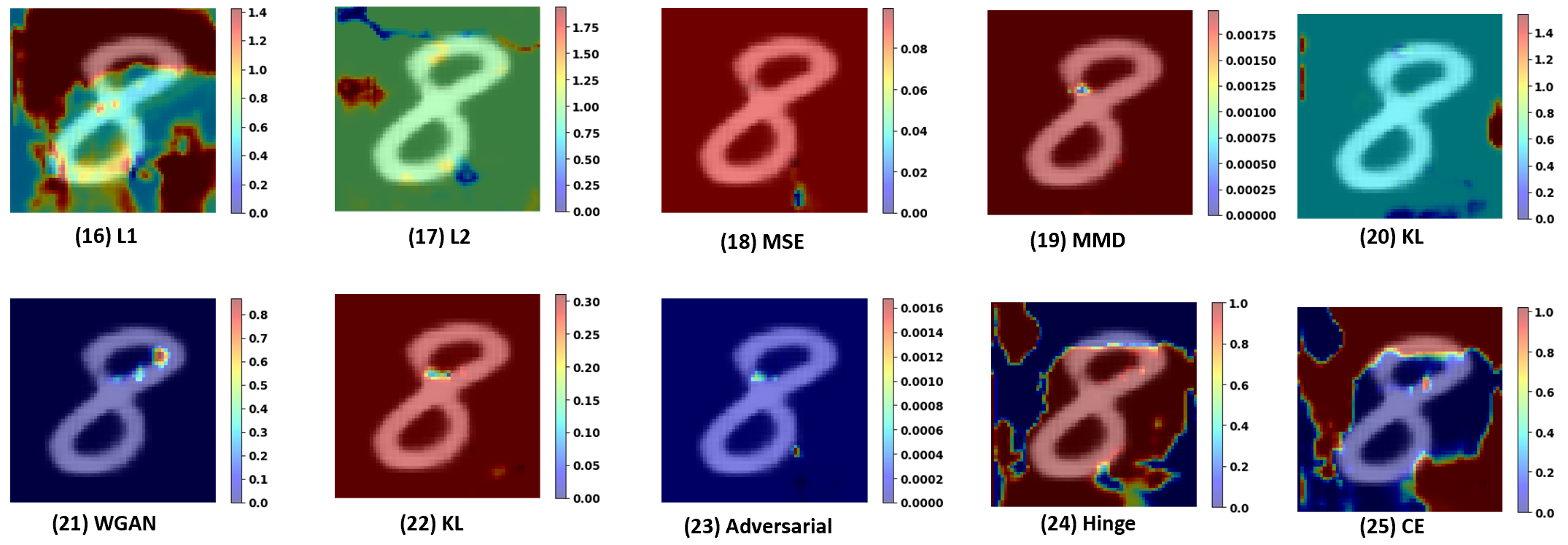}
\caption{Feature heatmap for each feature in network architecture and loss function predicted feature vector for MNIST data. }
\label{fig:heat_mnist}
\end{center}
\end{figure*}

\begin{figure*}[t]
\begin{center}
\includegraphics[width=\linewidth]{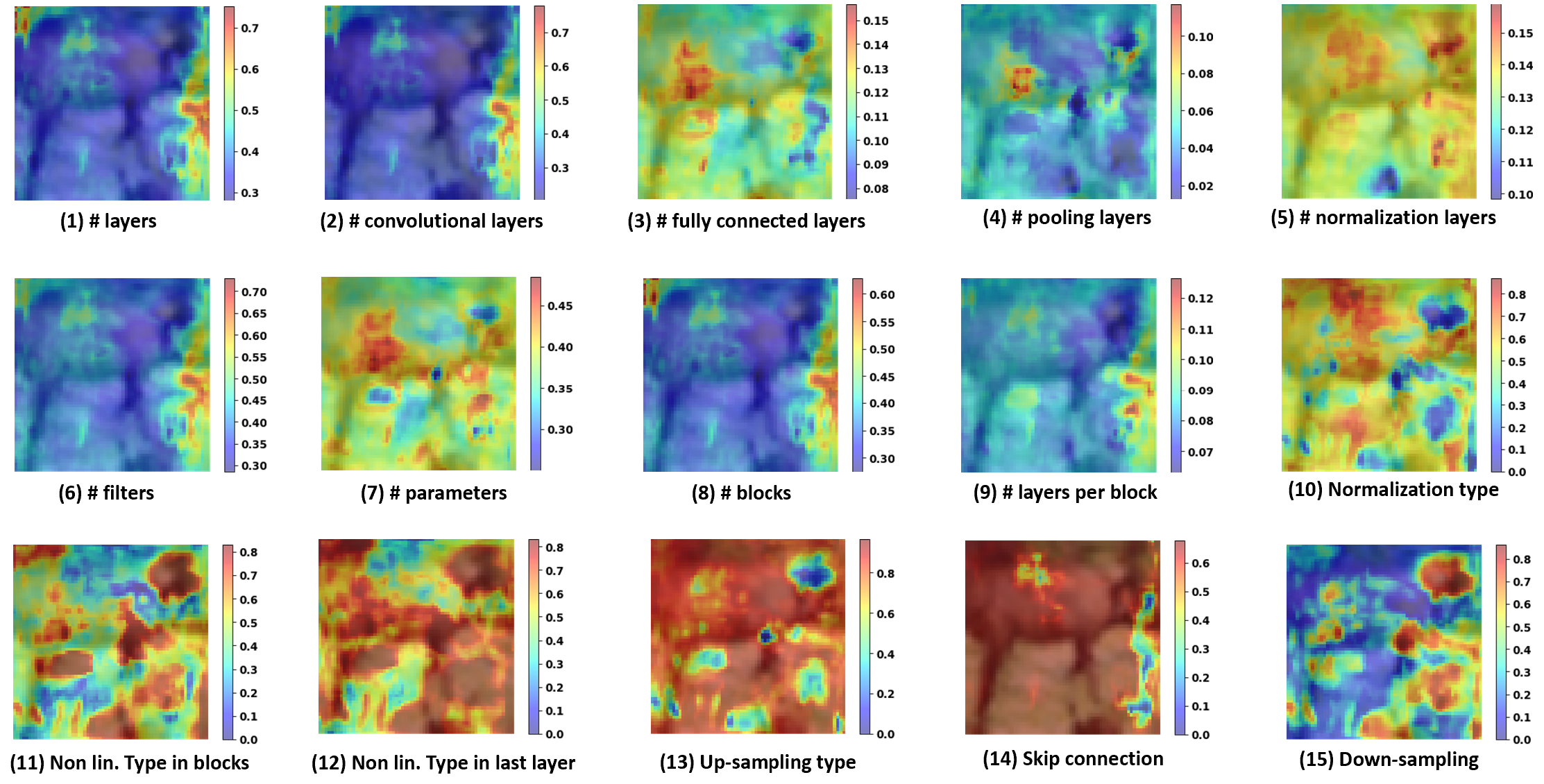}
\includegraphics[width=\linewidth]{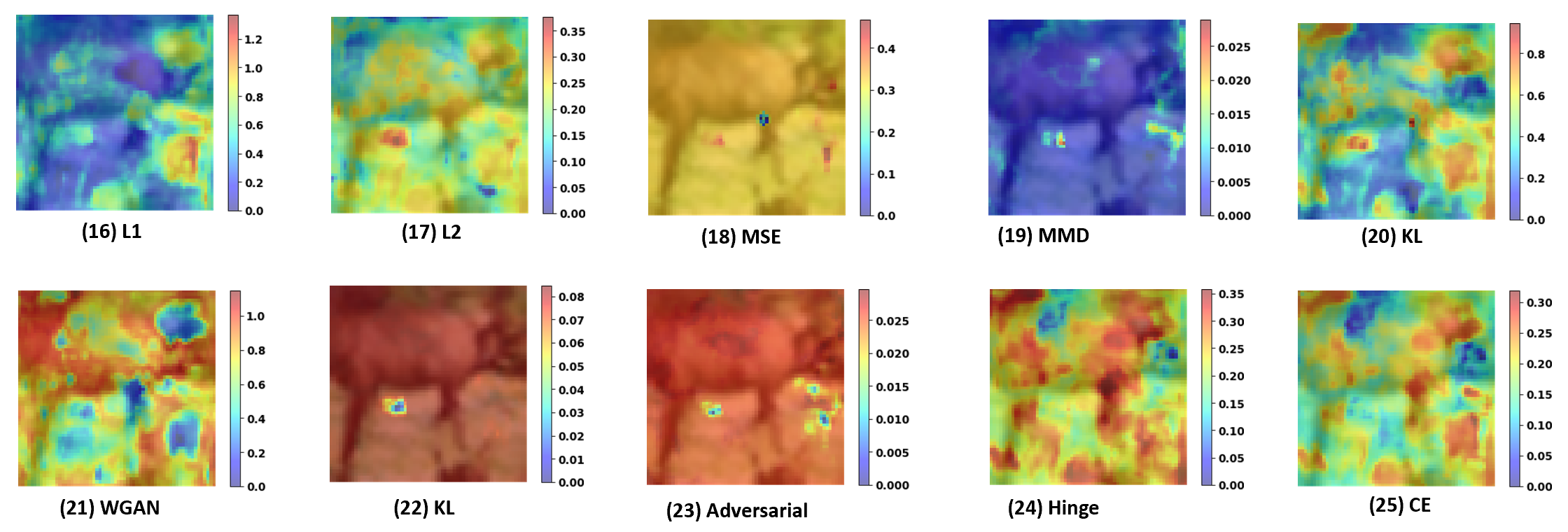}
\caption{Feature heatmap for each feature in network architecture and loss function predicted feature vector for CIFAR data. }
\label{fig:heat_cifar}
\end{center}
\end{figure*}

\begin{figure*}[t]
\begin{center}
\includegraphics[width=0.95\linewidth]{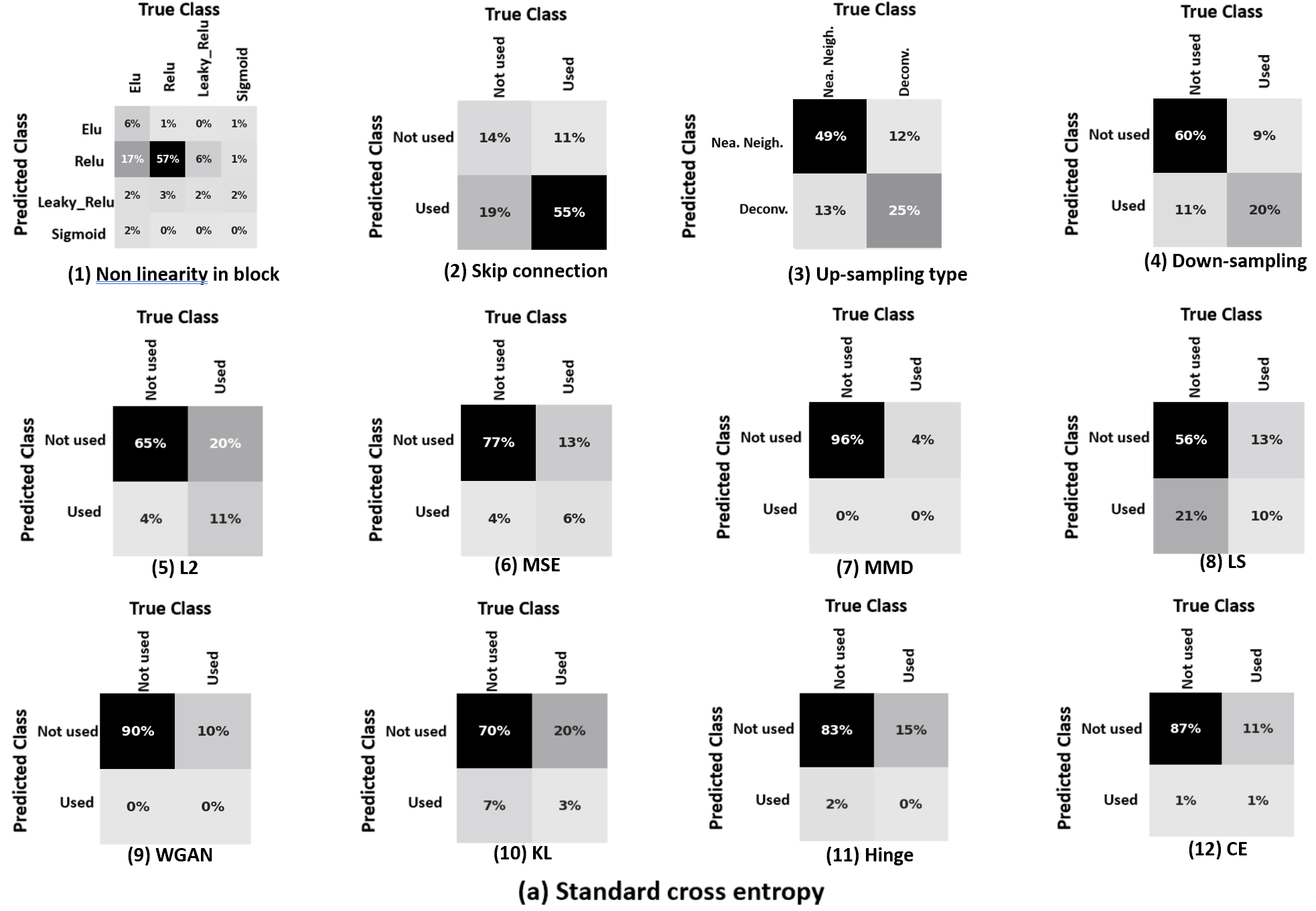}
\includegraphics[width=0.95\linewidth]{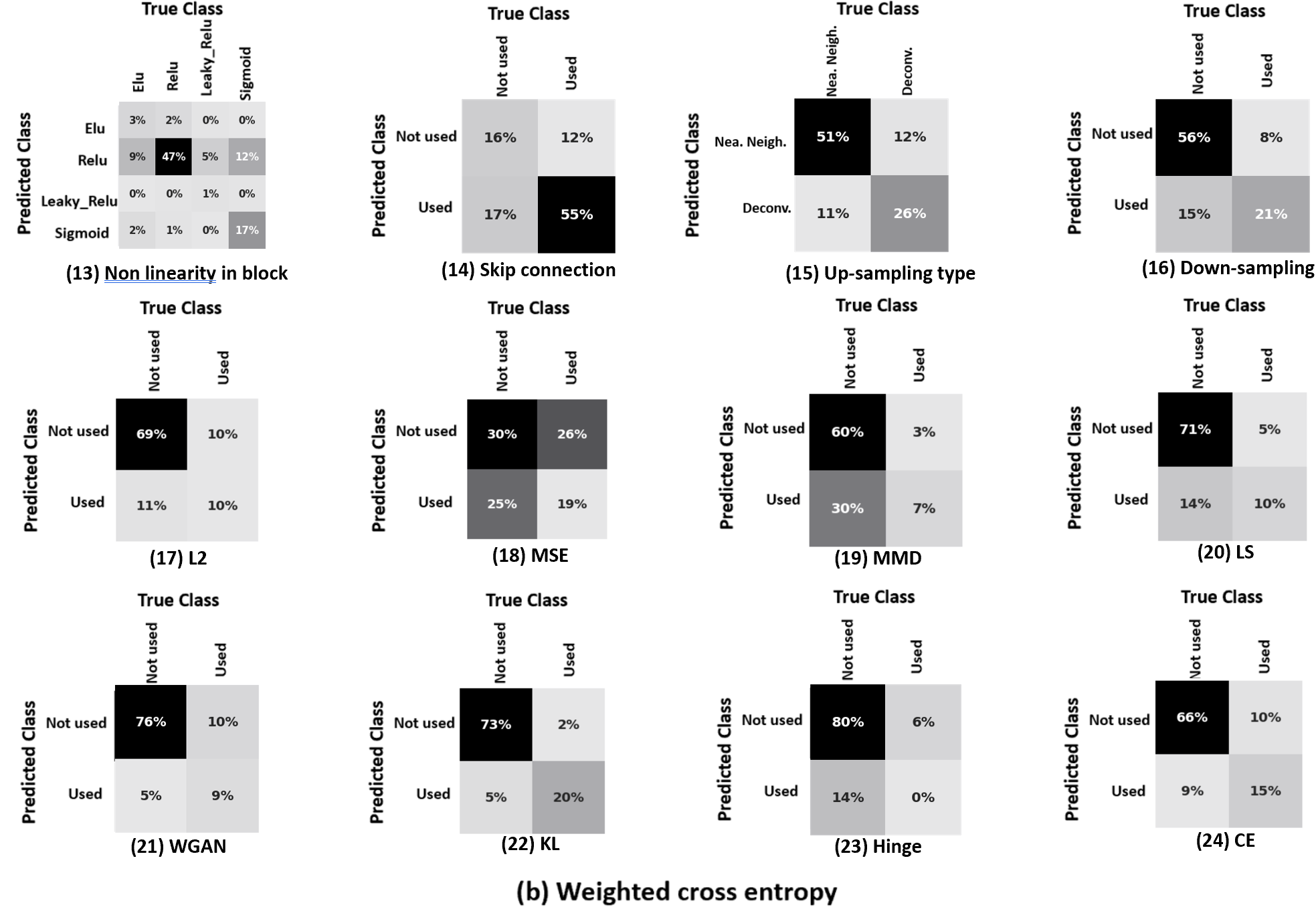}
\caption{Confusion matrix in the estimation of remaining parameters which were not shown in paper for network architecture and loss function. (1)-(12): Standard cross-entropy and
(12)-(24): Weighted cross entropy. Weighted cross entropy handles imbalance of data much better than the standard cross entropy which usually
predicts one class. }
\label{fig:con_mat_rem}
\end{center}
\end{figure*}

\section{Ground truth for GMs}
We collected a fake face dataset of $116$ GMs, each of them with $1,000$ generated images. We also collect the ground truth hyperparameters for network architecture and loss function types. Table~\ref{tab:net_gt} shows the ground truth representation of the network architecture where different hyperparameters are of different data types. Therefore, we apply min-max normalization for the continuous type parameters to make all values in the range of $[0,1]$. For multi-class and binary labels, we further show the feature value for different labels in Table~\ref{tab:net_gt_labels}. Note that some parameters share the same values but with different meanings. For example, F14 and F15 represent skip connection and down-sampling respectively. Table~\ref{tab:loss_gt} shows the ground truth representation of the loss function types used to train each GM where all these values are binary indicating whether the particular loss type was used or not.

\begin{figure*}[t]
\begin{center}
\includegraphics[width=0.95\linewidth]{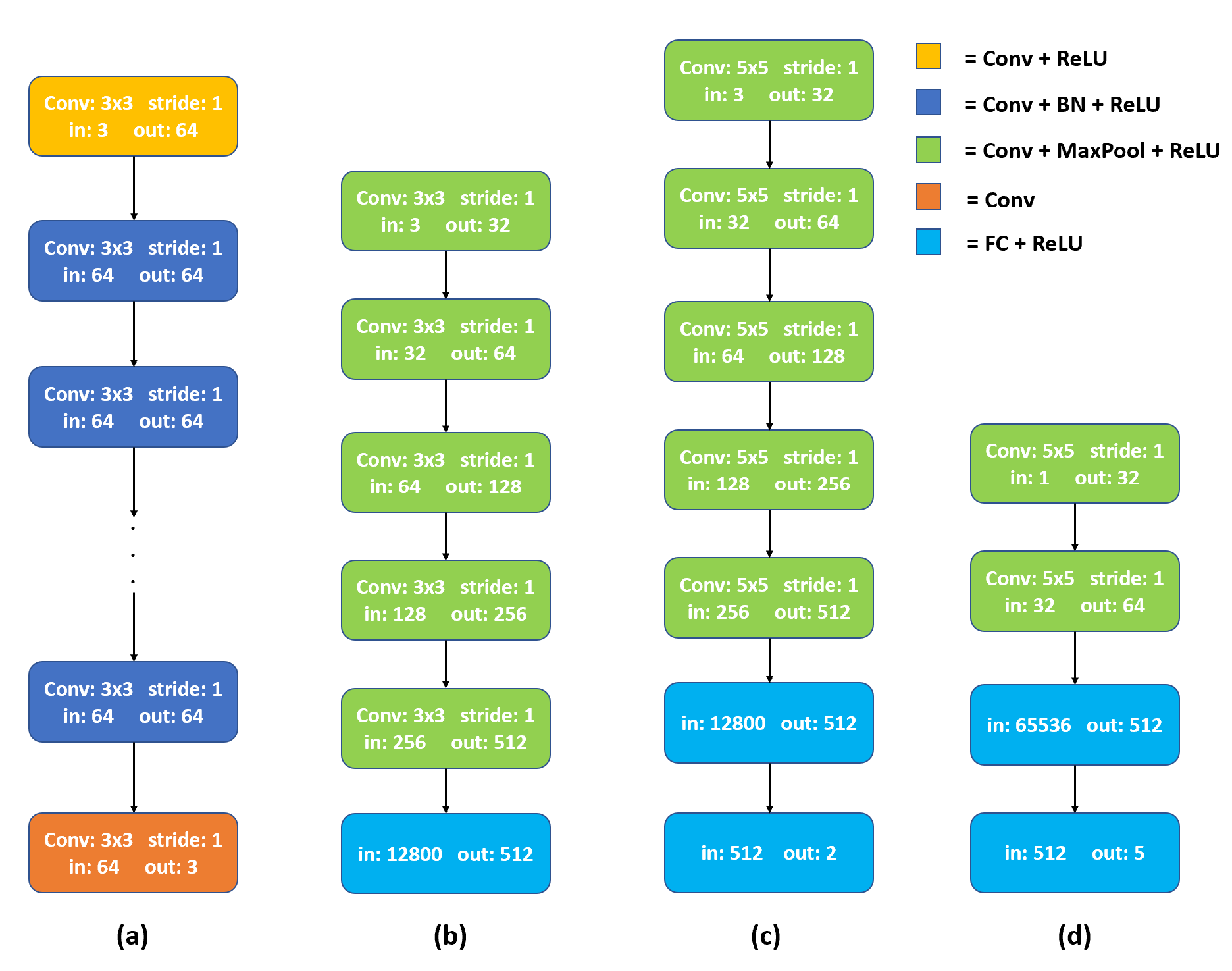}
\caption{Network architecture for various components of our method. (a) FEN (b) Mean and instance parser in PN (c) Shallow network for deepfake detection (d) Shallow network for image attribution.}
\label{fig:net_arch}
\end{center}
\end{figure*}

\section{Network architecture}
Figure~\ref{fig:net_arch} shows the network architecture used in different experiments. For GM parsing, our FEN has two stem convolution layers and $15$ convolution blocks with each block having convolution, batch normalization and ReLU activation to estimate the fingerprint. The encoder in the PN has five convolution blocks with each block having convolution, pooling and ReLU activation. This is followed by two fully connected layers to output a $512$ dimension feature vector which is further given as input to multiple branches to output different predictions. For continuous type parameters, we use two fully connected layers to output a $9$-D network architecture. For discrete type parameters and loss function parameters, we use separate classifiers with three fully connected layers for every parameter to perform multi-class or binary classification. 

For the deepfake detection task, we change the architecture of our FEN network as current deepfake manipulation detection requires much deeper networks. Thus, our FEN architecture has two stem convolution layers and $29$ convolution blocks to estimate the fingerprint. For further classification, we use a shallow network of five convolution blocks followed by two fully connected layers. 

For the image attribution task, we use the same FEN as used in model parsing, and a shallow network of two convolution blocks and two fully connected layers to perform multi-class classification.

\section{Feature heatmaps}
Every hyperparameter defined for network architecture and loss function type prediction may depend on certain region of the input image. To find out which region of the input image our model is looking at to predict each hyperparameter, we mask out $5\times5$ region from the input image. For the continuous type parameters, we compute the $L_1$ error between every predicted hyperparameter and its ground truth. This value of error will tell us how important is this $5\time5$ region in the input image to predict a particular hyperparameter. The higher the value of this error, the higher is the importance of that region in the prediction of the corresponding hyperparameter. For discete type parameters in network architecture and loss function, we estimate the probability of the ground truth label for every parameter. We subtract this probability from one to estimate the heatmap of the respective feature. Important regions will not affect the probability of the ground truth label for a particular feature. To obtain a stable heatmap, we do the above experiment on $100$ randomly chosen images across the different GMs and then calculate the average heatmap. 

Figure~\ref{fig:heat_face},~\ref{fig:heat_mnist} and~\ref{fig:heat_cifar} show the feature heatmaps for every hyperparameter of network architecture and loss type feature vector for Face, MNIST and CIFAR data respectively. For each hyperparmater, there are certain regions of the input that are more important than others. Each type of data has different type of heatmaps indicating different regions of importance. For face and CIFAR, these regions lie mostly in the central part but for MNIST, many of the features depend on the regions closer to edges. There are also some similarities between these heatmaps for a particular type of data. This can indicate the similarity of these hyperparameters.

%\fi

%\vfill


\end{document}